\documentclass[review]{elsarticle}

\usepackage{amsmath}
\usepackage{bm}
\usepackage{amsfonts}
\usepackage{lineno,hyperref,cleveref}
\usepackage{graphicx}
\usepackage{subfig}
\usepackage{multirow}
\usepackage{tikz}
\usepackage{adjustbox}
\usepackage[printonlyused]{acronym}
\usepackage[noend]{algpseudocode}
\usepackage{algorithm}
\def\algbackskip{\hskip-\ALG@thistlm}

\newcommand{\Ereject}{E_\mathrm{reject}}
\newcommand{\jointPC}{\mathcal{P}_j}
\newcommand{\point}{\mathbf{p}}
\acrodef{FMCW}		[FMCW]		{Frequency-Modulated Continuous Wave}

\modulolinenumbers[5]

\journal{Robotics and Autonomous Systems}

\begin{document}

\begin{frontmatter}

\title{CorAl: Introspection for Robust Radar and Lidar Perception in Diverse Environments Using Differential Entropy}

\author{Daniel Adolfsson, Manuel Castellano-Quero, Martin Magnusson, Achim J. Lilienthal, Henrik Andreasson}%\fnref{myfootnote}}
\address{MRO lab of the AASS research centre, \"Orebro University, Sweden}
\fntext[myfootnote]{This work has received funding from the Swedish Knowledge Foundation (KKS) projects ``Semantic Robots'' and ``NiCE''
    and the European Union's
    Horizon~2020 research and innovation programme  under grant
    agreement No %732737 (ILIAD)     and 
    101017274 (DARKO).}
 %\thanks{This work has received funding from the Swedish Knowledge Foundation (KKS) projects ``Semantic Robots'' and ``NiCE'' and the European Union's Horizon~2020 research and innovation programme  under grant agreement No %732737 (ILIAD)     and  101017274 (DARKO).
    %\newline 978-1-6654-1213-1/21/\$31.00 \textcopyright 2021 IEEE}%
%\author[mymainaddress]{Elsevier Inc}

%\author[mymainaddress]{Global Customer Service\corref{mycorrespondingauthor}}



\end{frontmatter}

\linenumbers

\section{Introduction}
\label{sec:introduction}

For mobile robots to be truly resilient to possible failure causes, during long-term hands-off operation in difficult environments, robust perception needs to be addressed on several levels.
Robust perception depends on failure \emph{resilience} as well as failure \emph{awareness}.
Several stages of the perception pipeline are affected: from the sensory measurements themselves (sensors should generate reliable data also under difficult environmental conditions), via algorithms for registration, mapping, localization, etc, to \emph{introspection} (by which we mean self-assessment of the robot's performance).

This paper presents novel work on robust perception that addresses both ends of this spectrum, namely \emph{CorAl} (from ``Correctly Aligned?''); a method to introspectively measure and detect misalignments between pairs of point clouds.
%This paper presents novel work on robust perception that addresses both ends of this spectrum.
%The main contribution is \emph{CorAl} (from ``Correctly Aligned?''), which is a method to introspectively measure and detect misalignments between pairs of point clouds.
In particular, we show how it can be applied to range data both from lidar and radar scanners; which means that it is well suited for robust navigation in all-weather conditions.

Lidar sensing is, compared to visual sensors, inherently more unaffected by poor lighting conditions (darkness, shadows, strong sunlight).
Radar is furthermore unaffected by low visibility due to fog, dust, and smoke.
However, radar as a range sensor has rather different characteristics than lidar, and interpreting radar data (e.g., for localization) has been considered challenging due to high and environment-dependent noise levels, multi-path reflections, speckle noise, and receiver saturation.
In this work we demonstrate how radar data can be effectively filtered to produce high-quality point clouds; and how even small misalignments can be reliably detected.

% THIS DATA SHOULD BE INCLUDED INTHE ITNRODUCTION OR RELATED WORK
%Radar has the advantageous property of being largely unaffected by dust, weather types and adverse conditions. However, compared to lidar, radar measurements have a high level of noise including multi-path reflections, speckle noise and receiver saturation. Hence, interpreting radar data (e.g., for localization) is considered a challenging task.

Many perception tasks, including odometry estimation~\cite{9636253}, localization~\cite{adolfsson_submap_2019}, %scene understanding,
and sensor calibration~\cite{della_corte_unified_2019}, rely on point cloud registration.
However, registration sometimes provides incorrect estimates; e.g., due to local minima of the registration cost function~\cite{9013051}, uncompensated motion distortion~\cite{zhang_loam_2014},
rapid rotation (leading to poor initial pose estimates)~\cite{gcnv2}, %there might be better papers to cite than this
or when the registration problem is geometrically under-constrained~\cite{8462890,softconstraints}.

Consequently, it is essential to equip these methods with failure awareness by measuring alignment quality so that misaligned point clouds can be rejected or re-aligned.
While a number of measures of alignment quality already exist, it is typically not easy to set a threshold to detect poor alignments that can be used in different environments -- as we also demonstrate in our experimental validations (see \Cref{sec:lidar_classification,sec:radar_classification}).
A particular benefit of the CorAl method is that it generalizes well, so that if it has been trained in one environment, the same parameters can be used in other, unseen, environments. 

Some examples of methods that have been used in practice to assess the alignment quality include
point-to-point or point-to-plane distances~\cite{p2l_chen_Medioni,rusinkiewicz-2001-fasticp},
point-to-distribution~\cite{magnusson-2009-phd,Almqvist} or distribution-to-distribution~\cite{Stoyanov2012ijrr,fuzzybnb} likelihood estimates, mean map entropy~\cite{Droeschel14localmulti-resolution}
or dense radar-image comparison~\cite{barnes_masking_2020}.
%
% These metrics can typically be used to measure a relative alignment error in the process of registration, but provide limited information on whether the point clouds are correctly aligned once registration has been carried out~\cite{Bogoslavskyi2017AnalyzingTQ}. 
However, except for some notable exceptions~\cite{Almqvist,adolfsson-2021-coral}, few studies in the literature have specifically and methodically targeted the measurement of alignment correctness.

\begin{figure}
    \centering
    {\includegraphics[trim={1.5cm 0cm 0.0cm 0cm},clip,width=\linewidth]{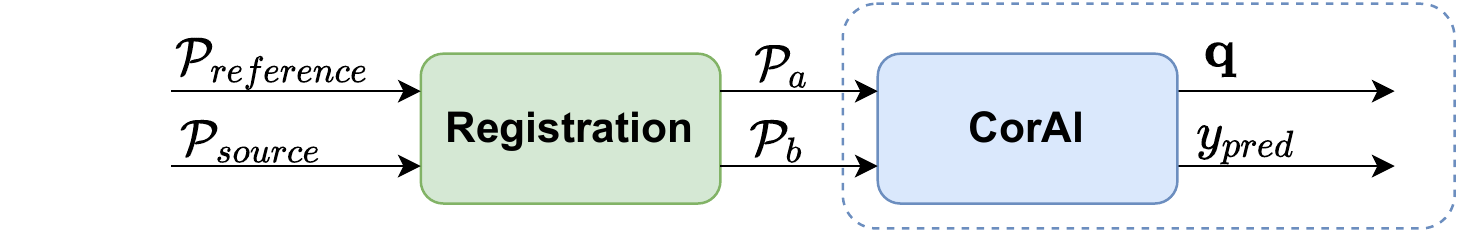}}
    \caption{CorAl, depicted in blue, operates on a pair of %previously registered 
    point clouds $\mathcal{P}_a,\mathcal{P}_b$ and can classify misalignment ($y_{pred}$) by comparing the differential entropy in the point clouds separately and jointly. Additionally, CorAl outputs a per-point quality measure $\mathbf{q}$ that highlights misaligned parts.\label{fig:coral_overview}\label{fig:my_label}}
\end{figure}

\begin{figure}
\vspace{-0.5cm}
\centering
\subfloat[][\raggedright  $\mathcal{P}_a$ colored by entropy.]{\includegraphics[trim={20.0cm 5cm 20.0cm 5cm},clip,width=0.49\linewidth]{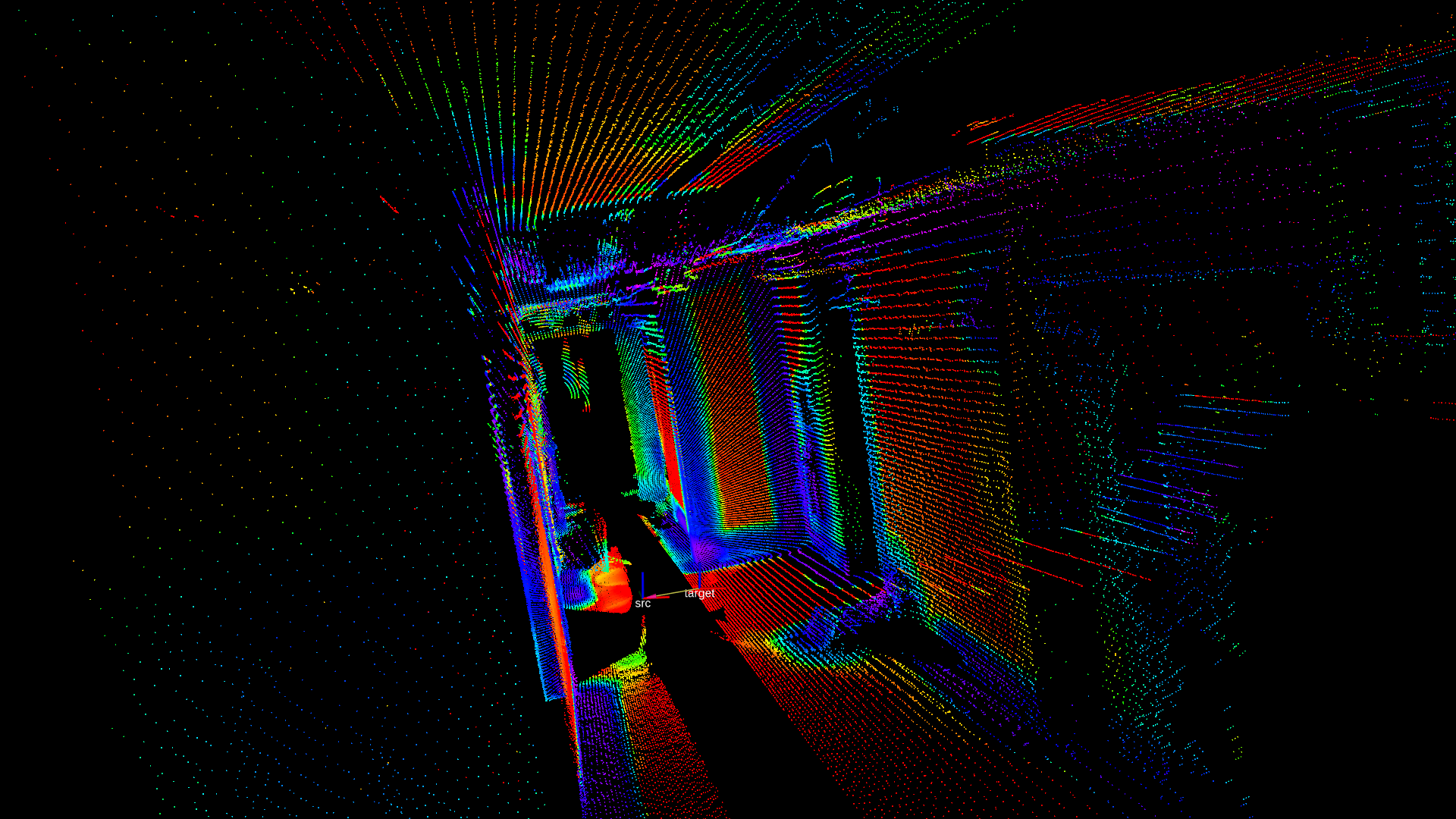}\label{fig:entropy_src}}
\subfloat[][\raggedright $\mathcal{P}_b$ colored by entropy.]{\includegraphics[trim={20.0cm 5cm 20.0cm 5cm},clip,width=0.49\linewidth]{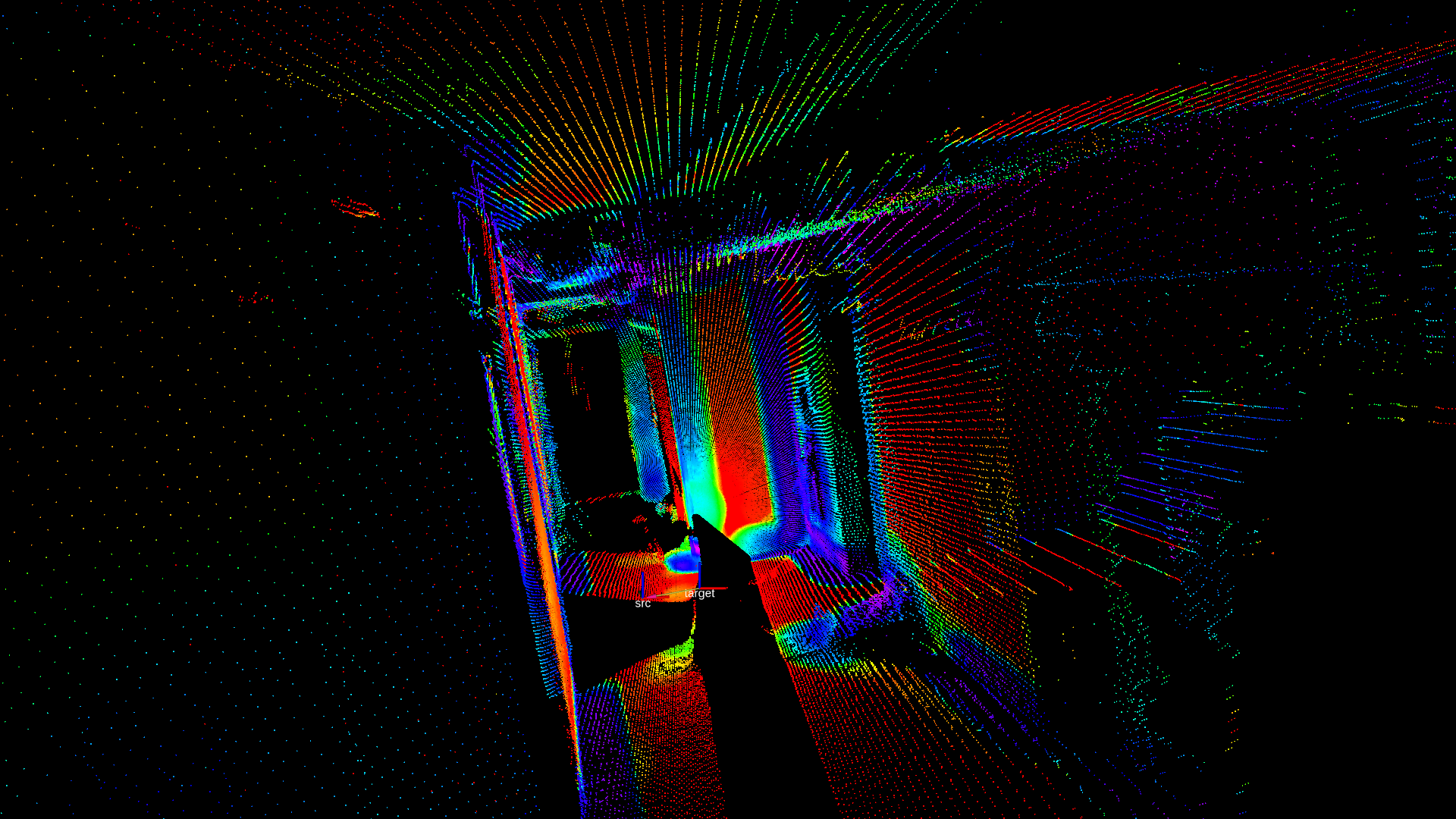}\label{fig:entropy_target}}
\\
\vspace{-0.2cm}
\subfloat[][\raggedright Correctly aligned $\mathcal{P}_a \cup \mathcal{P}_b$ colored by quality measure. In constrast to (a) and (b), red color indicate misalignment.]{\includegraphics[trim={20.0cm 5cm 20.0cm 5cm},clip,width=0.49\linewidth]{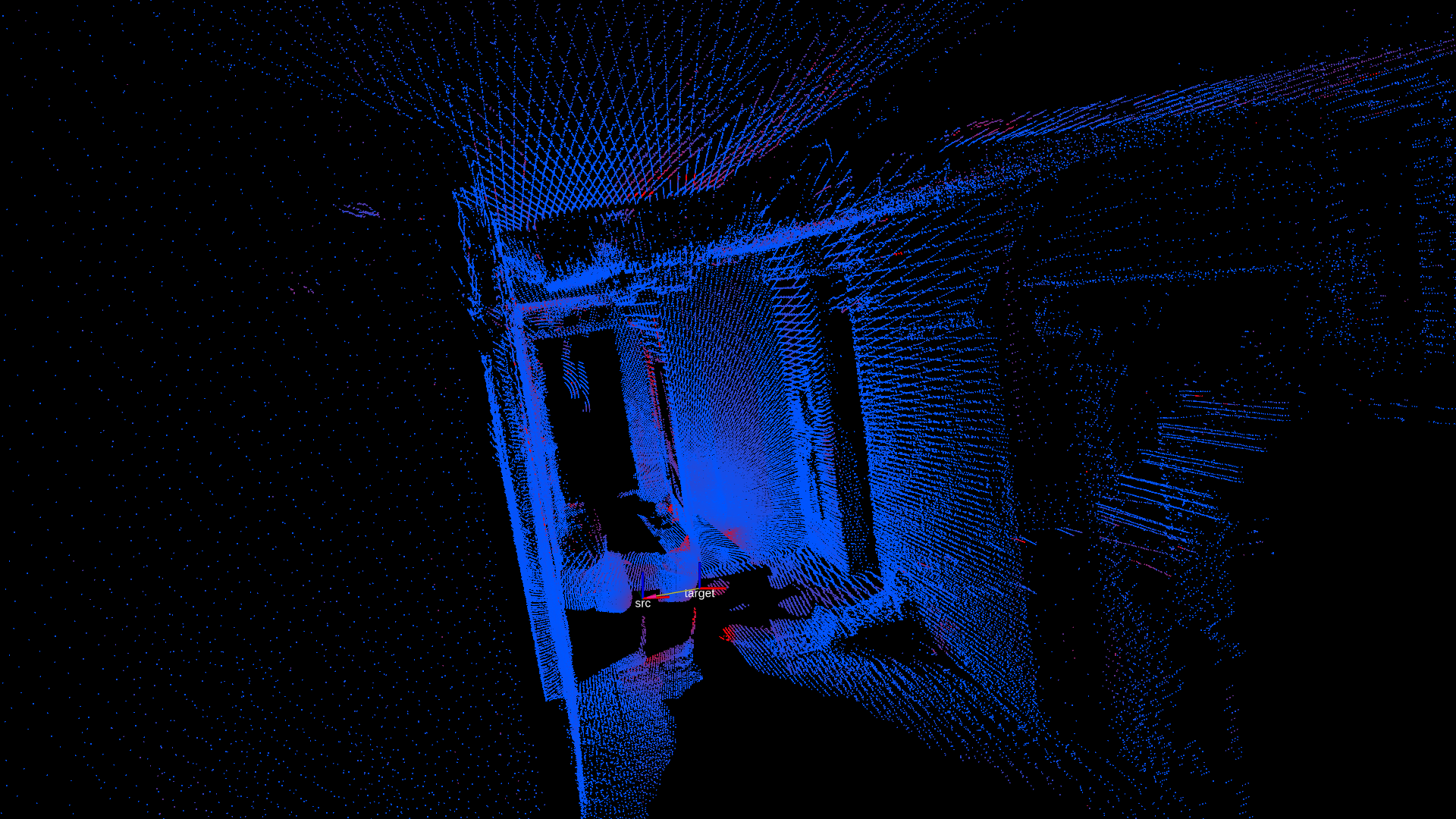}\label{fig:entropy_diff_aligned}}
\subfloat[][ Misaligned $\mathcal{P}_a \cup \mathcal{P}_b$ colored by quality measure when a small misalignment is added.]{\includegraphics[trim={20.0cm 5cm 20.0cm 5cm},clip,width=0.49\linewidth]{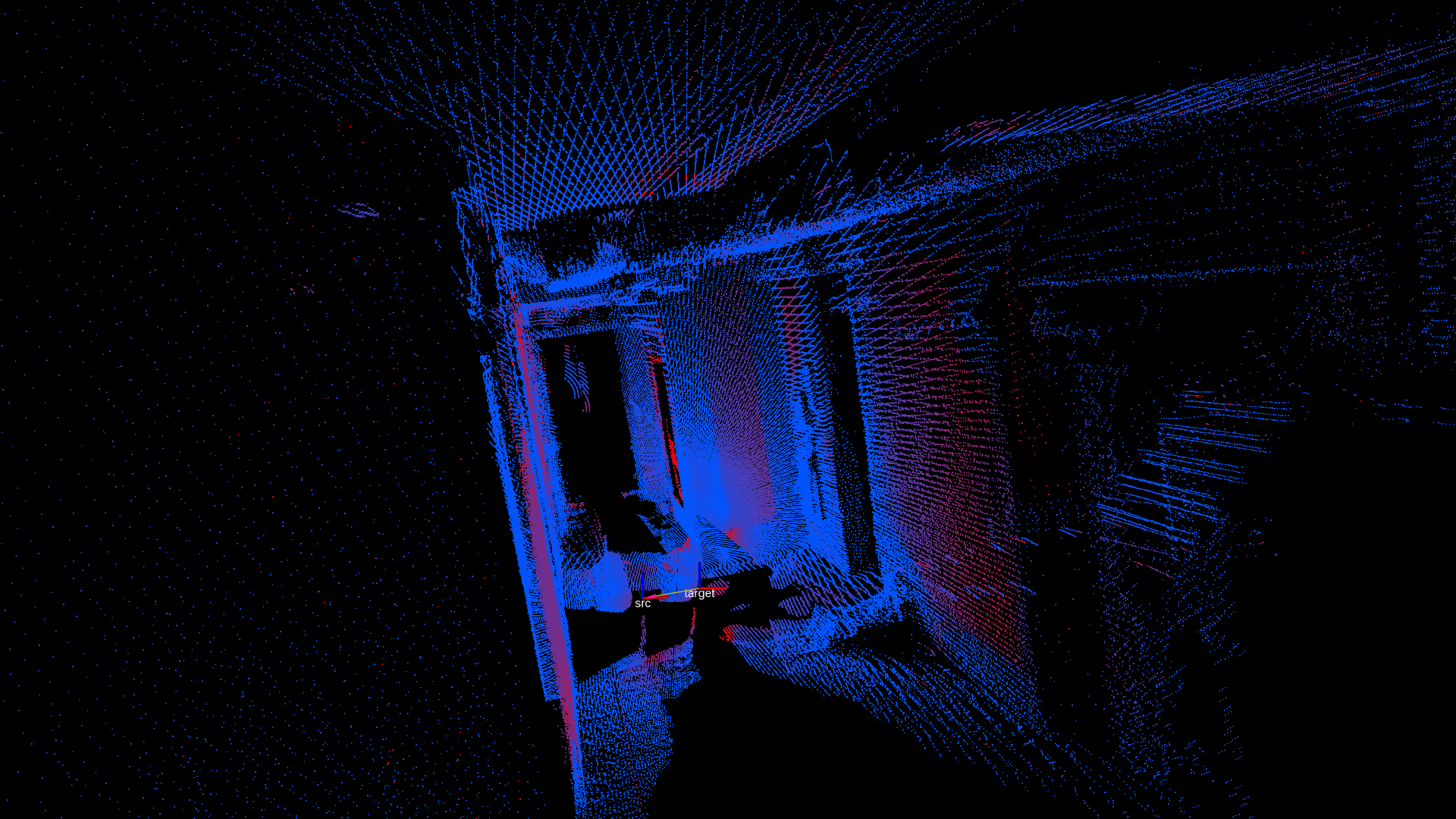}\label{fig:entropy_diff_misaligned}}
\caption{\label{fig:diff_entropy} \textit{Top:} Differential entropy in point clouds separately.
Colors range from red (low entropy) to blue and violet (high entropy).
\textit{Bottom:} The joint point cloud ($\mathcal{P}_a \cup \mathcal{P}_b$) colored by per-point quality measure $q_k(\mathcal{P}_a,\mathcal{P}_b)$ when aligned (c) and when misaligned (d). Blue and red indicate alignment and misalignment respectively.}
\vspace{-0.5cm}
\end{figure}

\begin{figure}
    \centering
    {\includegraphics[trim={0.5cm 7cm 3.7cm 1cm},clip,width=\linewidth]{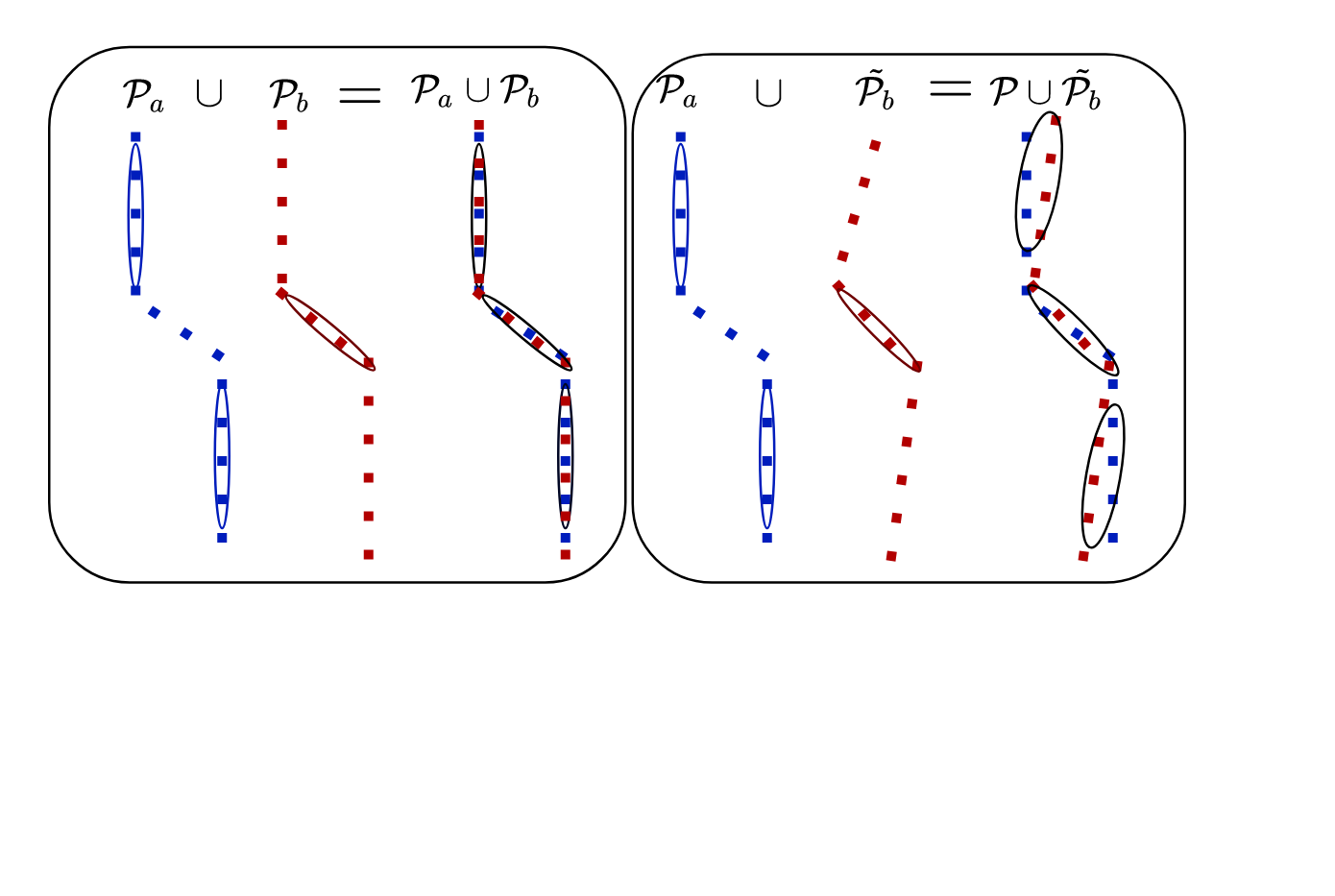}}
    \caption{Example how uncertainty (entropy) is preserved when joining aligned point clouds $\mathcal{P}_a\cup\mathcal{P}_b$ (left), but increases when joining misaligned point clouds (right). The entropy for aligned point clouds should be similar to the entropy in the separate point clouds and can be used when quantifying alignment quality.}
    \label{fig:illustration_coral}
    \vspace{-0.5cm}
\end{figure}

Our  method, CorAl, is well-grounded in information theory and gives an intuitive alignment correctness measure.
\Cref{fig:coral_overview} shows the general outline of the method, and \Cref{fig:diff_entropy} illustrates the output per-point alignment measure for a pair of point clouds.
CorAl computes the average differential entropy in two point clouds,  exploiting the difference when the entropy is computed for each point cloud separately, compared to the union of the point clouds.
For well-aligned point clouds, the joint and the separate point clouds will have similar entropy.
In contrast, misaligned point clouds tend to ``blur'' the scene, which can be measured as an increase in joint entropy as depicted in \Cref{fig:illustration_coral}.
A key idea is to estimate the entropy inherent in the scene from the entropy in the separate point clouds, which enables CorAl to accurately assess quality in a range of different environments.
In short, our previous contribution was an intuitive and simple measure of alignment correctness between point cloud pairs that generalizes across environments and highlights regions of misalignment.
%The method is described in detail in \Cref{sec:method}.
This paper extends our previous work on CorAl~\cite{adolfsson-2021-coral} 
by showing how it can be applied to FMCW (frequency-modulated continuous wave) radar data, thus pushing further towards truly robust perception, and includes quantitative evaluations on two new large-scale benchmarks and four new baselines.
In summarize, we present the following new contributions:

\begin{itemize} 
\item The first investigation (to our knowledge) that systematically evaluates alignment correctness classification using radar-based feature extraction and quality metrics without the aid of auxiliary sensors.
\item We include 4 new baselines based on recent research within spinning radar odometry.
\item A novel radar filtering strategy that allows CorAl to operate on noisy radar data, enabling alignment classification of small errors with high accuracy compared to previous radar feature extraction methods in urban traffic settings.
\item An ablation study that investigates parameter importance and how practical factors such as error magnitude and variation in distance between scans influence classification performance.
\item A cross-environment study that demonstrates that CorAl generalizes to new environments without retraining. 
\end{itemize}

\section{Related work}

\subsection{Cost functions for scan registration}
In practice, it is common to use the cost function of a scan registration method to estimate the  alignment quality after registration. 

One well-used alignment measure is the root-mean-squared (RMS) point-to-point distance, truncated by some outlier rejection threshold.  
This is also the function that is minimized by iterative closest point registration~\cite{icp,rusinkiewicz-2001-fasticp}.
However, this measure has been shown to be highly sensitive to the environment and the choice of the outlier threshold \cite{silva-2005-ga,Almqvist} when trying to detect small errors. Consequently, this is a poor measure for alignment correctness classification.

Another common alternative is the point-to-line distance~\cite{p2l_chen_Medioni,rusinkiewicz-2001-fasticp} or -- as a generalization -- a point-to-distribution~\cite{magnusson-2009-phd,Segal_2009} or distribution-to-distribution~\cite{Stoyanov2012ijrr} measure; where local surface descriptors are computed using the spatial distribution of points within a neighborhood.
In a similar vein, Liao et al.~\cite{fuzzybnb} propose distribution-to-distribution registration based on fuzzy clusters, and estimate coarse alignment quality via the dispersion and disposition of points around fuzzy cluster centers.
These methods have also been shown to generalize poorly for assessing 3D lidar scan alignment in different environments~\cite{adolfsson-2021-coral}.

%%%%%%%%%%%%%%%%%%%%%%%%%%%5
% [X] p2l_chen_Medioni,
% [X] rusinkiewicz-2001-fasticp}, 
% [X] point-to-distribution~\cite{magnusson-2009-phd,Almqvist} or
% [X] distribution-to-distribution~\cite{Stoyanov2012ijrr,
% [X] fuzzybnb} likelihood estimates, 

% [X] Mean Map Entropy~\cite{Droeschel14localmulti-resolution} or 
% [X] dense radar-image comparison~\cite{barnes_masking_2020}.

% [X] Bogoslavskyi et al.~\cite{Bogoslavskyi2017AnalyzingTQ} alignment quality pos+neg info

% [X] \cite{chandran-ramesh_assessing_2007} Assessing Map Quality using Conditional Random Fields Chandran-Ramesh

% [X] \cite{Makadia2006FullyAR} Makadia2006FullyAR Fully Automatic Registration of 3D Point Clouds - quality by consistency of normals.

%%%%%%%%%%%%%%%%%%%%%%%%%%%%%%%%%%%%

\subsection{Fault detection and uncertainty estimation}
Going beyond merely setting a threshold for the registration cost function, there are also some methods that have been devised specifically for fault detection and uncertainty estimation. 

%Similarly, 
Huan et al.~\cite{8996777} presented a failure detection based on logistic regression and using point cloud overlap, differences between 2d point cloud projections and mean and deviation of nearest point distance and normal directions. They show that various metrics can be combined for increased accuracy, however the work does not explicitly focus on detecting small errors and instead includes errors up to 5~m. 

Makadia et al.~\cite{Makadia2006FullyAR} use consistency between normals (``plane-to-plane``) as a post-registration alignment measure, where normals are computed from voxelized versions of two point clouds.
In an independent evaluation~\cite{Almqvist}, this method was found to perform poorly in unstructured outdoor environments.

%Fault detection in localization systems has also been investigated in the following papers. %~\cite{8462890,8500625,8917111,1642280}.
%
Nobili et al.~\cite{8462890} proposed a method to predict alignment risk prior to registration by combining overlap information and an alignment metric that quantifies the geometric constraints in the registration problem. Their alignment metric is based on point-to-plane residuals and has been evaluated in structured scenes with planar surfaces. In contrast, the method we propose in this paper can operate well even in unstructured environments. Additionally, our method seeks to estimate the alignment after registration has been completed to introspectively measure the registration success, as opposed to predicting the risk prior to registration.

Akai et al.~\cite{8500625} estimate reliability of vehicle localization from grid map and laser scan data, using a convolutional neural network (CNN). In our work, we are interested in alignment classification without prior knowledge for pairs of point clouds. Aldera et al.~\cite{8917111} learn detection of odometry failures in challenging conditions using weak supervision from  IMU or GPS. Their method analyzes eigenvectors of a pairwise compatibility matrix which contains scores between point correspondences. %What is the critique.
Finally, Sundvall and Jensfelt~\cite{1642280} propose a method for  fault detection with redundant positioning systems. In contrast, our method operates on lidar or radar without the need for any additional sensors during deployment and uses odometry or optionally ground truth during training.

A family of methods assessing alignment uncertainty based on an estimate of the pose covariance can be found in~\cite{David_Landry,bengtsson_robot_2003, nieto_scan-slam_2006,prakhya_closed-form_2015,censi-2007-accurate,magnusson-2009-phd,5152375}. Some of these use a Monte Carlo method and estimate uncertainty by sampling registrations in a region~\cite{bengtsson_robot_2003}. For mobile robotics, sampling strategies are tedious and unpractical. Other methods compute covariance in closed form based on the Hessian of the quality metric~\cite{censi-2007-accurate,magnusson-2009-phd} or from pose samples  weighted by dense correlation~\cite{5152375,barnes_masking_2020}.
%Almqvist et al.~\cite{Almqvist} include a covariance measure (NDT Hessian) in their evaluation which enables similar or less accuracy compared to RMS.
However, it is not generally possible to define a fixed set of covariance thresholds to distinguish good from bad alignments.

Bogoslavskyi and Stachniss~\cite{Bogoslavskyi2017AnalyzingTQ} define a quality metric that  takes into account free-space information, and use it to measure alignment error between range images of segmented 3D objects in a controlled experiment. Rather than focusing on objects, our method aims to classify the alignment quality of observed scenes in different environments. Additionally, their method operates on range images, which might not be easily available, while our method operates on unorganized point clouds.

Some methods work on the scale of a full map, rather than individual scans.
Chandran-Ramesh and Newman~\cite{chandran-ramesh_assessing_2007} convert a point cloud map into plane patches and train a conditional random field to detect plausible and suspicious plane configurations.
This method is not directly applicable for assessing pairwise point cloud alignment.
Droeschel and Behnke~\cite{Droeschel14localmulti-resolution} compute mean map entropy to measure the ``crispness'' of a point cloud map, with the aim to evaluate accuracy of scan registration in lieu of ground truth pose data.
Our work is inspired by this measure but, as further detailed and demonstrated below, mean map entropy does not generalize between structured and semi-structured environments.

\subsection{Feature extraction and quality assessment for spinning radar}
Lidar is a well-investigated sensor modality in robot perception.
Spinning FMCW radar is an alternative modality that has been receiving more attention in recent years, due to it being resilient to low-visibility conditions.
However, due to its challenging noise characteristics, how to efficiently interpret the data for robot perception is still considered an open research question. 
Hence, in our investigation of radar we include both feature extraction and quality assessment. Radar-based methods that use alignment quality measures can be categorized into dense methods~\cite{barnes_masking_2020,Fourier_Mellin,9197231}, which operate on raw radar images and do not explicitly perform data association, and sparse methods~\cite{hong2021radar,hong2020radarslam,burnett_we_2021,barnes_under_2020,burnett2021radar,8460687,8793990,kung2021normal,9636253}, which compute alignment quality using keypoint locations, shape and descriptors over a correspondence set. Previous sparse methods use (weighted) Point-to-Point~\cite{barnes_under_2020,8460687,8793990}, Point-to-distribution~\cite{kung2021normal} and Point-to-Line~\cite{9636253} metrics. Key points can be extracted via SURF, blob detection~\cite{hong2021radar}, gradient-based
feature detectors~\cite{8460687,8793990}, by a set of  oriented surface points~\cite{9636253} or distributions~\cite{kung2021normal} using a grid-based approach, or by semi-supervised~\cite{barnes_under_2020} and unsupervised~\cite{barnes_under_2020,burnett2021radar} deep learning methods. 

While these methods for feature extraction and alignment quality have been used as objective functions for the purpose of estimating odometry, it is currently not known to which extent these metrics can be used for alignment correctness classification.
%Hence, we argue that existing learning-free feature extractors are able to compute consistent features in presence from challenging radar data.
Previous work shows that the performance of deep learning-based odometry methods decreases in new environment types~\cite{barnes_masking_2020,burnett_we_2021}. However, some methods have been successful in estimating odometry and performing SLAM in widely different environments without deep learning and even without parameter tuning~\cite{9636253,hong2021radar}. In this paper, we show that learning-free feature extraction methods can produce stable key points over multiple environment types, and are suitable for alignment correctness classification in diverse environments. 
%Hence we refrain from learning at the earlier steps to extracting features, but utilize learning first at a classification stage to find linear decision boundaries based on logistic regression.
Hence, we can forego (deep) learning for feature extraction and only use logistic regression at the classification stage to learn linear decision boundaries, which ultimately only requires two parameters to be trained.

\subsection{Comparative studies of alignment assessment}
%%% THOWEVERHESE ARE USED IN AD HOC MANNER, FEW METHODS USES SYSTEMATIC EVALUATION
Most of the methods above have been  used in a more or less ad-hoc manner. 
Few systematic evaluations and comparative studies have been made on investigating their general capability for the task of alignment classification between point clouds pairs, i.e. to detect aligned vs. misaligned ones.
Two evaluations in this direction were presented by Almvqvist et al.~\cite{Almqvist} and in on our previous work CorAl~\cite{adolfsson-2021-coral} where a range of quality metrics was used to train classifiers using logistic regression. 
Almqvist et al. explored alignment classifiers based on point-to-point distances as well as a number of other  methods~\cite{rusinkiewicz-2001-fasticp,biber_normal_2003,magnusson-2009-phd,chandran-ramesh_assessing_2007,makadia-2006-fully,silva-2005-ga}, and investigated how to combine the measures with AdaBoost into a stronger classifier.
The classifiers were evaluated on two outdoor data sets, and although the best ones reached almost 90\,\%  accuracy for the hardest cases on each data set individually, accuracy drops to around 80\,\% when cross-evaluating between the data sets.
In their evaluations, the NDT score function~\cite{magnusson-2009-phd} proved to be the best individual measure for alignment assessment. The combined AdaBoost classifier did not have significantly higher accuracy, but reduced parameter sensitivity.  Our previous work confirms the finding by Almqvist: that detecting small alignment errors in a diverse range of environments without retraining is challenging.

%%%%% GLOBAL ALGORITHMS AND DATA ASSOCIATION
%Leordeanu correspondance 1544893
%Yang teaser rejection ~\cite{Yang20tro-teaser}
%Lusk clipper ~\cite{lusk2020clipper}
%% Alignment quality in Radar

\section{CorAl method}
\label{sec:method}

Our work takes inspiration from Droeschel and Behnke~\cite{Droeschel14localmulti-resolution} who used differential entropy measurements to assess map quality. 
The differential entropy measures the uncertainty or surprisal of a continuous variable, in their case for (3-dimensional) variables $X\in \mathbb{R}^3$ with multivariate Gaussian distribution $X \sim \mathcal{N}_3(\mathbf{\mu},\,\mathbf{\Sigma})$. Distributions are approximated by the sample mean and covariance of local point distribution within a radius $r$.
%\mathbf{x}\sim
In their work, they compute the ``Mean-Map-Entropy'' (MME), defined as the average per-point differential entropy over a set of point clouds, and use the measure to compare methods for lidar map refinement algorithms without the need for ground truth measurements and without imposing planar assumptions. As the entropies measure surprisal, everything else being equal, a lower MME measure can be interpreted as \textit{less surprisal} or \textit{crisper point clouds} and indicate the success of map refinement.
While MME is suitable for and has been used to quantify relative alignment improvement~\cite{Razlaw2015EvaluationOR,8461000}, the metric additionally depends on sensor noise and sample density and is highly dependent on scene geometry. Consequently, and as confirmed by our evaluation, the metric does not generalize over, e.g. structured and semi-structured environments. For that reason, we overcome sensitivity to variation in scene geometry by making use of dual entropy measurements computed 1) in point clouds separately and 2) in the joint point cloud. Intuitively, when joining a well-aligned point cloud pair, the entropy (or the blur) found in the joint point cloud should not increase compared to the entropy found in the separate point clouds, and instead, remain close to constant. In contrast, when joining point clouds with small misalignments, the joint entropy (blur) tends to increase compared to the separate entropy. By making use of dual differential entropy measurements, our method can account for the scene appearance and detect small alignment errors. Additionally, the measurements enable generalization to substantially different environments without retraining.

%Our work is inspired by the Mean-Map-Entropy (MME) measure proposed by Droeschel and Behnke~\cite{8461000} for map quality assessment. MME is based on differential entropy~\cite{article} and measures the randomness of multivariate Gaussian distributions. Droeschel and Behnke  used  MME in absence of accurate ground truth when evaluating map refinement.
%As shown in our evaluation, MME cannot be used as a general alignment quality measure as it is also affected by measurement noise, sample density and environment geometry. MME is more affected by changes in the environment compared to CorAl. Hence, the measure is not expected to generalize between, e.g., a structured warehouse and an unstructured outdoor forest environment. We overcome this effect using dual entropy measurements computed 1) in both point clouds separately and 2) in the joint point cloud. The intuition is that joining two well-aligned point clouds should not introduce additional uncertainty and entropy should remain constant if the point clouds overlap sufficiently.
\subsection{Assumptions and definitions}
CorAl operates on 2D or 3D point cloud pairs $\mathcal{P}_a$, $\mathcal{P}_b$ acquired by a range sensor such as lidar or radar. Points $\bold{p}_k\in \mathcal{P}$ are given in a common fixed world frame in $\bold{p}_k\in \mathbb{R}^2$ or $\bold{p}_k\in \mathbb{R}^3$. 
CorAl learns a linear decision boundary to separate aligned vs misaligned point-cloud pairs based on features extracted without learning.
During the learning phase, point clouds are assumed to be correctly aligned by an accurate ground truth or odometry system and free of distortions from motion. We aim to detect small alignment errors. Detecting large errors e.g. $>1$~m in large-scale-urban or larger than $>0.4$~m within indoor environments is generally not considered challenging for traditional metrics such as point-to-point or point-to-line metrics, thus, it is not the focus of our work.
For notation, we define the joint point cloud $\jointPC=\mathcal{P}_a \cup \mathcal{P}_b$; i.e., all points in $\mathcal{P}_a$ and  $\mathcal{P}_b$ together.

\subsection{Joint and separate entropy measurements}
\label{sec:coral-entropy}
%The differential entropy is computed around a single 
In order to compute the differential entropy around a single point, we first compute the sample covariance $\bold{\Sigma(p}_k)$ from all points within a radius $r$ around $\bold{p}_k$.
%are used to compute the sample covariance. Second,
%The per-point differential entropy is computed by 
%From all points within a radius $r$ around each point $\bold{p}_k$, we compute the sample covariance $\bold{\Sigma(p}_k)$. 
The differential entropy can then be computed from the determinant of the sample covariance according to:
%From the determinant of the sample covariance $\det(\bold{\Sigma}(\bold{p}_k))$  the diff:
\begin{equation}
    \label{eqn:point_entropy}
    h_{i}(\bold{p}_k)=\frac{1}{2}\ln((2\pi e)^N\det(\Sigma(\bold{p}_k))),
\end{equation}
%for the point cloud $i=a,b$ that contains $\bold{p}_k$. 
for a multivariate normal distribution with dimension $N$.
The property of this metric can be visually understood from Fig.~\ref{fig:entropy_src} where each point is colored according to the differential entropy. The metric serves as a geometric descriptor that describes the local geometry with a single value; the differential entropy.
% \cref{eqn:point_entropy}. 

\begin{figure}
\centering
\subfloat[Top view of the first two aligned point clouds $\mathcal{P}_a$(blue) and $\mathcal{P}_b$(red) in the ETH ``stairs'' dataset. 
]{\includegraphics[trim={0cm 4cm 7.cm 6cm},clip,width=\linewidth,angle=0]{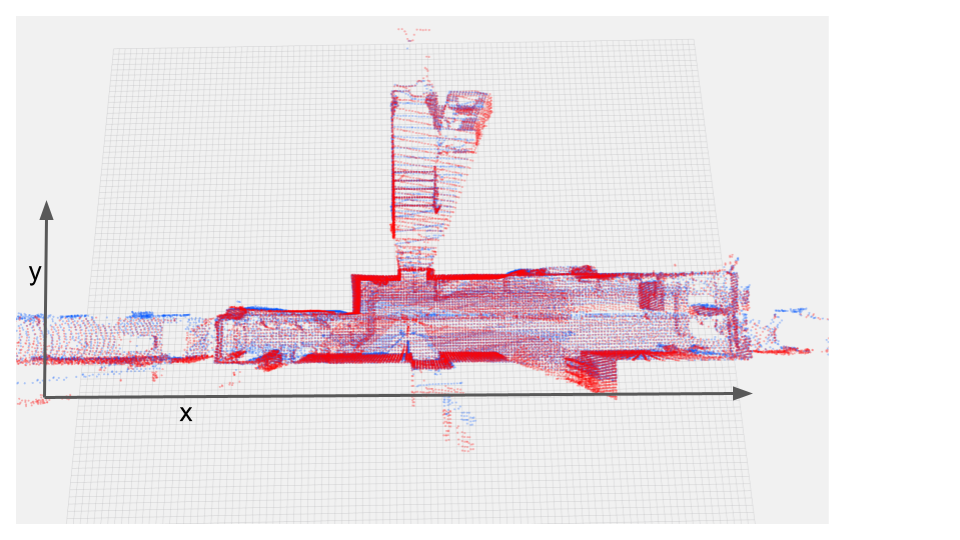}%
\label{fig:top_view_stais}}\\
\vspace{-0.2cm}
\subfloat[][\raggedright CorAl measure $Q(\mathcal{P}_a,\mathcal{P}_b)$ \cref{eqn:EntropyDifference} visualized by color for various (x,y) displacements.]{\includegraphics[trim={0.3cm 0cm 8.5cm 0cm},clip,width=0.8\linewidth]{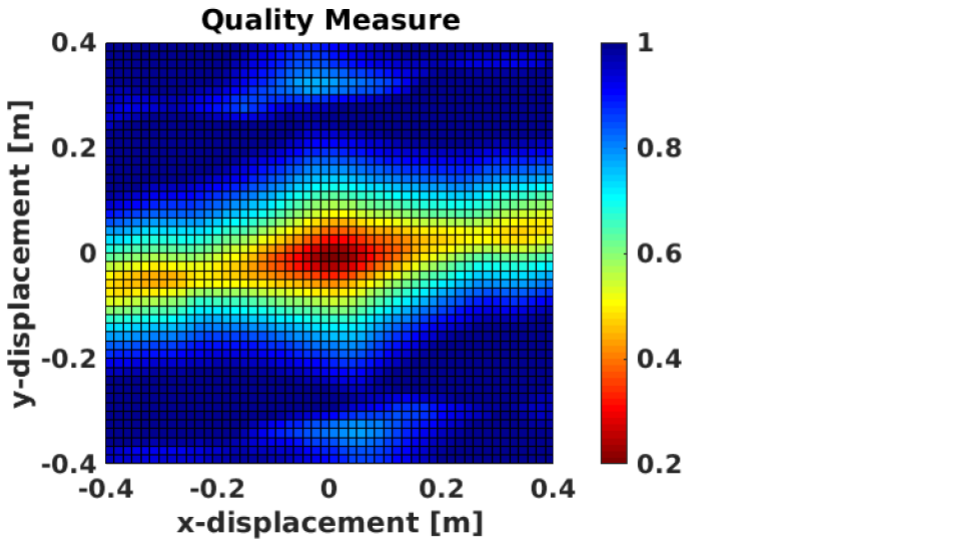}\label{fig:quality_xy_plot}}
%\subfloat[][\raggedright CorAl measure $Q(\mathcal{P}_a,\mathcal{P}_b)$ for various (x,$\theta$) displacements.]{\includegraphics[trim={0.0cm 0cm 0.5cm 0cm},clip,width=0.49\linewidth]{Journal/figures/ETH/objective/narrow_angle_objective.png}\label{fig:quality_xtheta_plot}}
\caption{
%Example of CorAl measure $Q(\mathcal{P}_a,\mathcal{P}_b)$ for various induced (x,y,$\theta$) alignment errors.
$Q$ has a minimum at the true position. The steepness of the surface around the true alignment indicates that CorAl is sensitive to small misalignments.
\label{fig:qualityMeasurePlot}}
\vspace{-0.5cm}
\end{figure}

The MME can then be computed by averaging Eq.~\ref{eqn:point_entropy} over the joint point cloud $\mathcal{P}_j$; i.e., computing the sum and then dividing by the number of terms as shown in Eq.~\ref{eqn:cloud_etntropy} and Eq.~\ref{eqn:Hjoint},
 
%sThe sum of differential entropy for $\mathcal{P}_i$ can then be computed as % in eq.\ref{eqn:cloud_etntropy}:
\begin{equation}
    \label{eqn:cloud_etntropy}
    H_i(\mathcal{P}_i)=\sum_{k=1}^{|\mathcal{P}_i|}h_i(\bold{p}_k)
    ,
\end{equation}
\begin{equation}
    \label{eqn:Hjoint}
    H_{\mathrm{joint}}=(MME)=\frac{H_j(\mathcal{P}_j)}{|\mathcal{P}_j|}=\frac{H(\mathcal{P}_a\cup\mathcal{P}_b)}{|\mathcal{P}_a|+|\mathcal{P}_b|}
    ,
\end{equation}
where $|\mathcal{P}_i|$ is the number of points in the point cloud $\mathcal{P}_i$. 
We extend this formulation by additionally computing the entropy in $\mathcal{P}_a, \mathcal{P}_b$ separately:
%Using Eq.~\ref{eqn:cloud_etntropy} we can derive measures of the \emph{separate} and \emph{joint} average differential entropy of two point clouds .
\begin{equation}
    \label{eq:Hseparate}
    H_{\mathrm{sep}}=\frac{H_{a}(\mathcal{P}_a)+H_{b}(\mathcal{P}_b)}{|\mathcal{P}_a|+|\mathcal{P}_b|},
\end{equation}
and obtain our quality metric by subtracting the average differential entropy from the joint:

\begin{equation}
    \label{eqn:EntropyDifference}
    Q(\mathcal{P}_a,\mathcal{P}_b)=H_{\mathrm{joint}}(\mathcal{P}_{j})-H_{\mathrm{sep}}(\mathcal{P}_a,\mathcal{P}_b)
    .
\end{equation}
The CorAl quality metric is hence ~\textit{the difference in differential entropy}. The CorAl quality can also be given on a per-point level by
\begin{equation}
    \label{eqn:point_quality}
    q_k(\bold{p}_k)=h_{j}(\bold{p}_k)-h_{i}(\bold{p}_k)
    ,
\end{equation}
where $i$ is the point cloud ($a$ or $b$) $\bold{p}_k$ originates from.
%where the point entropy is evaluated on the joint point cloud $j$ and the separate point cloud $i$ =   where  $\bold{p}_k$ originates from.
While the differential entropy in Eq.~\ref{eqn:point_entropy} varies according to the shape of the environment, the per-point entropy difference in Eq.~\ref{eqn:point_quality} has the property of being close to zero when point clouds are well aligned and increase with misalignment as depicted in Fig.~\ref{fig:entropy_diff_aligned}.
%according to ~\cref{eqn:point_quality} is depicted in Fig.~\ref{fig:entropy_diff_aligned} and \ref{fig:entropy_diff_misaligned}. 
The corresponding function surface of the quality measure $Q(\mathcal{P}_a,\mathcal{P}_b)$ is depicted in Fig. ~\ref{fig:qualityMeasurePlot}, which demonstrates how the score increases when displacing one of the point clouds in translation and orientation. 
%Typically, $Q(\mathcal{P}_a,\mathcal{P}_b)$ is close to zero for well-aligned point clouds and increases with the alignment error as depicted in ,
%which visualizes the function's surface for position and angular alignment errors around the correct alignment.

Well-aligned point clouds  $\mathcal{P}_a\cup\mathcal{P}_b$ acquired in structured environments have low differential entropy for most query points $\bold{p}_k$. This is reflected by low values for the determinant of the sample covariance. As the determinant can be expressed as the product of the eigenvalues of the sample covariance $\det(\bold{\Sigma(\bold{p}}_k))=\lambda_1\lambda_2\lambda_3$, we see that the measure is sensitive to an increase in the lowest of the eigenvalues when larger eigenvalues are constant. For example, the entropy of points on a planar surface is represented with a flat distribution with two large ($\lambda_1,\lambda_2$) and one small ($\lambda_3$) eigenvalue. Misalignment changes the point distribution in the joint point cloud from flat to ellipsoidal, which can be observed as an increase of the smallest eigenvalue $\lambda_3$. This makes the measure sensitive to misalignment of planar surfaces while generalizing well to other geometries. As shown in the evaluation, the CorAl measure can capture discrepancies between point clouds regardless of whether these are due to rigid misalignments or distortions that can occur when scanning while moving (e.g., because of vibrations or sensor velocity estimation errors). That means that the method may also be overly sensitive when used together with a registration method or odometry framework that does not compensate movement distortion or has a low accuracy.

\begin{figure}
\centering
\subfloat[Aligned point clouds: differential entropy distributions are similar.]{\includegraphics[clip,trim={2.0cm 0cm 2cm 0cm},width=0.49\linewidth]{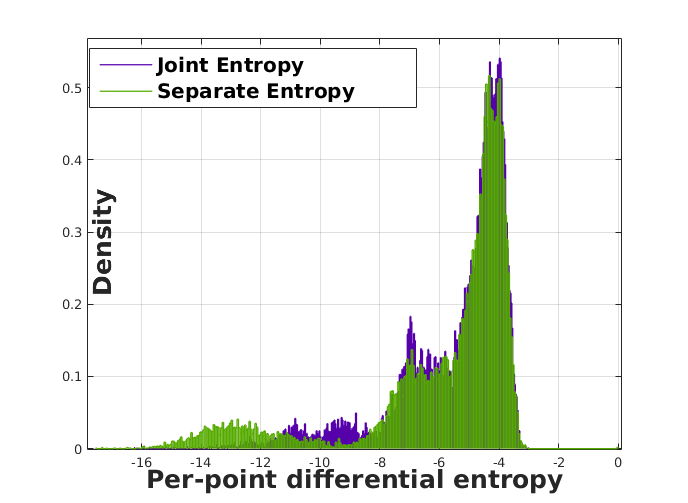}\label{fig:entropy_aligned}}\hfill%
\subfloat[Misaligned point clouds: joint differential entropy is higher.
]{\includegraphics[clip,trim={1.8cm 0cm 1.8cm 0cm},width=0.49\linewidth]{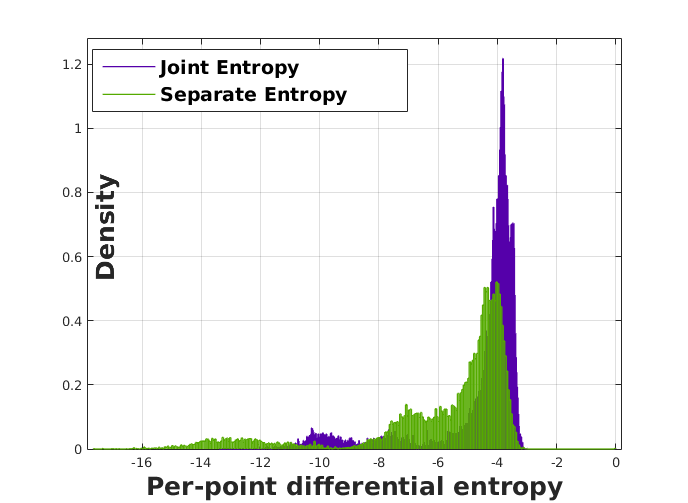}\label{fig:entropy_misaligned}}
\caption{\label{fig:entropypdf}
Probability distribution of per-point entropy \Cref{eqn:point_entropy} for joint and separate point clouds when aligned (a) and misaligned (b). Aligned point clouds have similar joint and separate entropy distributions. For misaligned point cloud pairs, the distribution of joint entropy is shifted compared to the separate entropy distribution. Joining misaligned point clouds tend to blur the scene, which can be observed by an entropy increase.
An exception can be seen in (a), within region $(-12,-8)$, where entropy mistakenly increases when joining aligned point clouds as described in ~\ref{sec:coral-modifications}.
}

\vspace{-0.3cm}
\end{figure}

%Overlap is required between point clouds to produce evidence of alignment. For that reason, we classify point clouds with less than 10\% overlap as misaligned. By defining the overlap as all points with a neighbour within $r$ in the other point cloud, non overlapping points have no effect on the quality measure in Eq. \ref{eqn:EntropyDifference}.

\subsection{Dynamic radius selection and outlier rejection }
\label{sec:coral-modifications}
For well-aligned point clouds, the quality measure $Q$ has values close to zero. In this case, the distribution of per-point differential entropies in the joint and separate point clouds are similar. The per-point entropy distributions are depicted in Fig.~\ref{fig:entropy_aligned} for a set of aligned and misaligned point cloud pairs when using fixed radius for computing entropies.
%meaning that the joint and separate point clouds have similar distributions of per-point entropy as depicted in %Fig.~\ref{fig:entropy_aligned}.
Unfortunately, the entropy in Eq.~\ref{eqn:point_entropy} is numerically unstable when the determinant of the covariance 
%in Eq.~\cite{eq:}
$\det(\bold{\Sigma}(\bold{p}_k))$ is ill-conditioned, hence a small increase of the determinant causes a large increase of the entropy. Accordingly, the lowest measured entropies can increase a lot (which mistakenly indicates misalignment) even when joining well-aligned point clouds as depicted in Fig.~\ref{fig:entropy_aligned}. 

One case where entropy mistakenly increases is when a 3d lidar observes floor regions with sparse ring patterns (due to space between vertical beams of the sensor). In the separate point clouds, the sparse rings give rise to ellipsoidal covariances (low entropy) with high uncertainty along the ring direction and low in the other directions. In the joint point cloud, the sparsity between rings is reduced, the computed covariances are instead planar according to the floor plane, with two high eigenvalues and one low, and therefore have a higher entropy. 

Ill-conditioned covariances occur where point density is low, typically for solitary points or far from the sensor where the radius $r$ is not large enough, e.g., to include multiple ``lidar rings'' within one entropy measurement, as in the example described above.

The ill-conditioned entropies can be mitigated by increasing the radius $r$ or making use of the options described below. Obviously, for a set of point cloud pairs, the parameters are typically well-calibrated
(and ill-conditioned entropies have been mitigated) if CorAl can separate between aligned and misaligned point clouds. This occurs when the joint $H_j$ and separate $H_{sep}$ entropies are linearly separable. The parameter can be calibrated by maximizing the ratio:

\begin{equation}\label{eq:Qs}
Q_s=Q_{misaligned}(\mathcal{P}_a,\mathcal{P}_b)/Q_{aligned}(\mathcal{P}_a,\mathcal{P}_b).
\end{equation} A large ratio indicates that the measure is able to discriminate between aligned and misaligned point clouds.
\begin{figure*}[htb!]
  \begin{center}
    \subfloat[]{\includegraphics[trim={0.0cm 2cm 0cm 0cm},clip,width=0.249\linewidth]{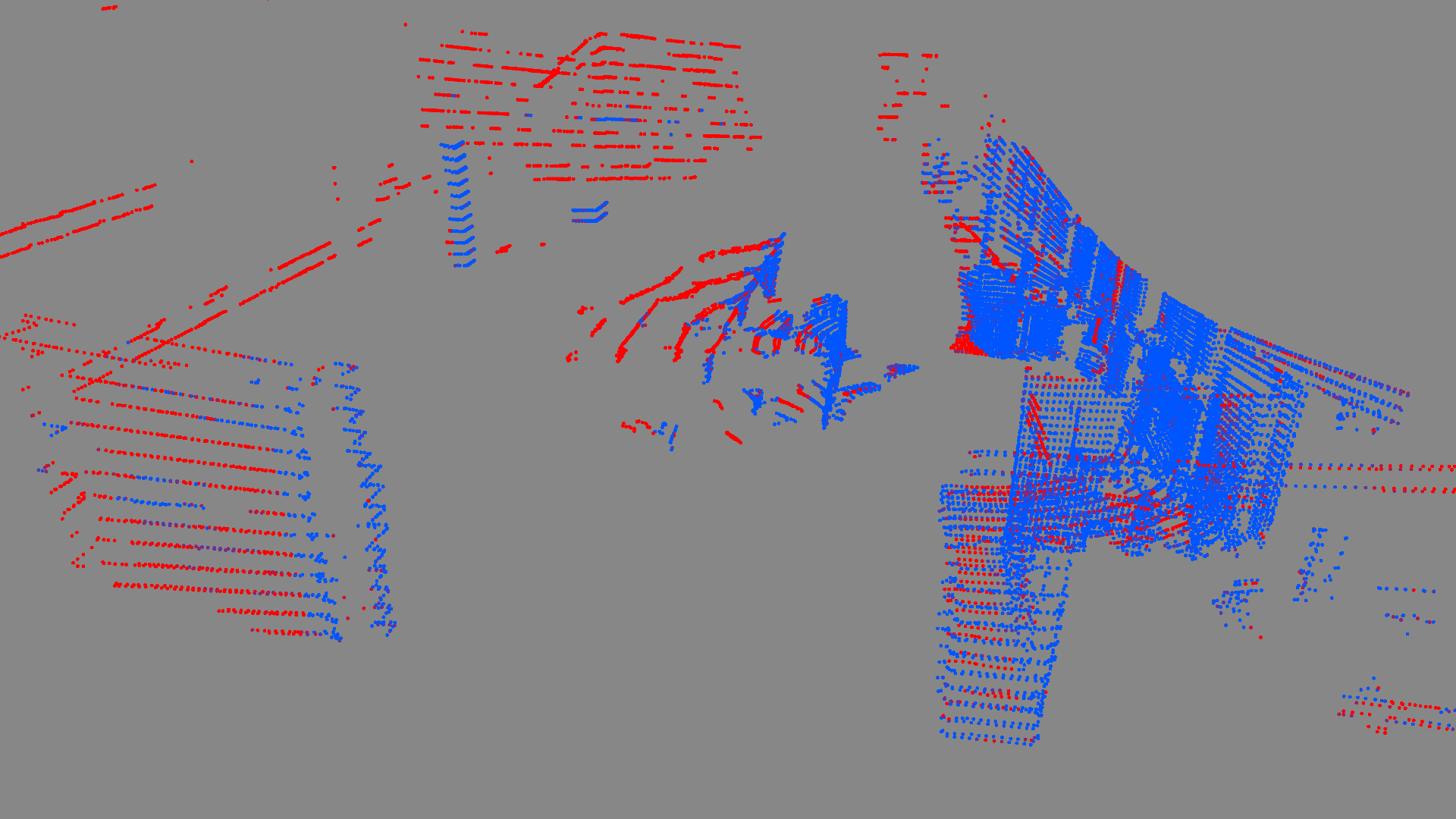}\label{fig:settings_a}}\hfill
    \subfloat[]{\includegraphics[trim={0.0cm 2cm 0cm 0cm},clip,width=0.249\linewidth]{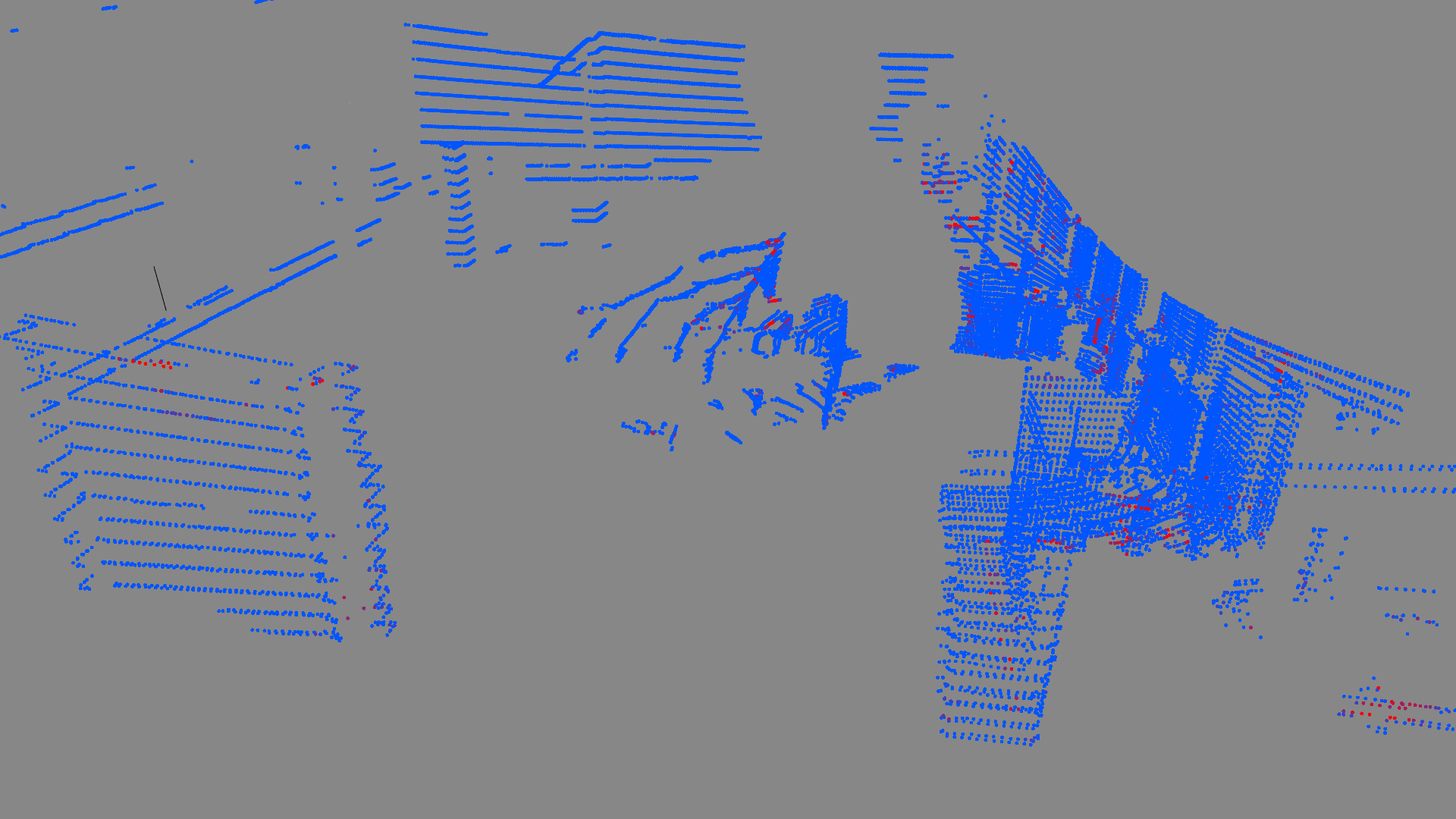}\label{fig:settings_b}}\hfill
    \subfloat[]{\includegraphics[trim={0.0cm 2cm 0cm 0cm},clip,width=0.249\linewidth]{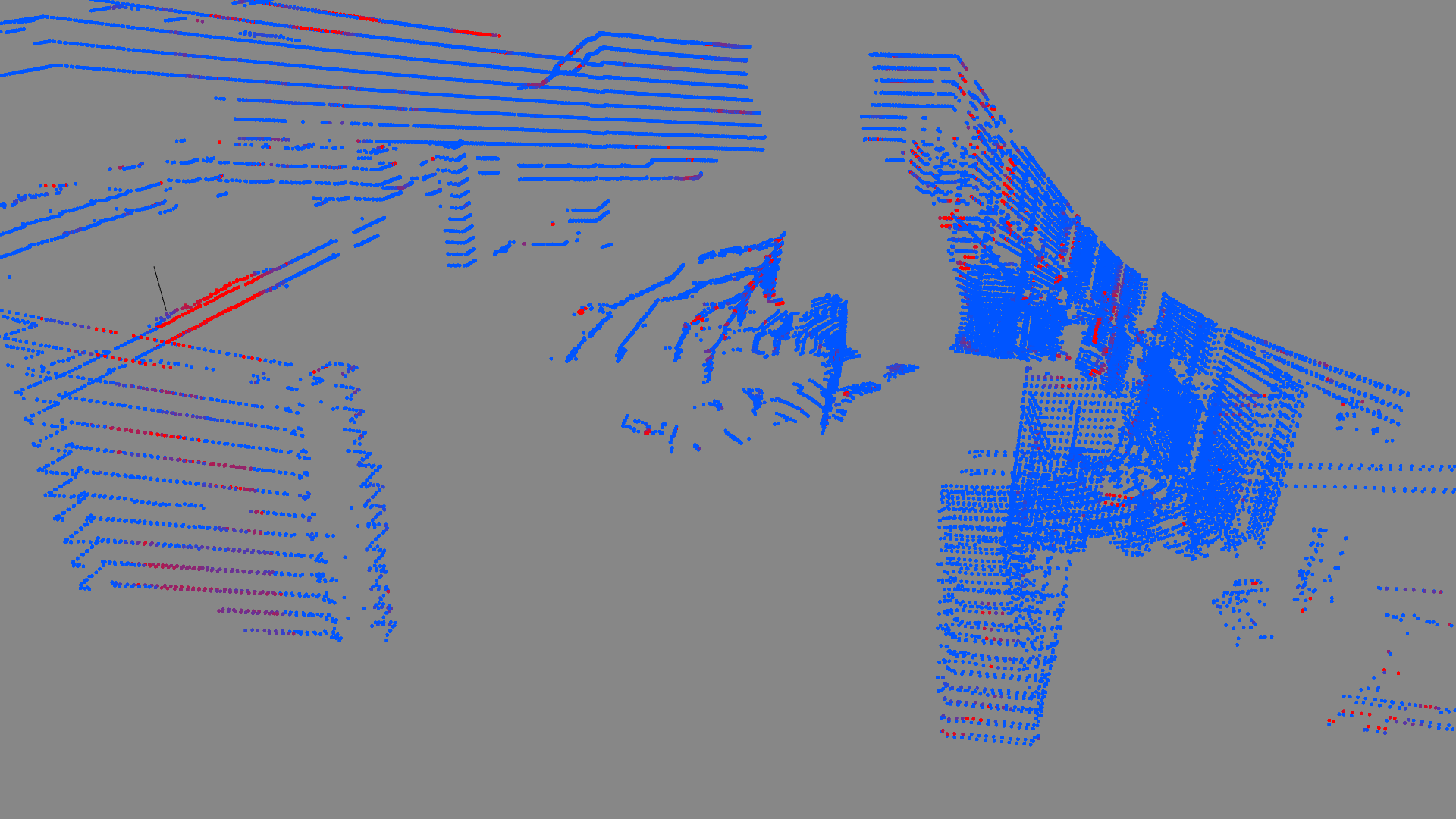}\label{fig:settings_c}}\hfill
    \subfloat[]{\includegraphics[trim={0.0cm 2cm 0cm 0cm},clip,width=0.249\linewidth]{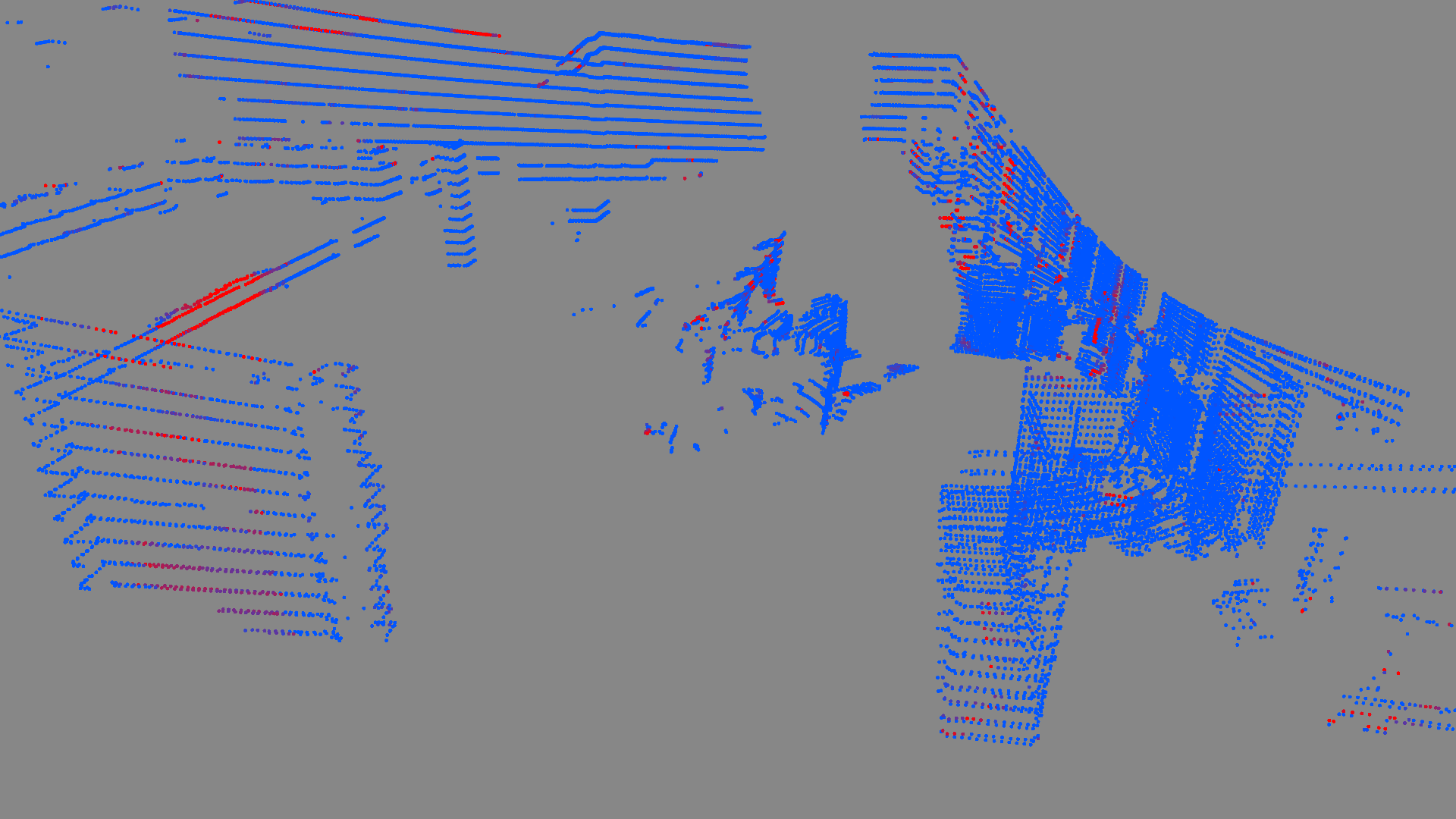}\label{fig:settings_d}}\\
    \vspace{-0.2cm}
     \subfloat[]{\includegraphics[trim={0.0cm 2cm 0cm 0cm},clip,width=0.249\linewidth]{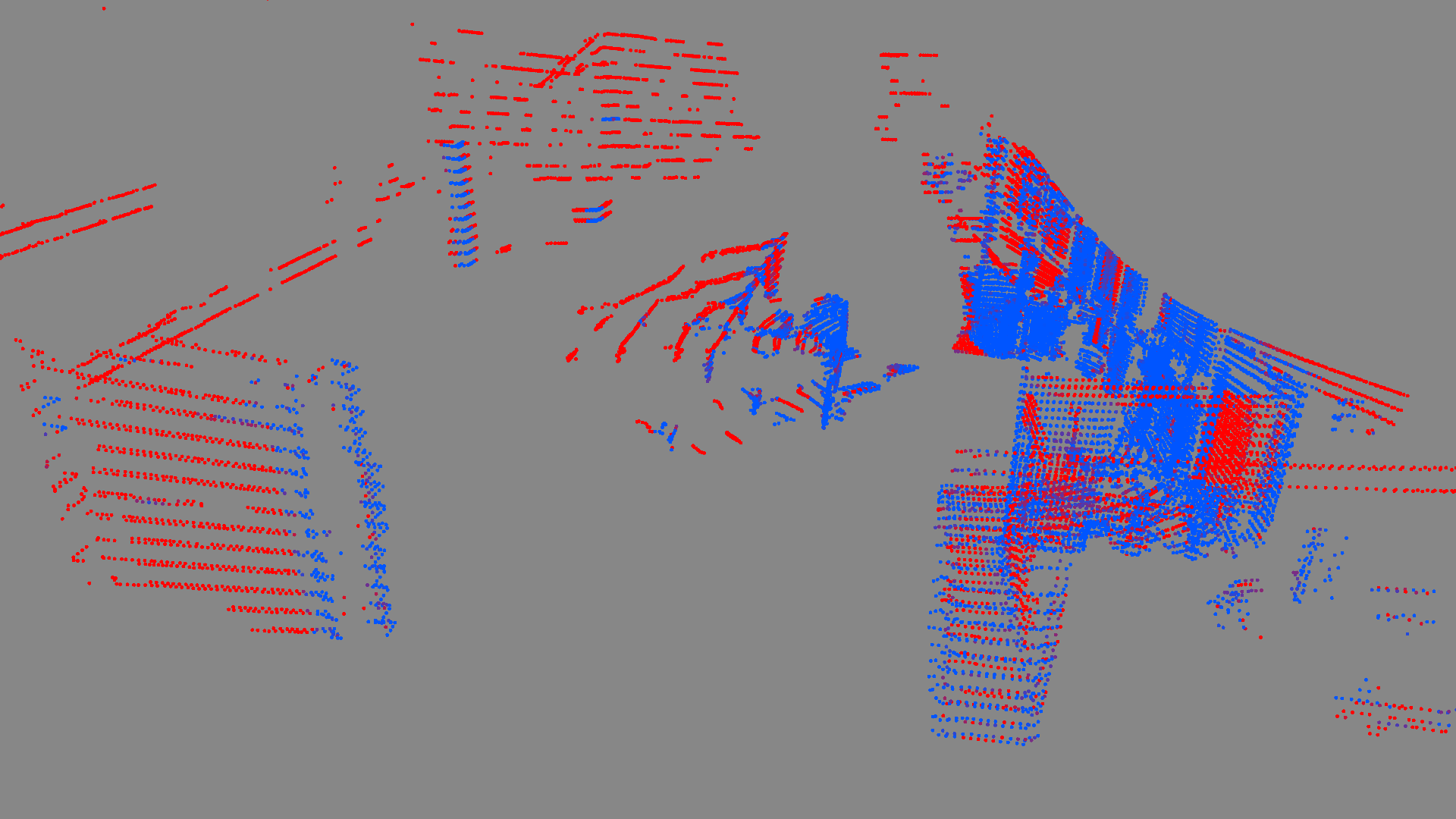}\label{fig:settings_e}}\hfill
    \subfloat[]{\includegraphics[trim={0.0cm 2cm 0cm 0cm},clip,width=0.249\linewidth]{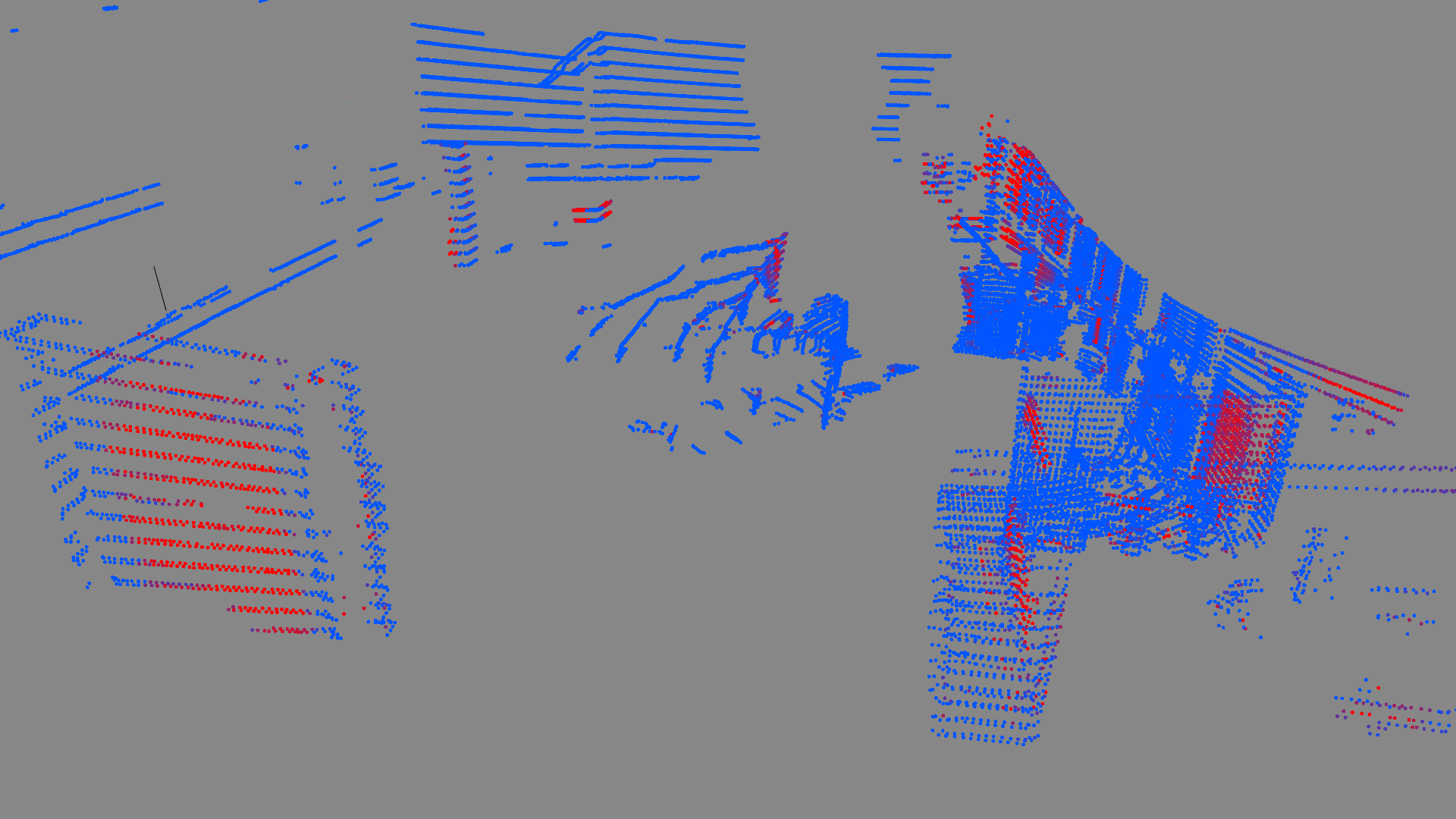}\label{fig:settings_f}}\hfill
    \subfloat[]{\includegraphics[trim={0.0cm 2cm 0cm 0cm},clip,width=0.249\linewidth]{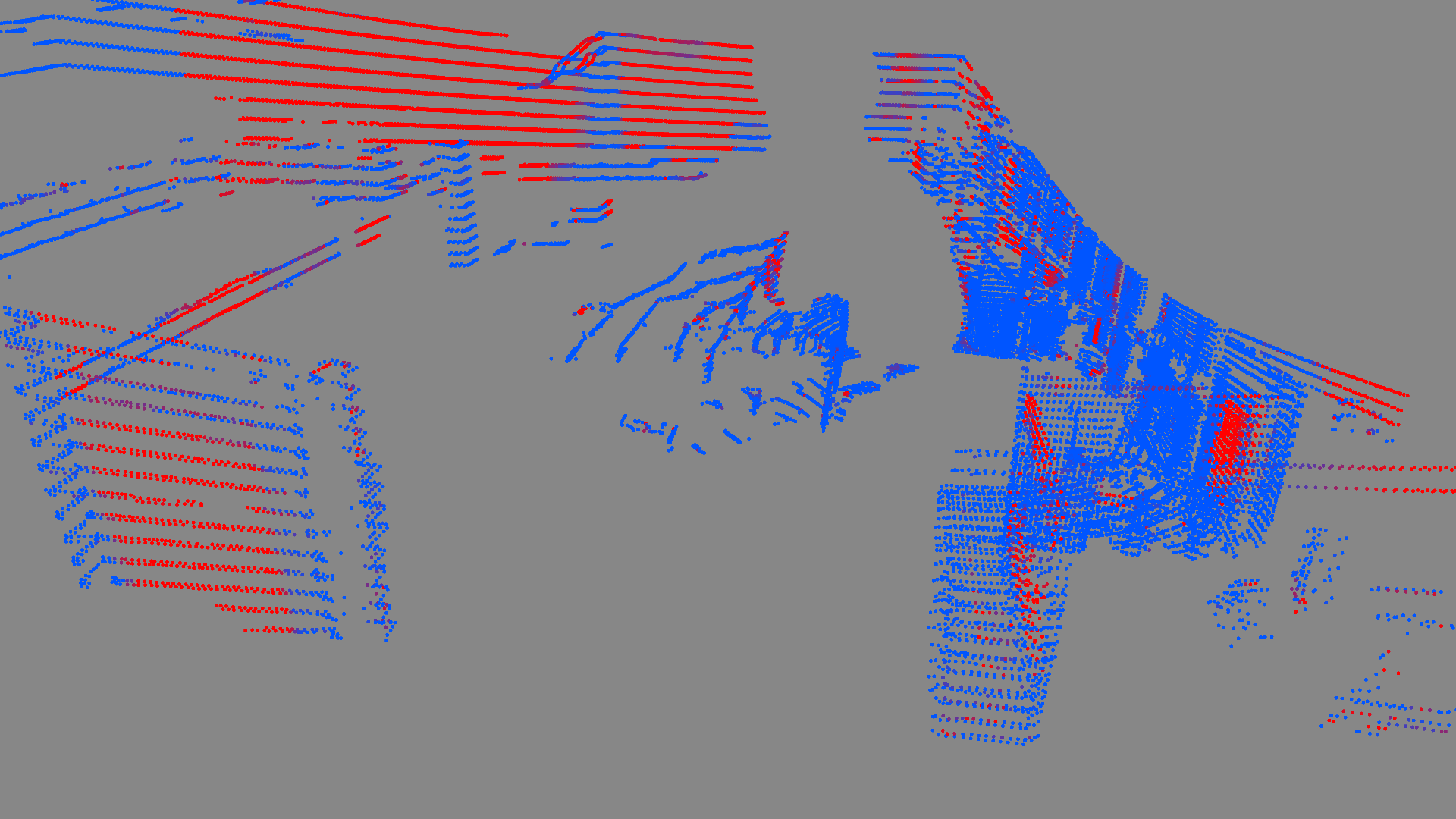}\label{fig:settings_g}}\hfill
    \subfloat[]{\includegraphics[trim={0.0cm 2cm 0cm 0cm},clip,width=0.249\linewidth]{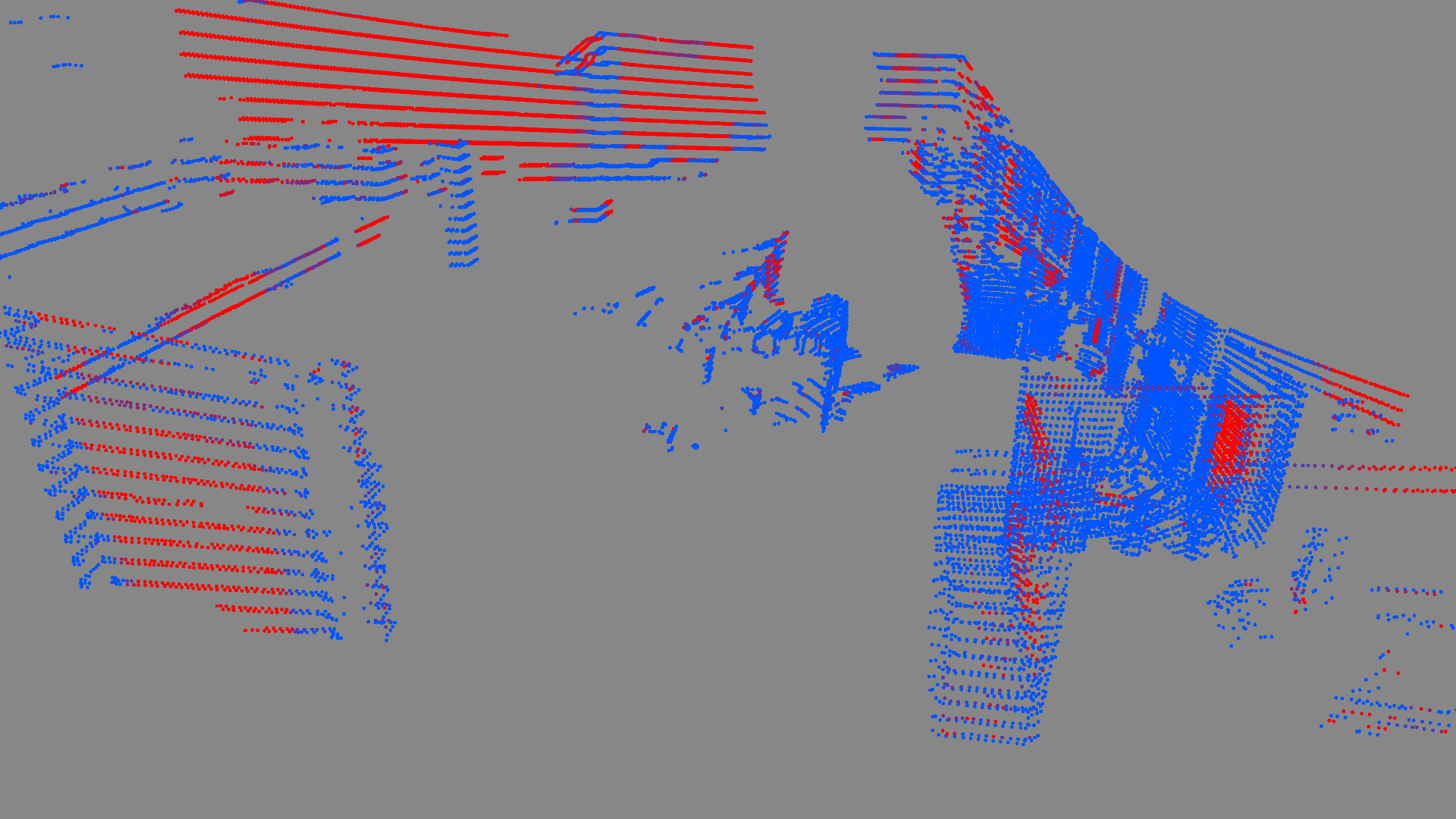}\label{fig:settings_h}}
        	\caption{
        	Joint point clouds colored by per-point quality $q_k$ ranging from blue (aligned) to red (misaligned). The location of the point cloud origin is highlighted in \Cref{fig:orkla}.
        	Columns depict the same parameters for aligned (top) and misaligned (bottom) point clouds. 
        	\textit{(a,e):} A fixed radius $r=0.3$~m gives a low $Q_s$ ratio from \Cref{eq:Qs}  $(Q_s=1.46)$, which makes it harder to learn a threshold to reliably separate aligned and misaligned point clouds. 
        	\textit{(b,f):}  Dynamically adjusted radius $r_{min}=0.3$~m,$\alpha=1.33,r_{max}=0.7$~m gives $(Q_s=2.93)$. 
        	\textit{(c,g):}  $\epsilon=10^{-8}$ is added $(Q_s=4.06)$. 
        	\textit{(d,h):} $\Ereject=10\%$ is added. Then $Q_s=4.3$, and the point clouds are clearly separable.
    	 }\label{fig:modifications}
  \end{center}
  \vspace{-0.3cm}
\end{figure*}

We propose three optional strategies to address the ill-conditioned covariances due to variations in sampling density originating from the sensor. Option (1) is specifically intended to address the characteristic ring pattern that produces variations in sparsity.
Option (2) that uses a dynamic radius can benefit both spinning lidar or radar where sparsity increase with range. Option (3) is intended mainly for 3d point clouds.

%Detecting small angular errors which displaces an observation proportionally to its range from the sensor origin.

(1): Eq.~\ref{eqn:point_entropy} is modified to $h_{i}(\bold{p}_k)=\frac{1}{2}\ln(2\pi e\det(\bold{\Sigma(p}_k))+\epsilon)$ where $\epsilon$ limits the lowest possible entropy. 
This makes sure that entropy is similar for points distributed along a line and a plane.
%The change make sure that points distributed along a line or a plane are assigned similarly high entropy. 
The improvement can be seen by comparing  %Fig.~\ref{fig:modifications}(a-b).
the first and second column in \Cref{fig:modifications}.

(2): A dynamic radius enables the quality measure to include more points far from the sensor and correctly detect alignment and misalignment at large distances as depicted in the third column of \Cref{fig:modifications} (c \& g).
Radius $r$ is chosen based on the distance $d$ between the point $\bold{p}_k$ and the sensor location to account for that point density decrease over distance. The radius is hence selected as: $r=d \sin(\alpha)$ in the range $r_{min} < r < r_{max}$ where $\alpha$ is the vertical resolution of the sensor provided by the data-sheet. %and is not tuned by the user
For other sensor types e.g. RGB-D, the resolution could be chosen similarly according to the angular sensing resolution.

(3): Remove $\Ereject$ percent of points $\bold{p}_k$ with the lowest entropies.
The effect is depicted in %Fig.~\ref{fig:modifications}(d). 
the rightmost column of \Cref{fig:modifications}.

\subsection{CorAl for spinning radar}\label{sec:classifier}
Given the presented CorAl approach, this section describes the additional steps used to enable CorAl to operate on spinning radar data. It contains a short introduction to the format of spinning radar data and proposes a novel feature extraction method that produces a high-quality point cloud from radar data.

%In this section we describe the format of spinning radar data, present a radar filter and propose a novel feature extraction strategy.

\begin{figure}
    \centering
    \includegraphics[width=\linewidth]{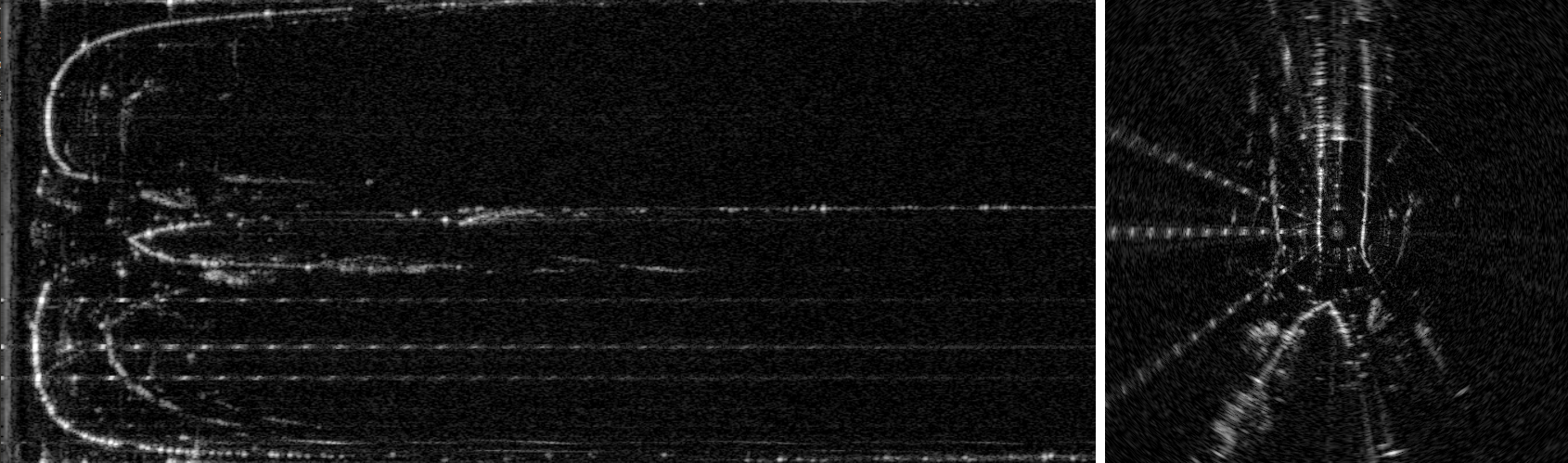}
    \caption{Polar and Cartesian spinning radar data.}
    \label{fig:radar_polar_cart}
\end{figure}

Spinning FMCW radar produces $360^\circ$ sweeps on polar coordinates seen in Fig.~\ref{fig:radar_polar_cart}. The raw data is represented as a matrix ($Z_{N_a \times N_r}$).
%For a given max range $R$ and range resolution $\gamma$, the polar image contains $N_r=R/\gamma$ range bins (columns in \Cref{fig:radar_polar_cart}).
The radar outputs $N_r$ range bins (the number of columns in \Cref{fig:radar_polar_cart}), given the max range $R$, the range resolution is $\gamma = R/N_r$.
Likewise, the radar provides $N_a$ azimuth bins (rows in \Cref{fig:radar_polar_cart}).
Each pixel $(a,r)$ with $a\in \{1..N_a\}, r\in\{1..N_r\}$ 
holds the reflected intensity and can be converted into Cartesian space as
\begin{equation}
%\begin{aligned}
\label{eq:cartesian}
    \mathbf{p} =
    \begin{bmatrix} p_x\\p_y
    \end{bmatrix}=
    \begin{bmatrix} r \gamma \text{cos}\left(\theta\right) \\ 
    r \gamma \text{sin}\left(\theta\right)
\end{bmatrix}
%&\mathcal{P}_f = \{\Mathbf{p}\}
%\end{aligned}
,
\end{equation}
 where $\gamma$ is the range resolution of the radar and the azimuth angle $\theta$ can be computed via $\theta=2\pi a/N_a$.

\subsection{Computation of Radar Intensity Peak (RIP)-features}
\label{sec:rip-features}
From raw radar data as seen in \Cref{fig:radar_polar_cart}, the goal is to produce accurate point clouds suitable for alignment classification.
We build on top of the radar filter ``$k$-strongest''~\cite{9636253} that efficiently computes a mask that removes noise and keeps significant features that are useful for odometry estimation. 
%$k$-strongest operates on azimuths 
Over all the $1..N_r$ range bins, the $k$ highest intensity bins that additionally exceed the expected noise level $z_{min}$ are selected. The noise level $z_{min}$ mitigates speckle noise and uncertain or false detections in absence of real obstacles. The limitation of $k$ returns per azimuth complements $z_{min}$ by mitigating multi-path reflections and receiver saturation under the assumption that true landmarks have higher intensity. 

\begin{figure}
    \centering
    \includegraphics[width=\linewidth,trim={2.7cm 3.1cm 10cm 4cm},clip]{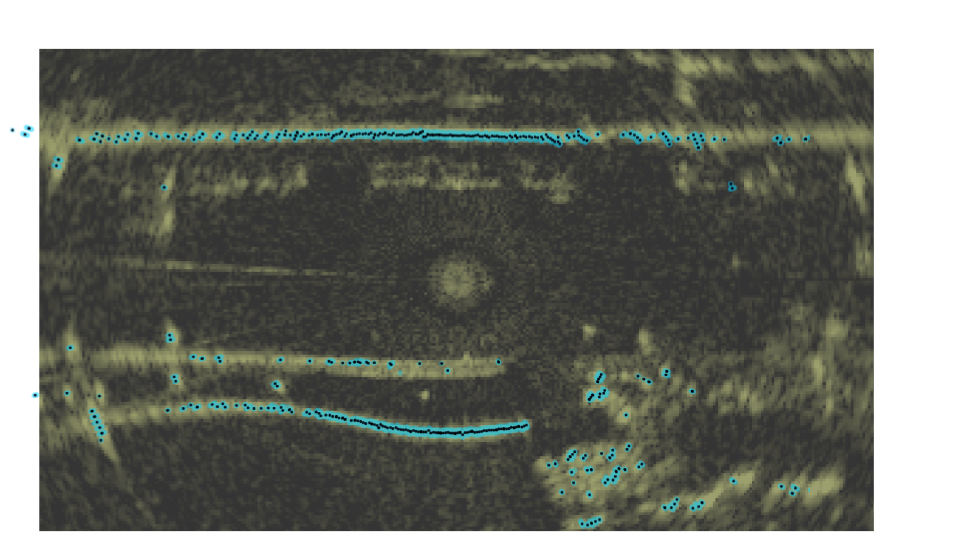}
    \caption{Sparse radar filtering presented in Cartesian space using %the proposed method. 
    the RIP (Radar Intensity Peaks) features proposed in \Cref{sec:rip-features}.
    The raw radar data (black-yellow) is first masked using ``$k$-strongest'' (cyan) to find areas that contain landmarks.
    Stable RIP-features (black dots) are then computed accurately around intensity peaks.}
    \label{fig:radar_comparison}
\end{figure}

This method efficiently provides a mask in regions around true landmarks and generalizes well across environment types (the same $k$ and $z_{min}$ values work well in our odometry pipeline~\cite{9636253} for the road as well as underground applications). However, it does not accurately reconstruct landmark surface locations. We tested this representation and found that the method is unsuitable for detecting small misalignments. For that reason, we propose an additional step aiming to further analyze masked regions and compute stable features located on Radar Intensity Peaks (RIP)-features. 
We aim to efficiently and accurately detect surface locations by finding peaks within masked regions where intensity is consistently high over a local neighbourhood. To do so, we combine non-maximum-suppression on the 1D intensity-range signals together with a region strength criterion. Azimuths bins are analyzed independently without considering neighboring azimuth bin.
For each range bin within the masked radar image, and neighboring range bins outside the masked region, the region strength is computed as the average intensity within a window (with size $w$) of neighboring range bins.
Second, we select all range bins where region strength exceeds neighboring bins with the additional criteria that region strength must exceed the expected region noise floor. The algorithm is formally described in Alg.~\ref{alg:frip}. 
Typical behavior is depicted in Fig.~\ref{fig:radar_comparison}.
Structures such as buildings and vehicles appear with higher intensity and give rise to the most stable features; these are good candidates for detecting misalignment as surface location uncertainty is low. When there are secondary landmarks within a single beam observed with intensity, e.g. a bush in front of a building, these give rise to features more scarcely. Moving vehicles have little impact on the CorAl score if driving sufficiently fast, i.e. for a radar spinning at $4$~Hz, vehicles moving faster than $14.4$~km/h will be outside the CorAl radius ($r=1$~m) boundary within consecutive observations, and will not contribute to misalignment.

%We build upon the previous radar filtering technique $k-strongest$~\cite{9636253} that masks the polar radar image per by thresholding based on expected noise level $z_min$ and by additionally limit the nuber of returns to a maximum of $k$ per azimuth. This produces a mask that efficiently provides attention to the most important landmarks in the scene while discarding frequently occuring multipath reflections and less important secondary detections. 
%As an additional step, we use a peak detector that operate per azimuth on masked regions in the image together with non-maximum suppression.  

\newlength\myindent
\setlength\myindent{2em}
\newcommand\bindent{%
  \begingroup
  \setlength{\itemindent}{\myindent}
  \addtolength{\algorithmicindent}{\myindent}
}
\newcommand\eindent{\endgroup}

\begin{algorithm}
    \caption{RIP-Feature extraction}\label{alg:frip}
    \textbf{Input:} $Z_{N_a \times N_r}$ \textit{Polar radar image}\\
    \textbf{Parameters:} window size $w$, masked points per azimuth $k$, noise filter $z_{min}$\\
    %$w$ //window size\\ 
    %\hspace{2cm} $k$ // $kstrongest$\\
    %\hspace{2cm} $z_{min}$ // $kstrongest$\\
    \textbf{Output:} $\mathcal{F}_{rip}=\{(r,a)_i\}$ \textit{set of range and angle tuples}
    \begin{algorithmic}[1]
    \State $\mathcal{F}_{rip} \gets \emptyset$
    \For{$a \gets 1..N_a$ } \textit{//all azimuth bins}
        %\State //Masking via $k-strongest$, subset of range bins
        \State $\mathcal{K} \gets kstrongest(\: Z(a,1..N_r),\: k,\: z_{min})$ %//bins of interest
        \State Initialize table \textbf{S} \textit{//Scores}
        \For{$r \in \mathcal{K}\cup neighbors(\mathcal{K},w)$}
        \State $Scores[r]= 1/(2w+1)\sum_{n=r-2}^{r+2}  = Z(a,n)$ 
        \EndFor
        \For{$r \in \mathcal{K}$} \textit{//only search} masked points
            \If{$\textbf{S}[r]>=max(\: neighbors(r,w)\: )$ }
                \If{$\textbf{S}[r] > z_{min}$ }
                    \State $\mathcal{F}_{rip} \gets \mathcal{F}_{rip} \cup \{(r,a)\}$
                \EndIf
            \EndIf
        \EndFor
    \EndFor
    \end{algorithmic}
\end{algorithm}

\subsection{Self-supervised learning of alignment classification}\label{sec:classifier}
We learn alignment classification based on our quality measure in a self-supervised fashion from an accurate sensor pose signal. In this paper, we use either an external ground truth system or a lidar/radar odometry estimator. For each pair of scans, we compute the quality measure 1) directly and 2) after inducing a (sensor frame) offset in position and orientation  on the later scan location. This allows the system to produce positive (aligned) and negative (misaligned) data points for training and to learn classification boundaries according to the magnitude of induced errors. To avoid over-fitting and to produce valuable insight into the alignment class separability based on the evaluated score functions, we use a simple logistic regression as a model for classification:
\begin{equation}
\begin{split}
    \label{eq:class-pz}
    &p=\frac{1}{1+e^{-z}},\\
    &z=\beta_0+\beta_1x_1+\beta_2x_2,\\
\end{split}
\end{equation}

\begin{equation}
    \label{eq:classifier}
       y_{pred}= 
\begin{cases}
  \mathrm{aligned} & \mathrm{if} \: p\geq t_h \\
      \mathrm{misaligned} & \mathrm{if} \: p<t_h,
\end{cases}
\end{equation}
where $p$ is the class probability and $x_1,x_2$ are input variables to which we pass quality measures. For the CorAl quality presented here, we refrain from passing the one variable quality measure $Q=H_{joint}-H_{sep}$ and instead pass the joint and separate entropies as: $x_1=H_{joint},x_2=H_{sep}$. This allows the model to learn the mapping from $H_{sep}$ and $H_{joint}$ to alignment probability $p$ implicitly.

%Instead of passing the quality measure $x_1=Q(\mathcal{P}_a,\mathcal{P}_b)$, $H_{sep}$ and $H_{joint}$ are passed separately to $x_1=H_{joint}$ and $ x_2=H_{sep}$.
The model parameters in Eq.~\ref{eq:class-pz} ($\beta_0$, $\beta_1$, $\beta_2$) are learned during training. The classification probability threshold $t_h$ can be adjusted after training has been carried out in order to balance sensitivity to false-positive rate. Increasing the threshold will cause fewer well-aligned pairs to be correctly classified as aligned (decrease recall), but will reduce the rate of misaligned pairs being falsely be classified as aligned (increase precision). In our experiments we used the default threshold $t_h=0.5$. During training, weights of data points were adjusted inversely proportional to class occurrence to mitigate bias.

%For example, in mobile robotics, it is desired that misaligned point clouds are not accidentally reported as aligned (false positives), potentially causing a system failure. In contrast, aligned point clouds classified as misaligned are typically harmless. For that reason, $t_h$ can be increased to reject false positives and hence improve robustness. We used the default threshold $t_h=0.5$. During training, weights of data points was adjusted inversely proportional to class occurrence to mitigate bias.

%begin{algorithm}
%\caption{An algorithm with caption}\label{alg:cap}
%\hspace*{\algorithmicindent} \textbf{Input:} $Z_{N_a \times N_r}$ \\
% \hspace*{\algorithmicindent} \textbf{Output} 
%\begin{algorithmic}
%\State asd
%\state input polar radar image $Z_{N_a \times N_r}$ 
%\State output a vector of range bins with landmarks
%\State $r$ \gets $k strongest(Z_{a, 1..N_r})$ %//vector of range bins
%\State $score$ is empty vector
%\State for each  {$i \in r $}
%\state $score_i$ = average intensity of window with size $w$\\
%\state peaks \gets peaks\\ %defined as all range bins which have scores larger than the score of their neighbours
%\end{algorithmic}
%\end{algorithm}

%FOR EACH AZIMUTH $a\in Z$
%$\mathcal{R}=Kstrongest(a)$ %get range bins with highest intensities
%$r \in R$
%for each remaining point in $p_r$,
%calculate the score by average intensity   $z_{mean}$ within a window of size $w=2$ neighbouring pixels in the raw radar image.
%$s_r =1/(window+1)\sum z_r$
%select all range bins where the score is higher than their closest neighbour and the score is higher than 

%%END FOR
`
\section{Evaluation on lidar data}
\label{sec:lidar_classification}
%To allow for comparison with previous work on lidar alignment classification, we follow the convention of inducing errors in previous work

In this section, we present a quantitative evaluation of alignment quality classification with CorAl for 3D lidar data.
%More details about the lidar evaluation can also be found in~\cite{adolfsson-2021-coral}.
An evaluation for 2D radar data will follow in \Cref{sec:radar_classification}.

In order to compare the method presented in this paper to previous work related to 3D point cloud alignment assessment, we follow the procedure of evaluation as carried out in~\cite{Almqvist}.
For the results in this section, we use an equal portion of aligned and misaligned point clouds, where misaligned point clouds are created by adding an offset for each point cloud pair: an angular offset ($e_{\theta}=0.57^\circ = 0.1$~rad) around the sensor's vertical axis and a random translation $(x,y)$ offset at a distance ($e_d=0.1$~m) from the ground truth. These errors are large enough to be meaningful to detect in various environments, yet challenging to classify.
%\paragraph{Input to the classifier}
%CorAl, FuzzyQA and Rel-NDT output two decision variables that are passed as input variables  $x_1,x_2$ to the classifier (\ref{sec:classifier}). The other evaluated methods output a single variable $x_1$, and $x_2=0$ is fixed.

\subsubsection{Evaluated lidar methods}
\label{sec:methods}
 
The evaluated methods; MME, CorAL, CorAl-median, NDT, Rel-NDT and FuzzyQA are briefly summarized here together with their most important parameters. For CorAl, FuzzyQA and Rel-NDT, two values (that represent the quality measure) are passed to $x_1$ and $x_2$, for MME and NDT, a single value is passed to $x_1$ while $x_2$ is set to zero. To make a fair comparison, we use a similar radius for NDT, MME and CorAl.

\paragraph{MME} Mean Map Entropy (MME) as proposed by Droeschel and Behnke \cite{8461000}. 
The quality measure corresponds to $H_{joint}$ from Eq.~\ref{eqn:Hjoint}. The parameter is the same radius $r$ used for computing the per-point differential entropy. The MME is passed to the classifier as: $x_1=H_{joint},x_2=0$.

\paragraph{CorAl} as described in \Cref{sec:coral-entropy,sec:coral-modifications} and in \cite{adolfsson-2021-coral}.
%(proposed in the paper)} 
%Separate and registered entropy $H_s ,H_j$ as described in~\cref{eq:Hseparate,eq:Hjoint}. 
Parameters are $r_\mathrm{min}$, $r_\mathrm{max}$ and $\alpha$ to determine nearby points radius, and $\Ereject$ to set outlier rejection ratio and $\epsilon$ for mitigating ill-conditioned entropies. The joint and separate entropy are passed separately to the classifier: $x_1=H_{joint},x_2=H_{sep}$, an intuition is presented in ~\ref{sec:classifier}.

\paragraph{CorAl-Median} % (proposed in the paper)} 
$H_s, H_j$ are modified to calculate the \textit{median} entropy rather than the mean entropy, we hypothesize that this modification can be more robust to outliers. Except for modification, we used the same parameter and methodology described in the previous paragraph.

\paragraph{NDT (point-to-distribution normal-distributions transform)}
The NDT quality describes the likelihood of finding the points in $\mathcal{P}_b$, given the NDT representation of $\mathcal{P}_a$.
The method uses the 3D-NDT~\cite{magnusson-2007-jfr} representation, which constructs a voxel grid over one point cloud, and computes a Gaussian based on the points in each voxel.
The likelihood of finding the points in $\mathcal{P}_b$ is computed as
%\begin{equation}
%    p(x)=d_1e^{-(d_2/2(x-u)\Sigma^{-1}(x-u)}
%\end{equation}
\begin{equation}
    s = \frac
    {\sum_{k=1}^{n} \tilde p(\point_k)}
    {n}
    ,
    \label{eqn:ndt_point}
\end{equation}
where $n$ is the number of overlapping points, defined as those points that fall in an occupied NDT voxel, or in a voxel that is a direct neighbor of an occupied voxel, and $\tilde p$ is the probability density function associated with the nearest overlapping NDT-cell.
This is similar to the ``NDT3'' variant in~\cite{Almqvist}.
The most important parameter for NDT is the voxel size $v$ which is set equal to $2*r$ in our evaluation, as this makes the volumes used for the sample covariance of NDT cells and the entropy in CorAl comparable. The NDT quality $s$ is passed as: $x_1=s$.
%Neighborhood of association is set to a maximum of one voxel.

\paragraph{Rel-NDT}
%(presented in the paper)} 
as described in~\cite{adolfsson-2021-coral}.
This variant aims to improve the re-utilization of learned NDT classification parameters when applied to new environments. %We wanted to investigate if entropy can be used to improve the generalization of NDT to different environments. 
The idea (similar to CorAl) is that environment type is reflected in the average entropy of the scene and can be combined with the NDT score to improve the classification. Similar to the MME, we compute the average differential entropy. However, instead of recomputing the sample covariances around each point $p_k$, we use the %MME over from 
covariances directly from the NDT representation of $\mathcal{P}_a$. This is only done in overlapping regions and no additional parameter to NDT is required. The NDT score $s$ and the overlapping NDT differential entropy (NDT-entropy) is passed as: $x_1=s, x_2=NDT-entropy$

%We did this by computing the average entropy of all NDT-covariances associated with $\point_k$  in the point-likelihood terms  and feed that together with the NDT score \eqref{eqn:ndt_point} to the classifier. 

\paragraph{FuzzyQA} FuzzyQA \cite{fuzzybnb} measures the alignment quality by a ratio $\rho=\frac{\mathrm{AFCCD}}{\mathrm{AFPCD}}$, where AFCCD and AFPCD are two indexes describing the points' disposition and dispersion around fuzzy cluster centers. In \cite{fuzzybnb}, two point clouds are considered to be coarsely aligned if $\rho<1$. In this paper, we pass AFCCD and AFPCD separately to the classifier input $x_1=AFCCD,x_2=AFPCD$ in order to learn a generalizable separator.
%The two point clouds are coarsely aligned if $\rho<1$. However, AFCCD and AFPCD are passed separately to the classifier input $x_1,x_2$.
%\\[\baselineskip]
\subsection{Qualitative evaluation, live robot data}
\label{sec:orkla-eval}
\begin{figure}
    \centering
    \includegraphics[width=\linewidth]{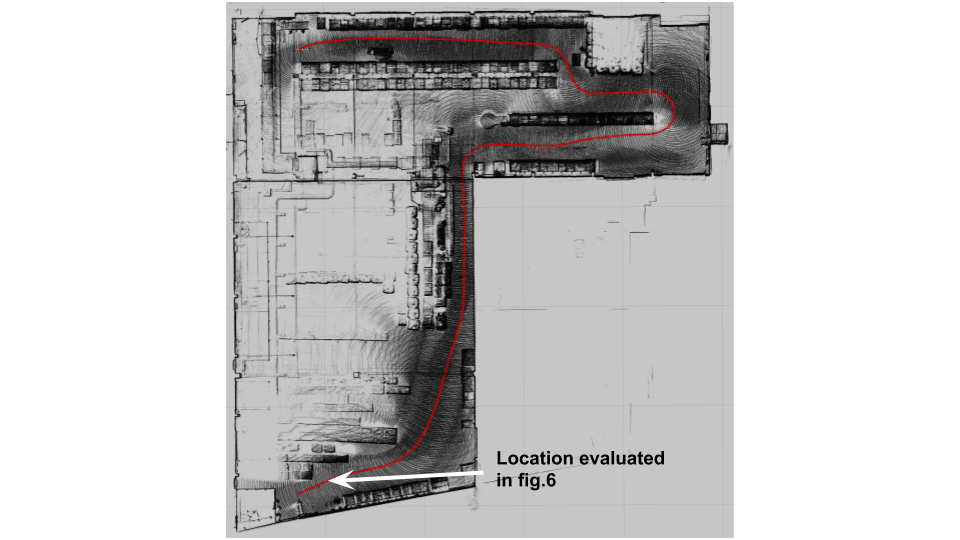}
    \caption{\label{fig:orkla}Data acquired by a truck in a warehouse environment. The sensor trajectory is drawn in red. 
    The environment in the figure is 50 m $\times$ 50 m and the sequence length is 102~m.
    In the first segment of the trajectory, starting at the bottom left, the walls are clearly visible. The final segment is located between aisles where walls are typically out of sight and the sensor observes more complex structures such as shelves. The truck has worn out  slightly oval wheels which introduces a large degree of vibrations, hence alignment quality is expected to be low.
    }
    \vspace{-0.3cm}
\end{figure}

%We started our evaluation on a real robot scenario in a warehouse environment. 
First, we present qualitative results from real-world data in a %structured
warehouse environment.
A forklift equipped with a Velodyne HDL-32E 3D laser scanner was manually driven at fast walking speed in the warehouse depicted in \Cref{fig:orkla}. The environment covered in this data set varies from large and open areas with walls in line of sight to small and narrow aisles between shelf racks.
To generate ground-truth alignments for the warehouse dataset, we first aligned the point clouds 
using a scan-to-map approach~\cite{softconstraints}. We have then inspected the alignment between subsequent scans and found that at least 40/484 (8.3\%) point clouds were impaired by rigid misalignments or non-rigid distortions from vibrations and motion to the extent that these could be visually located. Alignment classification was then performed on the remaining scans by inducing errors as described in \cref{sec:lidar_classification}. 
We used the following parameters as they provided a relatively high value of $Q_s$ for the first scan pair in the dataset: $\alpha=0.92^{\circ}$, $\Ereject=0.2$, $r_{min}=0.2$~m, $r_{max}=1.0$~m and voxel size $v=2r_{min}=0.4$~m.
We found that CorAl-mean, MME and NDT reached an accuracy of $96\%$, $70\%$ and $99\%$ respectively. In this case, NDT performs slightly better than CorAl. We believe that the relatively lower result is due to the experimental setup. CorAl is highly sensitive to low quality in training data. In this case, even after removing the worst data points in the ground truth set, the level of vibrations from the worn-out wheels remain high in a large number of data points used for training and evaluation. This makes it challenging to learn the detection of small errors using the CorAl quality measure.
%the CorAl quality measure is highly sensitive to small errors, learning to detect small errors with low quality data is challenging.
Whether this is high sensitivity is the desired behavior depends on the application.

\subsection{Quantitative evaluation, ETH benchmark data set}
\label{sec:eth-eval}
Our main quantitative evaluation of CorAl's performance on 3D lidar point clouds is done by using the public ETH registration dataset \cite{Pomerleau_2012}. It contains several sequences representing a wide variety of environments, which will serve us to evaluate how well CorAl generalizes across different kinds of them. Specifically, this dataset includes 3 sequences in structured environments (Apartments, ETH Hauptgebaude, Stairs), 3 sequences in semi-structured environments (Gazebo in summer, Gazebo in winter, Mountain plain) and 2 challenging sequences in unstructured environments (Wood in summer, Wood in autumn). %These sequences cover a wide variety of environments, which will serve us to evaluate how well CorAl generalizes across different kinds of them.  %These varied environments allow us to understand how CorAl generalize across environment type.
In \Cref{fig:ETH_intra_medium,fig:ETH_generalization_medium,fig:ETH_Joint_medium}, the training results for structured environments are shown in blue, the ones for semi-structured environments in brown, and the ones for unstructured environments in green.
Each sequence contains between 31 and 47 scans acquired from stationary positions.
%For more information we refer to the original data-set paper~\cite{Pomerleau_2012}.
The dataset contains accurate ground truth positions, required to evaluate the different methods. 
In order to make the evaluation fairer, more realistic, and applicable to real applications, we downsampled the original dense point clouds using a voxel grid of 0.08~m. In all the training experiments, CorAl has been run on an Intel Core i7-7820X desktop CPU, achieving an overall run-time of $0.246\pm 0.095$ seconds per point cloud pair. Also, since this dataset has less variation in sampling density compared to the warehouse one employed in \Cref{sec:orkla-eval}, we used a fixed radius $r=0.3$~m %($\alpha=0, r_{min}=r_{max}=r)$ 
and set $\Ereject=20\%$ and $\epsilon=0$. Finally, the NDT voxel size was set equal to two times the CorAl radius, i.e, $v=2r=0.6$~m. This way, the diameter of influence for CorAl and the width of NDT cells are similar.

In these conditions, we carried out three different kinds of training, which are aimed at showing how the proposed CorAl method can achieve generalization. Training has been performed with increasing difficulty, as explained below.
%We carried out three types of training... The aim is to understand how generalization can be achieved... separate to generalizaiton easy to hard
%\paragraph{Performance}
%CorAl has an overall run-time of $0.246\pm 0.095$ seconds per point cloud pair on an Intel Core i7 and depends on the point cloud density. 

\label{sec:eval}

\subsubsection{Separate training}
\begin{figure}
    \centering
    \includegraphics[trim={2cm 0cm 2cm 0cm},clip,width=\linewidth,angle=0]{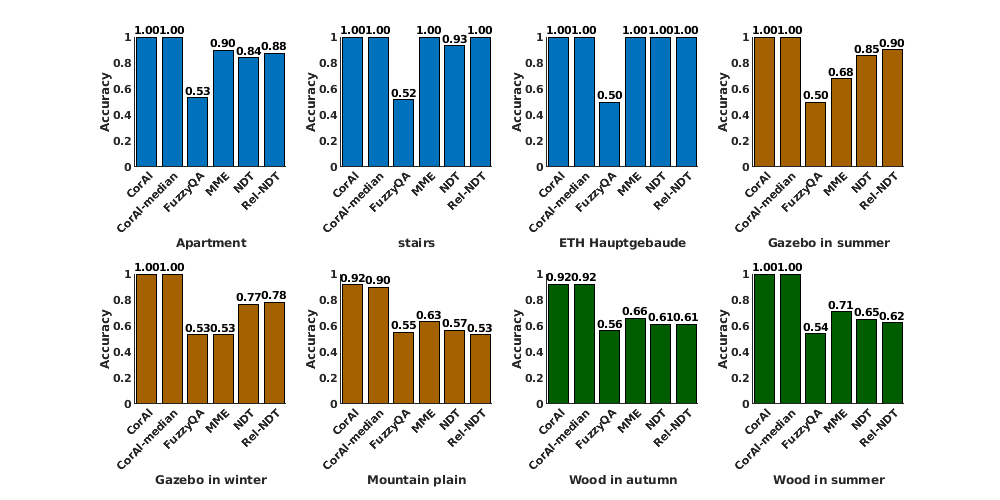}
    \caption{Separate training. 
    The overall accuracy was CorAl: $98\%$, CorAl-median: $98\%$ FuzzyQA: $53\%$ MME: $77\%$ NDT: $78\%$ Rel-NDT: $80\%$}
    \label{fig:ETH_intra_medium}
    \vspace{-0.5cm}
\end{figure}

First, we evaluate the capability to learn classification in a specific type of environment. 
The classifiers were trained and evaluated on each sequence separately, using 5-fold cross-validation.
This evaluation  serves as a reference for the cross-environment evaluations below. 
Results are shown in \Cref{fig:ETH_intra_medium}.
We found that all methods except FuzzyQA performed well in structured environments. For instance, MME scored around 90--100\%, which clearly indicates that even a method that is highly influenced by the environment can successfully assess alignment quality if the environment is structured and not changing substantially. In contrast, we did not expect that FuzzyQA would achieve a good classification performance since it is specifically designed to classify coarse alignment.

%We did not expect that FuzzyQA would handle this as it is specifically designed to classify coarse alignment. For instance, MME scored around 90--100\% on the structured environment, which indicates that these methods can successfully assess alignment quality in highly structured %and controlledenvironments.

In the semi-structured and unstructured sequences, only CorAl and CorAl-median performed well, with consistently $>$90\% accuracy, even in the most challenging sequences. All other methods are only slightly better than random, except for the gazebo sequences. Rel-NDT slightly outperforms NDT, however not consistently.
Both NDT methods performed decently (77--90\%) in the gazebo sequence, but poorly in the unstructured Wood sequences (60--65\%), indicating that NDT requires at least some structure or surfaces free from foliage to be effective as an alignment correctness measure. %We believe this is because entropy alone provides little information about the environment. This is supported by the low overall accuracy of MME.

\subsubsection{Joint training}
\begin{figure}
    \centering
    \includegraphics[trim={2cm 0cm 2cm 0cm},clip,width=\linewidth,angle=0]{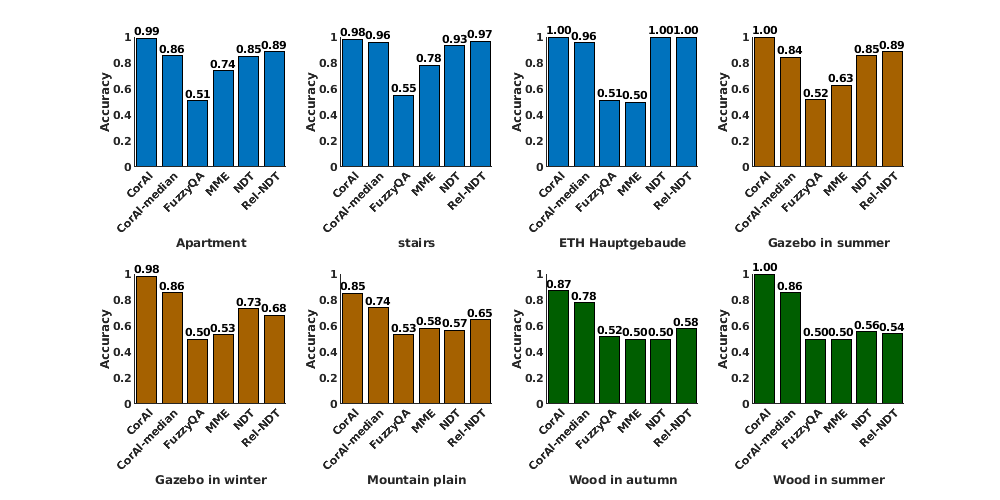}
    \caption{Joint training. Overall accuracy CorAl: $96\%$, Coral-median: $95\%$ FuzzyQA: $52\%$ MME: $60\%$ NDT: $75\%$ Rel-NDT: $78\%$}
    \label{fig:ETH_Joint_medium}
    \vspace{-0.5cm}
\end{figure}

The second test evaluates how the methods are able to learn alignment classification when trained in a variety of environments. To do that, the methods need to be versatile. The training was performed on all the ETH sequences together, and evaluation was then done on each sequence individually. The results are shown  in \Cref{fig:ETH_Joint_medium}.
As can be expected, the accuracy of all classifiers decreased compared to the previous test. CorAl still performs best, with an accuracy  of 85--100\% in all cases. CorAl-median reached a slightly lower accuracy compared to CorAl. Rel-NDT performed better than NDT in most cases, however not consistently. The generally high accuracy of CorAl indicates that it is possible to find general parameters that make the method valid in a range of substantially different environments.

\subsubsection{Generalization to unseen environments}
\label{sec:lidar-generalization}
\begin{figure}
    \centering
    \includegraphics[trim={2cm 0cm 2cm 0cm},clip,width=\linewidth,angle=0]{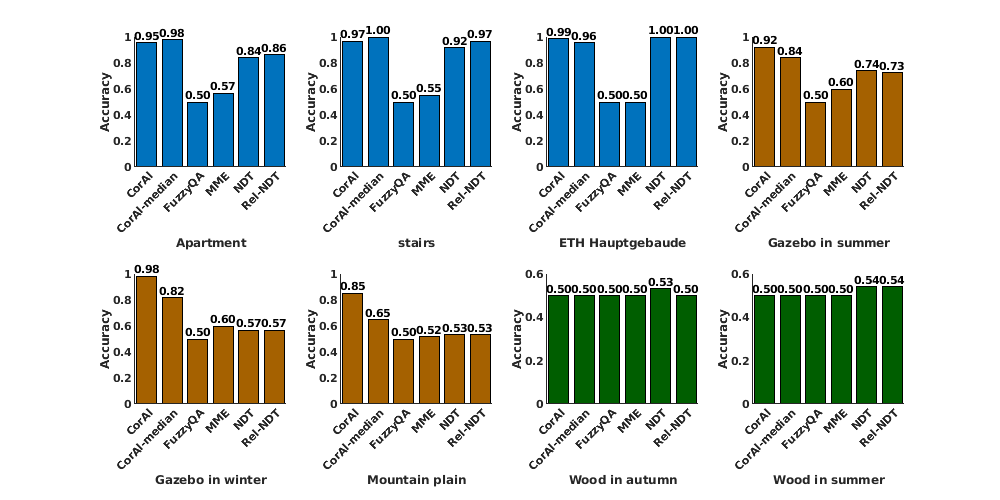}
    \caption{Evaluation on unseen environments. Overall accuracy: $83\%$, CorAl-median: $79\%$ FuzzyQA: $50\%$ MME: $54\%$ NDT: $72\%$ Rel-NDT: $72\%$, In structured and semi-structured environments: $95\%$, Coral-median: $88\%$ FuzzyQA: $50\%$ MME: $56\%$ NDT: $78\%$ Rel-NDT: $79\%$.}
    \label{fig:ETH_generalization_medium}
    \vspace{-0.5cm}
\end{figure}

The final test evaluates how classifiers perform in environments with different characteristics than those observed in the training set. We trained and evaluated different sequences and environments. The 3 structured environments were used for training and the remaining 5 (semi-structured and unstructured) were used for evaluation and vice versa. The classification accuracy is depicted in \Cref{fig:ETH_generalization_medium}.
When trained on structured  and evaluated on semi-structured environments, CorAl performed accurately (85--98\%) and other methods performed close to random except NDT for Gazebo summer ($74\%$). No method generalized well when training on structured data and evaluated on  unstructured environments. 
On the other hand, learning from semi-structured and unstructured environments was enough to afford very high accuracy in structured environments with CorAl -- very close to what was attained with joint training on all sequences. 
The previous joint evaluation shows that it is possible to train a model that is simultaneously accurate in all environment types. Hence, we believe that the reason the classifier trained in a structured environment does not generalize to an unstructured environment is that the model overfits when not trained with sufficiently diverse and challenging data.

%In all the training experiments above, CorAl has been run on an Intel Core i7-7820X desktop CPU, achieving an overall run-time of $0.246\pm 0.095$ seconds per point cloud pair. This execution time depends on the point cloud density itself.

\section{Evaluation of large-scale radar data}
\label{sec:radar_classification}
In this section, we present a thorough evaluation of the problem of alignment correctness classification using data acquired by different spinning radars. In these experiments, we have employed both CTS350-X and CIR2014-H models by Navtech. We highlight multiple challenges in alignment correctness classification of radar data and consider the impact of practical challenges such as variation of parameters, distance between scans, and error magnitudes. Similar to the generalization training carried out in \Cref{sec:eth-eval}, we again use datasets with different characteristics for training and testing in order to understand how CorAl generalizes across different environment types. We compare our method to recently published radar-specific baselines~(\Cref{sec:radar-baselines}).
% I removed "-Radar" from CorAl name
%, we include four previously published radar-specific methods for feature and alignment quality measures found in recent research. % with and without motion prior

\begin{figure}
    \centering
    \includegraphics[clip,trim={0cm 7.5cm 0cm 0cm},width=\linewidth]{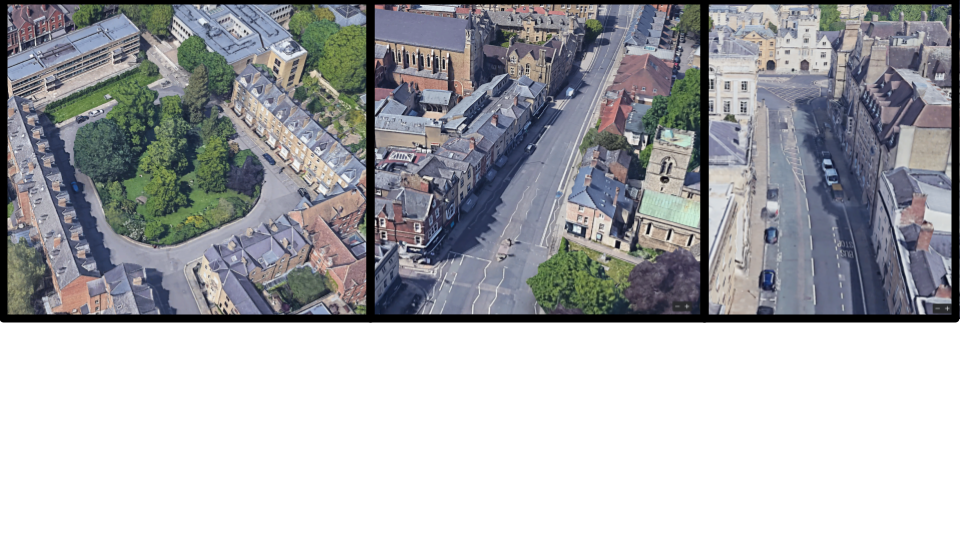}
    \caption{Locations traversed in the urban Oxford dataset.}
    \label{fig:Oxford_overview}
\end{figure}

\begin{figure}
    \centering
    \includegraphics[clip,trim={0cm 0cm 7.2cm 0cm},width=\linewidth]{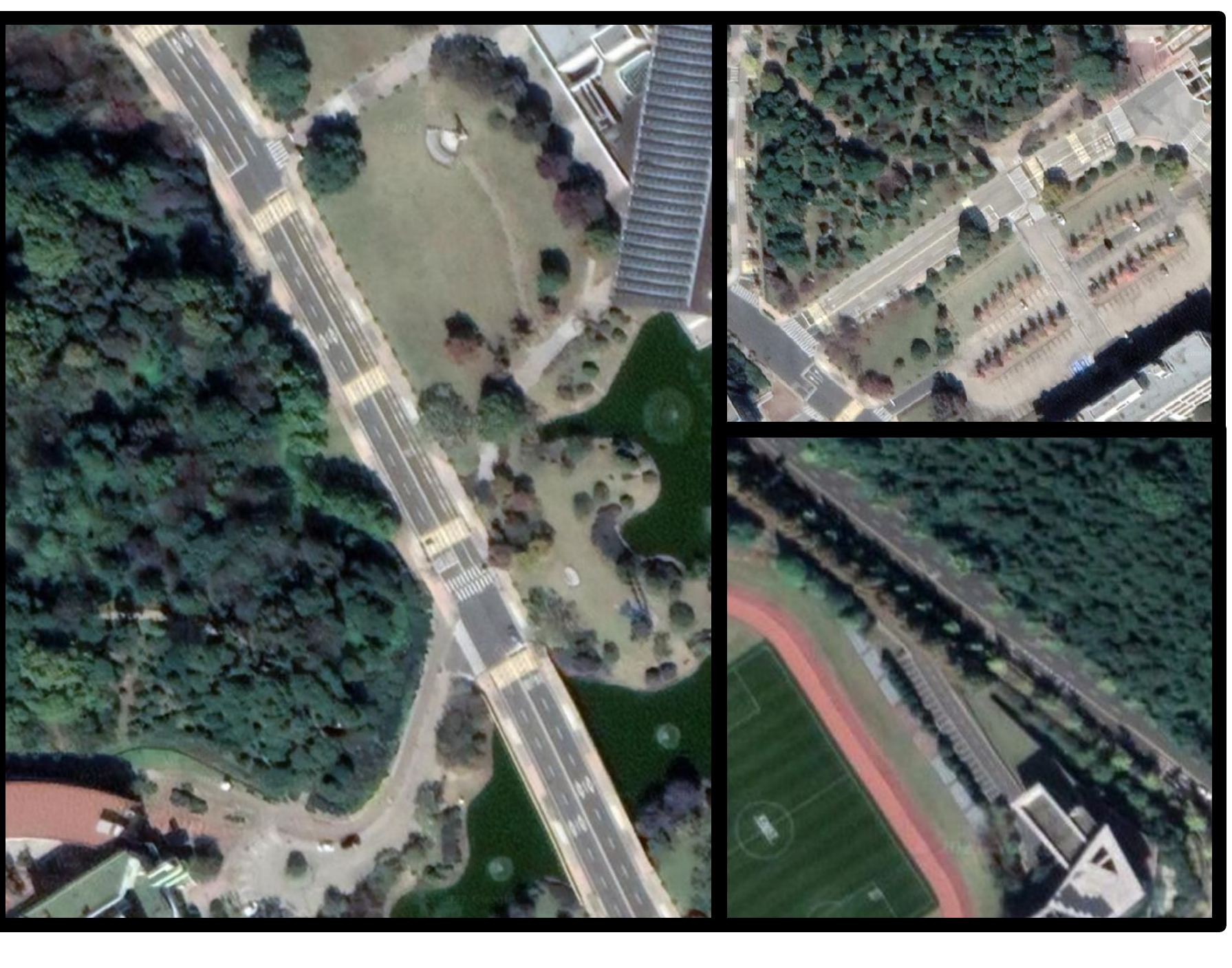}
    \caption{Examples of semi-structured regions in the MulRan dataset. }
    \label{fig:Mulran_overview}
\end{figure}

Currently, there exist four public datasets for spinning radar localization research: Boreas,\footnote{Boreas Autonomous Driving Dataset was accessed on 2022-01-10 from \url{https://registry.opendata.aws/boreas}.}
Radiate~\cite{sheeny2020radiate}, Mulran~\cite{gskim-2020-mulran} and the most established Oxford Radar RobotCar dataset~\cite{RadarRobotCarDatasetICRA2020}. We selected Oxford (Fig.~\ref{fig:Oxford_overview}) and Mulran (Fig.~\ref{fig:Mulran_overview}) which both have accurate ground truth positioning and similar sensor range resolution ($0.0432$~m and $0.0595$~m respectively). Additionally, MulRan contains a diverse variety of surrounding environment types including urban, mountain, and fields which allows us to test how the methods generalize outside of urban environments. In Sec.~\ref{sec:generalization_radar} we used the sequence ``10-12-32'' from Oxford and ``KAIST02'' from Mulran. In all other experiments, each data point and deviation is computed over the sequences \textit{10-12-32}, \textit{18-14-14}, \textit{18-14-46} and \textit{18-15-20} from the Oxford dataset. Both datasets contain a variety of weather conditions and traffic from vehicles, pedestrians and bikers. Hence these sequences constitute realistic urban scenarios.

To make the method evaluation more realistic for urban applications, we generate 4 misalignments symmetrically around each aligned data-point: two in the longitudinal directions of the driving direction (forward and backward) where localization uncertainty due to motion and landmarks is generally higher,
and two in the lateral direction (left and right) with lower uncertainty. Training weights and evaluation metrics are balanced accordingly. 

\begin{figure}
    \centering
    \includegraphics[clip,trim={0.6cm 0cm 0cm 0cm},width=\linewidth]{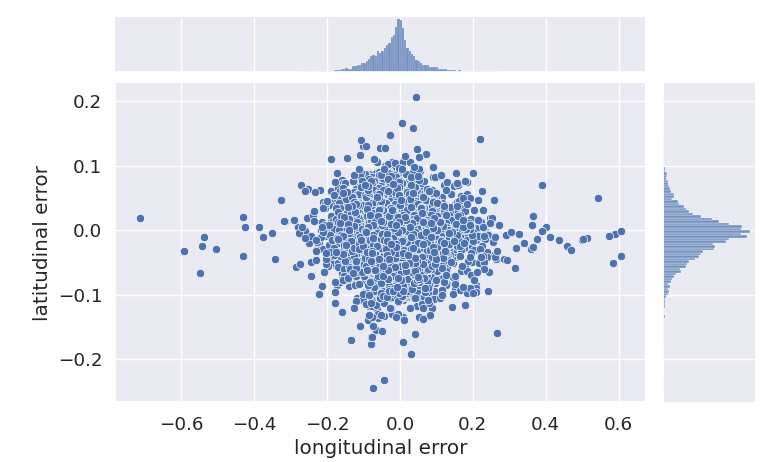}
    \caption{Investigation of error distribution in radar odometry.}
    \label{fig:error_distribution}
\end{figure}

Depending on the application and sensor, various error levels can be of interest. E.g. odometry is expected to be more accurate compared to loop closure and relocalization and requires detection of smaller errors.
We produced the position error distribution of the currently most accurate method for radar odometry estimation \textit{CFEAR}, the distribution is depicted in Fig.~\ref{fig:error_distribution}. 
%Infrequently, the error exceeds $0.3~m$ in the longitudinal direction.
The $0.995$ quantile corresponds to a longitudinal and lateral error of $0.29$~m and $0.12$~m, respectively. Detecting errors above this point $(>~0.29)$~m would be meaningful for the task of odometry. These are larger error levels compared to what we considered in our previous evaluation on lidar ($0.1$~m) in Sec.~\ref{sec:lidar_classification}. %motivate why larger errors make sense
Using the same step size between error levels as in~\cite{Almqvist}, we define small, medium and large errors as $0.3$~m, $0.5$~m and $0.7$~m respectively. These errors are higher than the ones defined for lidar sensors, which is necessary as localization uncertainty is expected to be higher, e.g. due to the larger scale, higher motion and low spinning rate of radar used in our experiments. For a vehicle moving at 50 km/h with a radar spinning at 4~Hz, the motion distortion from translation only is 3.5~m between the first and last segment of a scan unless carefully compensated for.  


\subsection{Evaluated radar methods}
\label{sec:radar-baselines}
Here we give a brief introduction of all compared methods for feature extraction, quality measure and parameters. 
%All feature representations have been 
In all cases, we have
compensated for motion distortion effects as described in~\cite{9636253}, and features within a minimum range of $2.5$~m are removed in order to discard false detections located on the experimental setup itself.

\paragraph{Cen2018} A method for extracting radar features and estimating ego-motion proposed by Cen and Newman~\cite{8460687}.
We use their method for extracting features from intensity gradients and peaks with the improved configuration described by Burnett~\cite{burnett_we_2021}, where the probability threshold is increased to $z_q=3.0$ and a Gaussian filter is added with $\sigma=17$ as described in~\cite{burnett_we_2021}. We do not carry out data association as described in the original publication but instead perform a radius search and compute the point-to-point quality measure with association radius $radius=3$~m. The sum of point-to-point distances is normalized by the number of points and passed to $x1$.
\paragraph{CFEAR} current state-of-the-art in Radar odometry estimation~\cite{9636253}.
We use the CFEAR-feature extraction method, with the quality measures (P2P, P2L and P2D). Radar data is first filtered using $k$-strongest as described in Sec.~\ref{sec:rip-features}
by first applying the $k$-strongest filter and then using a grid-based approach that estimates a set of oriented surface points for each grid cell that contains points. In our experiments, we used the same parameters as described in~\cite{9636253} except with minor changes to $z_{min}=60$ and $radius=3$~m for consistency with Cen2018. For this method, we extend the logistic regression model with a third input dimension $x_3$ to incorporate more available information. Specifically, the absolute score, the number of correspondences and the normalized score are passed to $x_1, x_2$ and $x_3$ respectively.
\paragraph{CorAl-Radar (ours)}
We set $r_\mathrm{min}=r_\mathrm{max}=1$, $\Ereject =0$ and $\epsilon=0$. CorAl-Radar extends the previous parameter CorAl set with window size $w=2$, $k=12$ and $z_{min}=70$ for computing RIP-features (\Cref{sec:rip-features}). The latter parameters ($k$ and $z_{min}$ have the same meaning as in the method CFEAR and are fixed in our experiments and equal for CFEAR and CorAl. The joint and separate entropy are passed separately to the classifier: $x_1=H_{joint},x_2=H_{sep}$ as described in Sec.~\ref{sec:classifier}.

\subsection{Method and performance analysis}
\begin{figure}
    \centering
    \includegraphics[width=\linewidth,trim={1.4cm 0.5cm 2cm 0cm},clip]{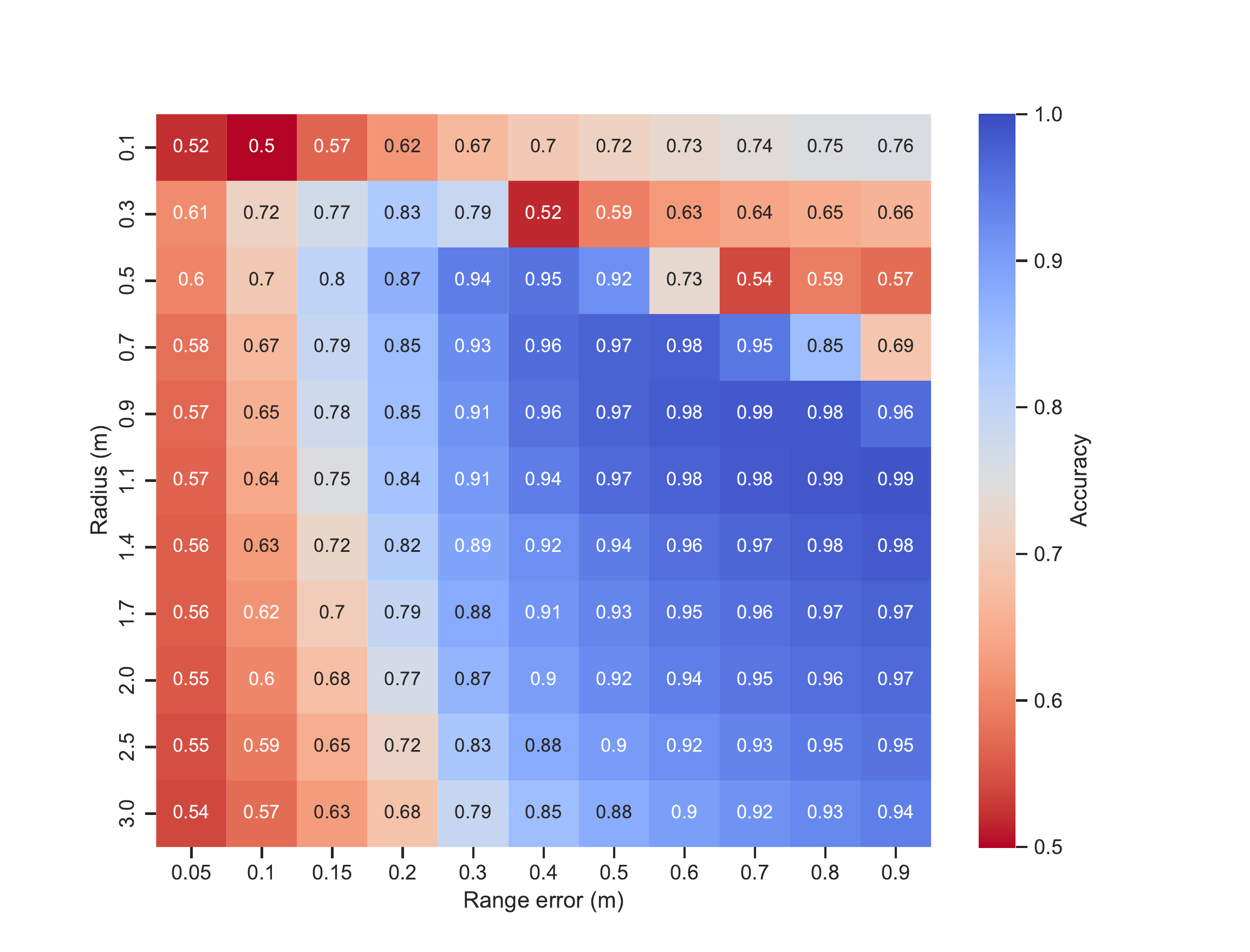}
    \caption{Heat map for the classification accuracy obtained by using the CorAl method, as a function of the translation error and the association radius employed. While the radius is optimally chosen slightly larger compared to the translation maximum error CorAl intends to detect, $r=1$~m provides good capabilities to detect errors larger than $0.2$~m.}
    \label{fig:coral_heatmap}
\end{figure}

We start our evaluation by studying the radius parameter. While a low radius is required to detect the smallest alignment errors, a too conservatively chosen radius will make larger errors more challenging to detect. This is because the computation of entropy is only sensitive to the displacement of points within the radius. Hence is the capability to detect errors with different magnitudes is related to the selection of radius as seen in Fig.~\ref{fig:coral_heatmap}. Generally, the radius should be chosen larger than the maximum error to detect. however as confirmed by our evaluation, detecting very large alignment errors $>1$~m is not a challenging task and CorAl can therefore be complemented with a quality metric such as point-to-point or point-to-line. We found that a radius set to $r=1$~m provides a good trade-off and allows the detection of fairly large errors.

The runtime performance of CorAl is shown in Tab.~\ref{tab:executionTime_table}. All timing statistics have been computed out on an Intel i7-11850H laptop CPU. All representations and quality metrics are efficient to compute ($<$5~ms single-threaded) except Cen2018 which runs at 14~ms with a multi-thread implementation.

\begin{table}
    \centering
\begin{adjustbox}{width=0.99\hsize}
\begin{tabular}{l|lllll}
& \multicolumn{2}{c}{\textbf{Feature extraction}} & \multicolumn{2}{c}{\textbf{Quality measure}} &\\
\textbf{Method} & \textbf{Mean [ms]} & \textbf{Std. dev.} ($\sigma$) & \textbf{Mean [ms]} & \textbf{Std. dev.} ($\sigma$) & Threads\\
\hline
CFEAR-P2P & 3.93 & 0.10 & \textbf{0.69} & 0.04 & single \\
CFEAR-P2L & 4.88 & 0.08 & 0.83 & 0.09 & single\\
CFEAR-P2D & 3.92 & 0.08 & 0.75 & 0.05& single\\
CEN2018-P2P & 14.21 & 0.25 & 3.05 & 0.41 & multiple\\
CorAl (ours) & \textbf{2.07} & 0.03 & 3.47 & 0.29 & single\\
%CFEAR-P2P & 3.9327 & 0.0975 & \textbf{0.6915} & 0.0349 & single \\
%CFEAR-P2L & 4.8832 & 0.0824 & 0.8312 & 0.0936 & single\\
%CFEAR-P2D & 3.9199 & 0.0767 & 0.7521 & 0.0544& single\\
%CEN2018-P2P & 14.2104 & 0.2512 & 3.0448 & 0.4148 & multiple\\
%CorAl (ours) & \textbf{2.0742} & 0.0333 & 3.4704 & 0.2944 & single\\
\end{tabular}
\end{adjustbox}
\caption{Computation time (milliseconds) for feature extraction and quality measure.}\label{tab:executionTime_table}

\end{table}

\subsection{Detecting errors with various magnitudes}
\begin{figure}
\centering
\subfloat[][Accuracy vs translation error.]{\includegraphics[clip,trim={1.4cm 0.5cm 2cm 0cm},width=\linewidth]{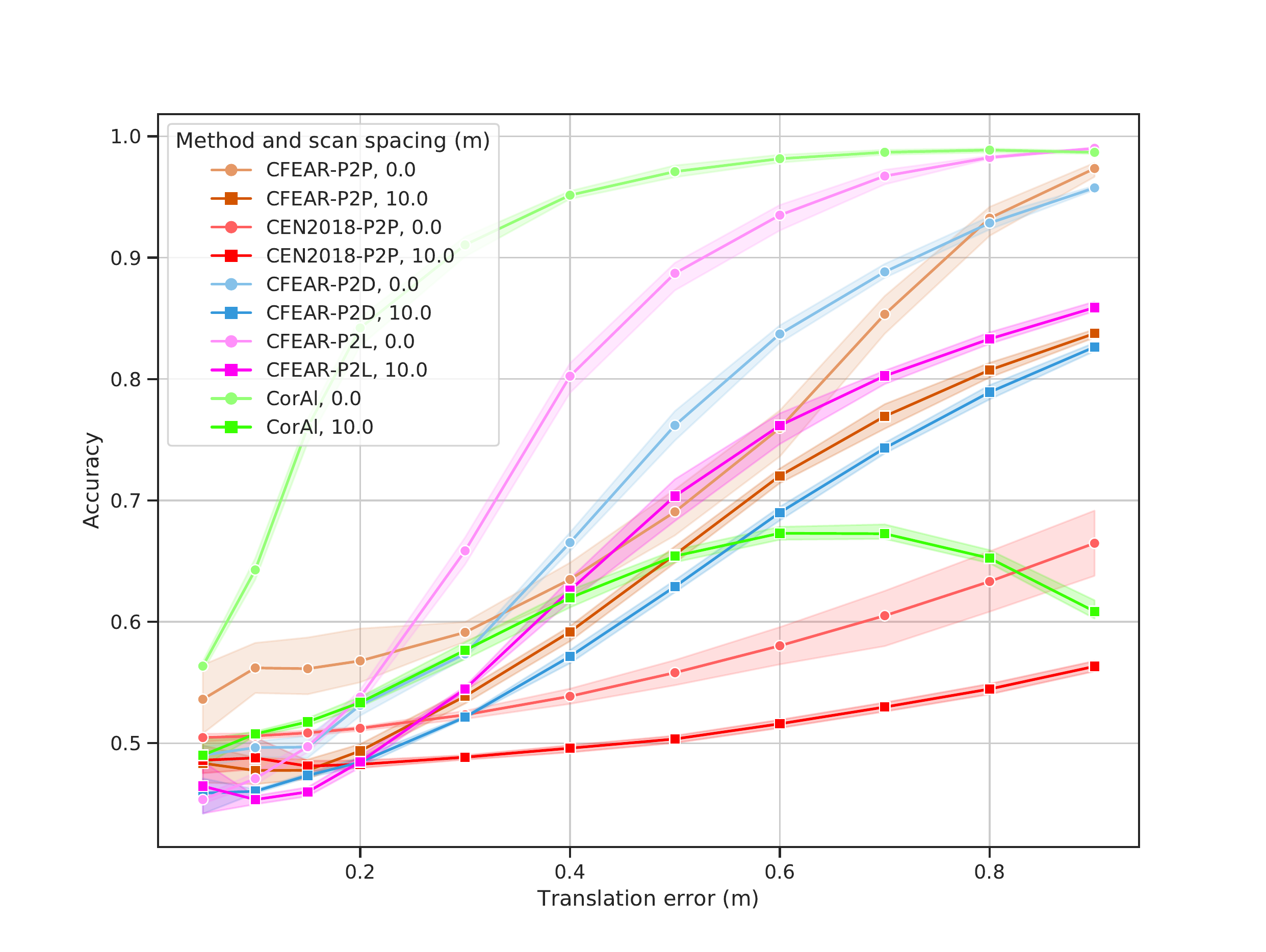}\label{fig:range_error_acc}}\hfill
\vspace{-0.2cm}
\subfloat[][AUC vs translation error.]{\includegraphics[clip,trim={1.4cm 0.5cm 2cm 2cm},width=\linewidth]{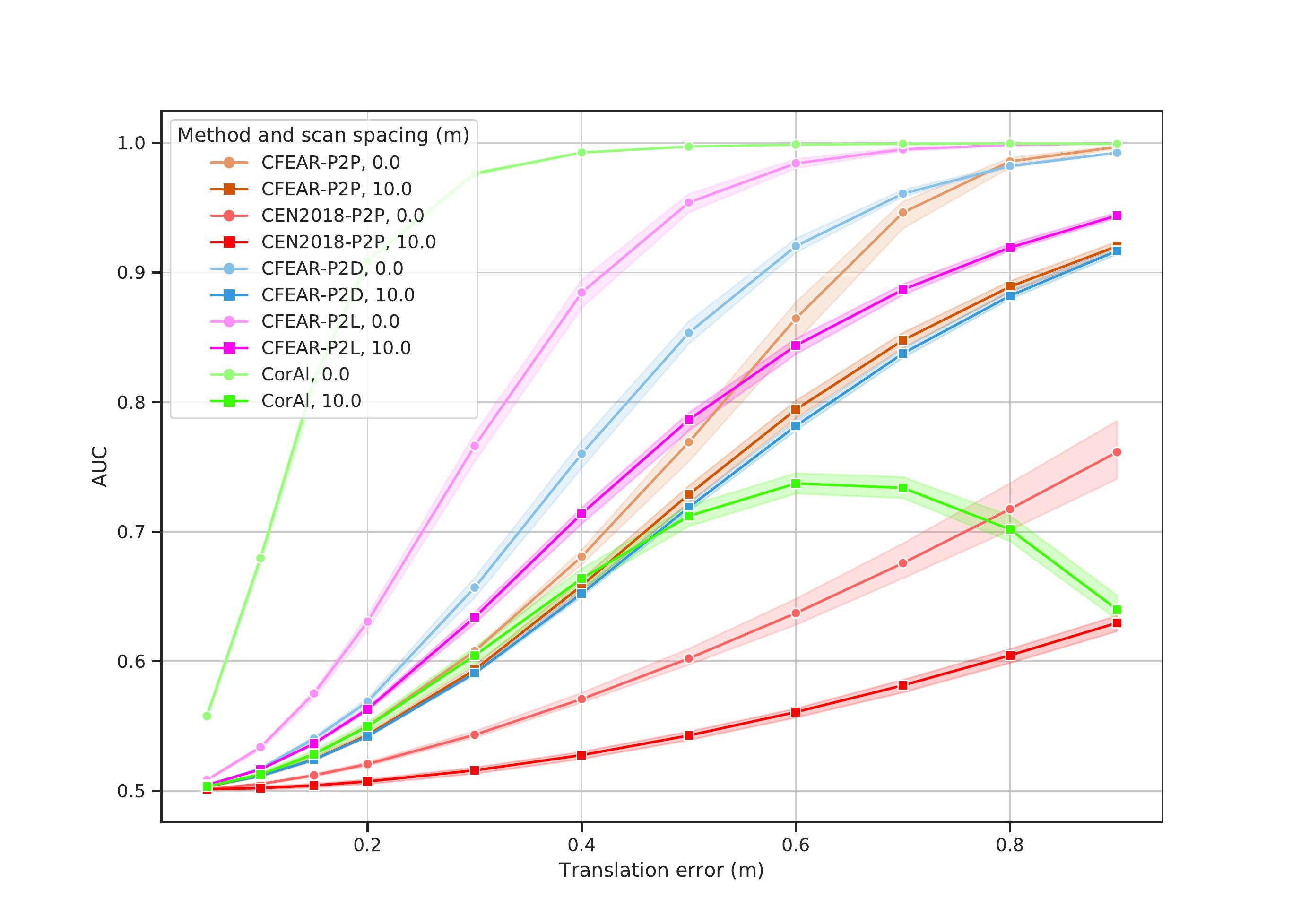}\label{fig:range_error_auc}}
\caption{\label{fig:accuracy_auc_range}Accuracy and AUC vs translation error. (a) CorAl is the most accurate and can even detect small alignment errors of $0.3$~m with $>90\%$ accuracy when distance between scans is low. CFEAR-P2L requires $0.2$~m larger errors to reach similar level of accuracy. For large scan distances (less overlap), CFEAR-P2L is the most accurate with $~85\%$ accuracy. (b) CorAl generally achieves the best performance (well above $90\%$ even for small translation errors) when distance between scans is low. For large scan distances, CFEAR-P2L is the most robust method for any translation error.}
\vspace{-0.3cm}
\end{figure}

%We have performed a set of experiments in which translation (or range) errors are artificially added to subsequent scans in a sequence. 
We aim to investigate the extent to which errors of different magnitudes ranging from $0.05$ up to $0.9$ meters can be detected. 
The performance is quantified using accuracy and the area under the ROC curve (AUC). The obtained results are depicted in Fig.~\ref{fig:accuracy_auc_range}. 
In these plots, we report results for two classes of scans, with varying distances between scans: scans taken at least 0~m apart (consecutive scans) and scans taken at least 10~m apart. %Also, as a reference, we report errors with two minimum distances ($0$ and $10$~m) between scans.
In general, these results show that the proposed CorAl method achieves the best classification accuracy and AUC in the evaluated range when the spacing between scans is low. Under these conditions, none of the other methods reach similar performance. The next-best method is CFEAR-P2L, which robustly detects large translation errors but only when they are greater than 0.7 meters. When scan spacing is large ($10$~m), CorAl is still the most accurate for small errors (less than 0.4 meters) although the accuracy achieved is not very high. A summary of these results can be found in Tab.~\ref{tab:accuracy_AuC_table}. 

\begin{figure}
\centering
\subfloat[][Accuracy vs orientation %theta
errors.]{\includegraphics[clip,trim={0.0cm 0.0cm 0cm 0cm},width=\linewidth]{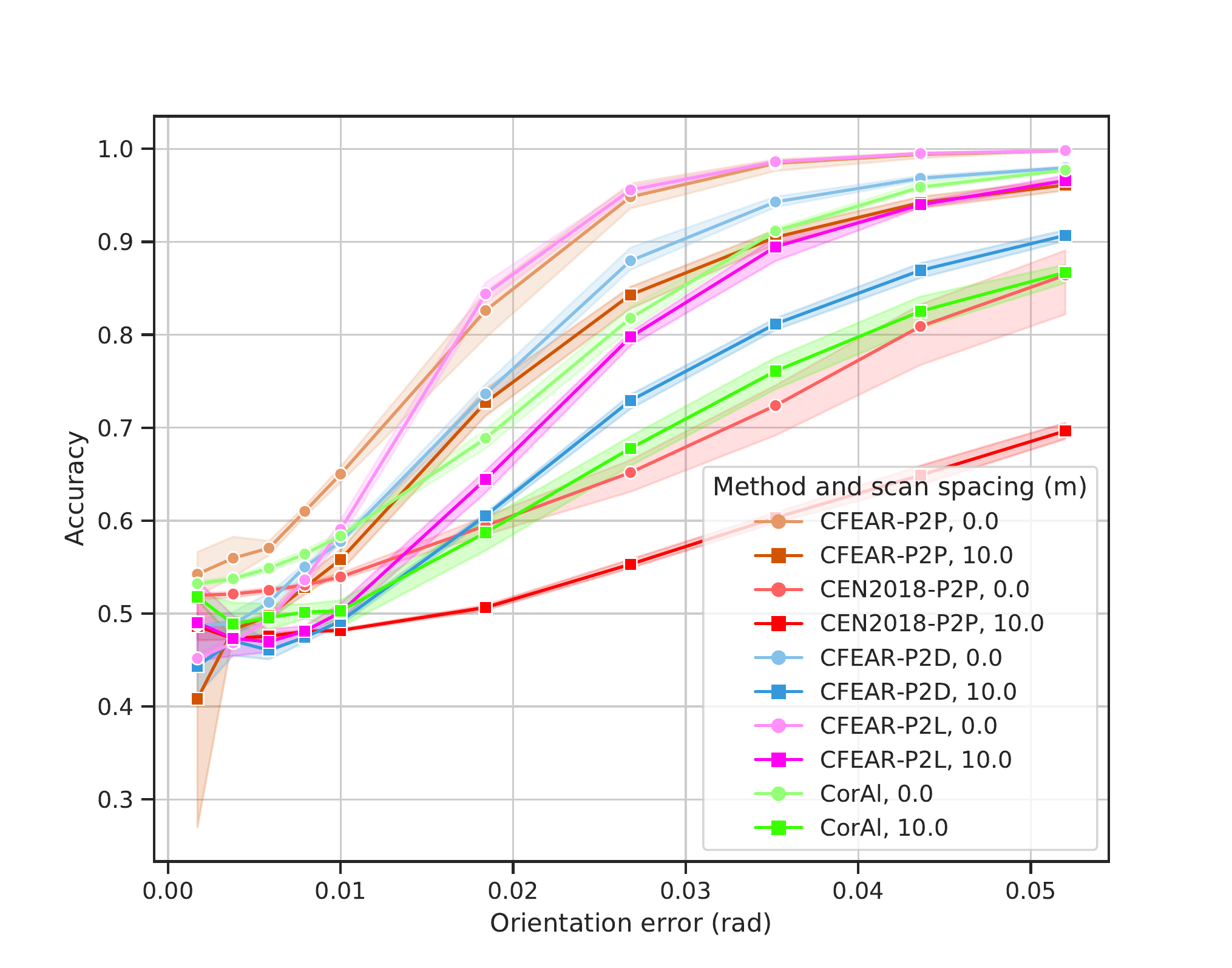}\label{fig:theta_error_acc}}\hfill
\vspace{-0.2cm}
\subfloat[][AUC vs orientation %theta
errors.]{\includegraphics[clip,trim={0cm 0cm 0cm 2cm},width=\linewidth]{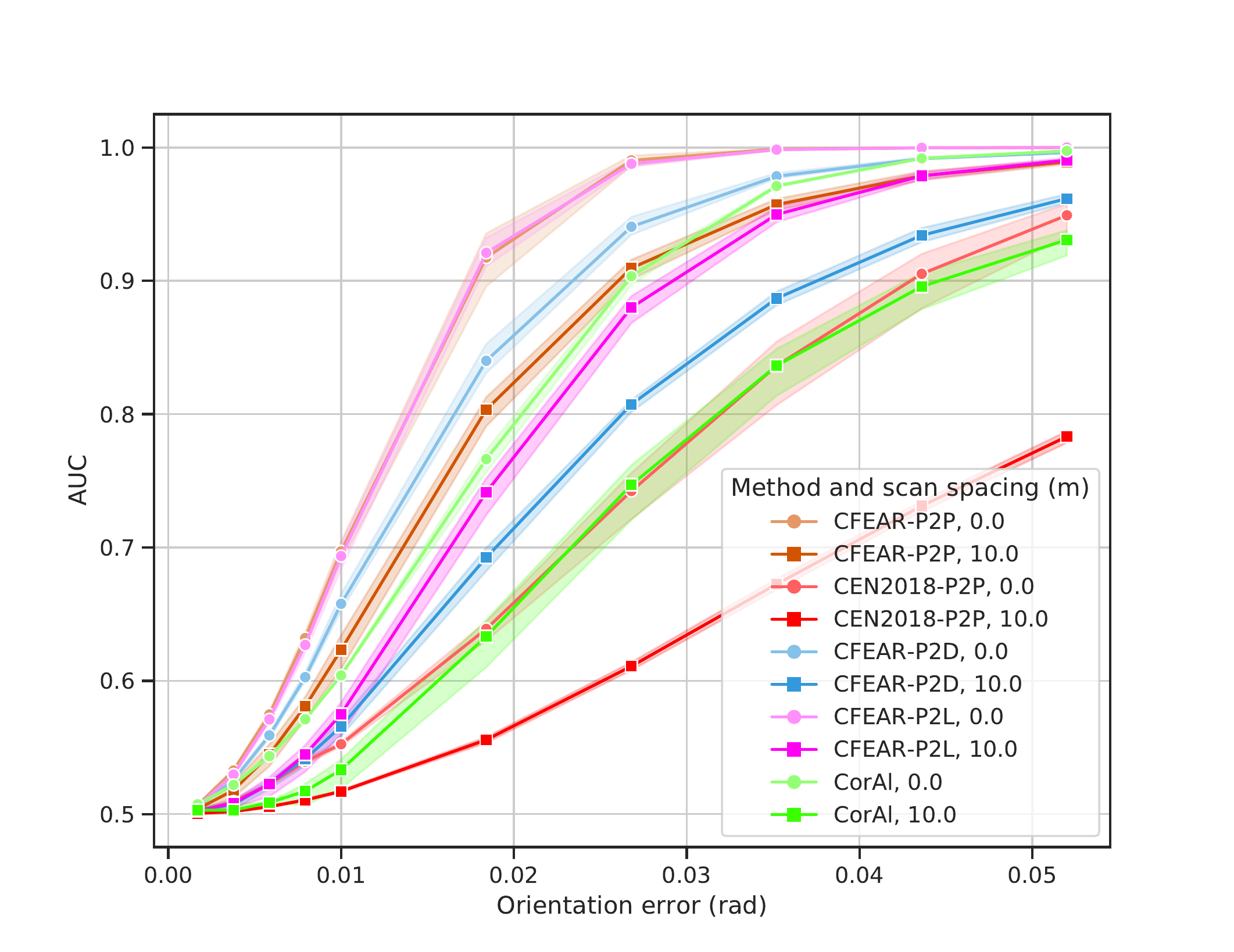}\label{fig:theta_error_auc}}
\caption{\label{fig:theta_error}Classification accuracy vs orientation error. The radius parameter is set to $r=3$~m. }
\vspace{-0.3cm}
\end{figure}

Orientation errors are evaluated separately as depicted in Fig.~\ref{fig:theta_error}. Such errors are added here in the same way as in the case of evaluation on lidar data, i.e., by adding an angular offset $e_{\theta}$ around the sensor's vertical axis. In general, the CFEAR and Cen2018 quality metrics demonstrate similar or improved performance compared to CorAl for orientation errors.
We believe this is because orientation errors displace observations proportionally to the distance of observation. At large distances, the conservative data association of CorAl makes the metric less sensitive, unless combined with the optional parameter that dynamically increases radius accordingly.
\begin{table}
    \centering
\begin{adjustbox}{width=\hsize}
\begin{tabular}{l|cccc}
& \multicolumn{3}{c}{\textbf{Translation error }} 
\\
\textbf{Method} & Small (0.3 m) & Medium (0.5 m) & Large (0.7 m)\\
\hline
CFEAR-P2P & $0.59/0.61$ & $0.69/0.77$ & $0.85/0.95$\\
CFEAR-P2L & $0.66/0.77$ & $0.89/0.95$ & $0.97/ 0.99$\\
CFEAR-P2D & $0.57/0.66$ & $0.76/0.85$ & $0.89/0.96$\\
CEN2018-P2P & $0.52/0.54$ & $0.56/ 0.60$ & $0.61/ 0.68$\\
CorAl (ours) & $\mathbf{0.91}/\mathbf{0.98}$ & $\mathbf{0.97}/\mathbf{1.00}$ & $\mathbf{0.99}/\mathbf{1.00}$\\

%CFEAR-P2P & $0.5913/0.6078$ & $0.6906/0.769$ & $0.8534/0.9462$\\
%CFEAR-P2L & $0.6585/0.7662$ & $0.8871/0.954$ & $0.9672/ 0.9949$\\
%CFEAR-P2D & $0.5735/0.6569$ & $0.7619/0.8534$ & $0.8883/0.9608$\\
%CEN2018-P2P & $0.5233/0.5433$ & $0.5581/ 0.6021$ & $0.6051/ 0.6757$\\
%CorAl (ours) & $\mathbf{0.9106}/\mathbf{0.9763}$ & $\mathbf{0.9709}/\mathbf{0.9971}$ & $\mathbf{0.9868}/\mathbf{0.9992}$\\

%CFEAR-P2P & $0.5913/0.6078$ & $0.6906/0.769$ & %$0.8534/0.9462$\\
%CFEAR-P2L & $0.6585/0.7662$ & $0.8871/0.954$ & %$0.9672/ 0.9949$\\
%CFEAR-P2D & $0.5735/0.6569$ & $0.7619/0.8534$ & %$0.8883/0.9608$\\
%CEN2018-P2P & $0.5233/0.5433$ & $0.5581/ 0.6021$ %& $0.6051/ 0.6757$\\
%CorAl (ours) & $\mathbf{0.9106}/\mathbf{0.9763}$ %& $\mathbf{0.9709}/\mathbf{0.9971}$ & %$\mathbf{0.9868}/\mathbf{0.9992}$\\
\end{tabular}
\end{adjustbox}
\caption{Quality of classification results corresponding to each method used in this work, versus different translation errors. The values reported for each configuration are the mean accuracy and mean AUC, respectively, which also correspond to 0 meters of scan overlapping.}\label{tab:accuracy_AuC_table}

\end{table}
%Regarding AUC, CFEAR-P2L is most accurate, at least when the error is very low. Generally speaking, all the previous results prove that our method is the most robust when dealing with low range errors, which is precisely the focus of this work. A summary of the results is reported in

%~\ref{tab:accuracy_AuC_table} and figure %
%Regarding association radius, we have run all methods with $radius=3$ except CorAl, which uses $radius=1$. [EXPLAIN WHY].

%\begin{figure}
%    \centering
%    \includegraphics[clip,trim={1.4cm 0.5cm 2cm 0cm},width=\linewidth]{Journal/figures/radar_eval_methods/accuracyVSrange_error.pdf}
%        \caption{Accuracy vs range errors. CorAl is the most accurate and can even detect small alignment errors of $0.3$~m with $>90\%$ accuracy when distance between scans is low. CFEAR-P2L requires $0.2$~m larger errors to reach similar level of accuracy. For large scan distances (less overlap), CFEAR-P2L is the most accurate with $~85\%$ accuracy. }%wow nice! :) Ready!! cool!, repeat for all figures :D?, yes, i did
%    \label{fig:accuracy_range}
%\end{figure}

%\begin{figure}
%    \centering
%    \includegraphics[clip,trim={1.4cm 0.5cm 2cm 2cm},width=\linewidth]{Journal/figures/radar_eval_methods/aucVSrange_error.pdf}
%    \caption{AUC vs range errors. CorAl generally achieves the best performance (well above $90\%$ even for small range errors) when distance between scans is low. For large scan distances, CFEAR-P2L is the most robust method for any range error.}
%    \label{fig:auc_range}
%\end{figure}

%--------------
%------------------

%\begin{figure}
    %\centering
    %\includegraphics[clip,trim={0.8cm 0.2cm 1.5cm 0cm},width=\linewidth]{Journal/figures/radar_eval_methods/accuracy_th_error.pdf}
%    \caption{Accuracy vs theta errors.}
%    \label{fig:my_label}
%\end{figure}

%\begin{figure}
    %\centering
    %\includegraphics[clip,trim={1.0cm 0.5cm 2cm 0cm},width=\linewidth]{Journal/figures/radar_eval_methods/auc_th_error.pdf}
%    \caption{AUC vs theta errors.}
%    \label{fig:my_label2}
%\end{figure}

\subsection{Variation in distance between scans}
\begin{figure}
\centering
\subfloat[][Closely located radar scans.]{\includegraphics[clip,trim={27.0cm 1.5cm 22cm 0cm},width=0.48\linewidth]{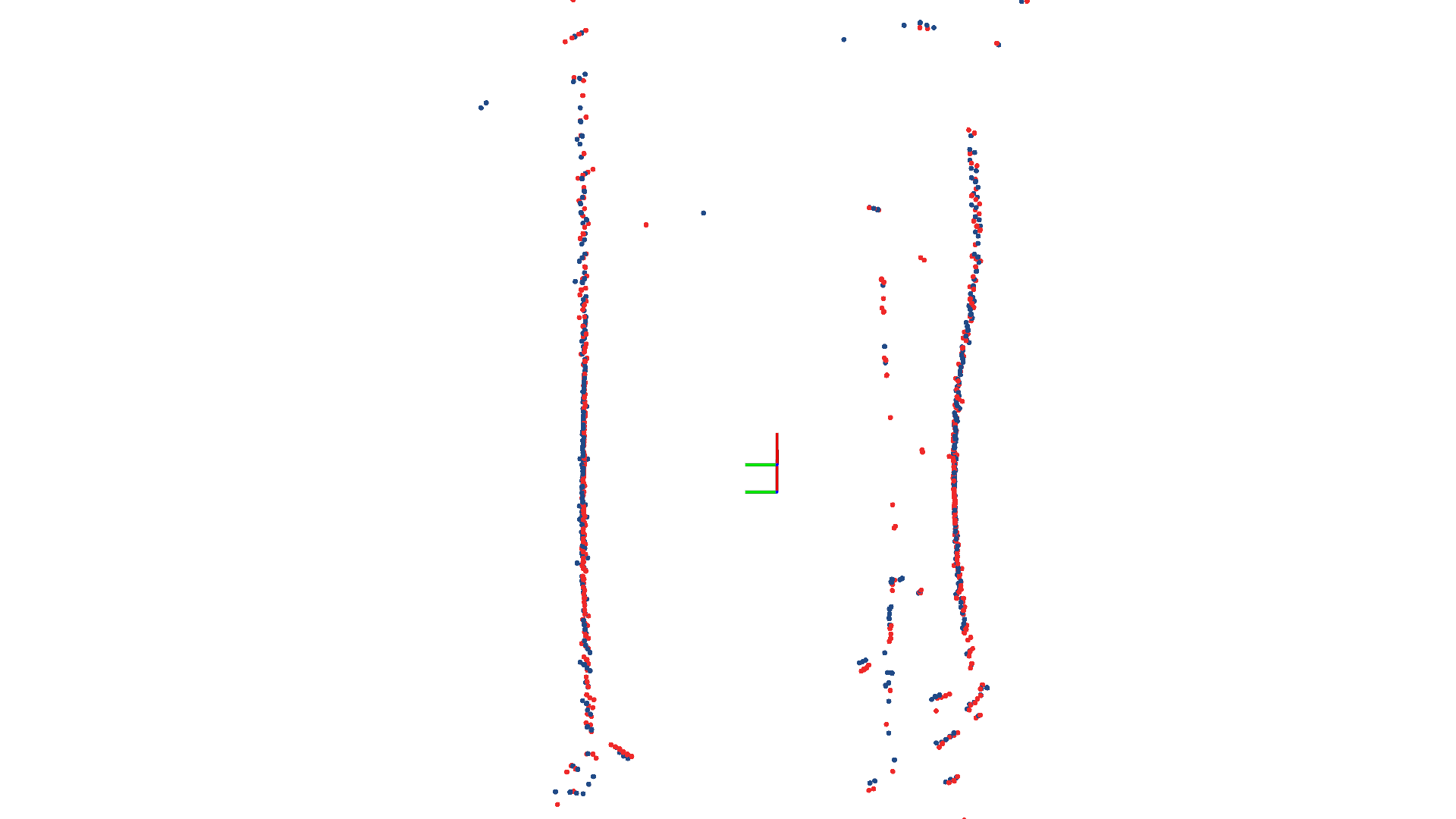}\label{fig:qualitative_radar_overlap_high}}\hfill
\subfloat[][Distant located radar scans.]{\includegraphics[clip,trim={26.5cm 0cm 22.2cm 0cm},width=0.48\linewidth]{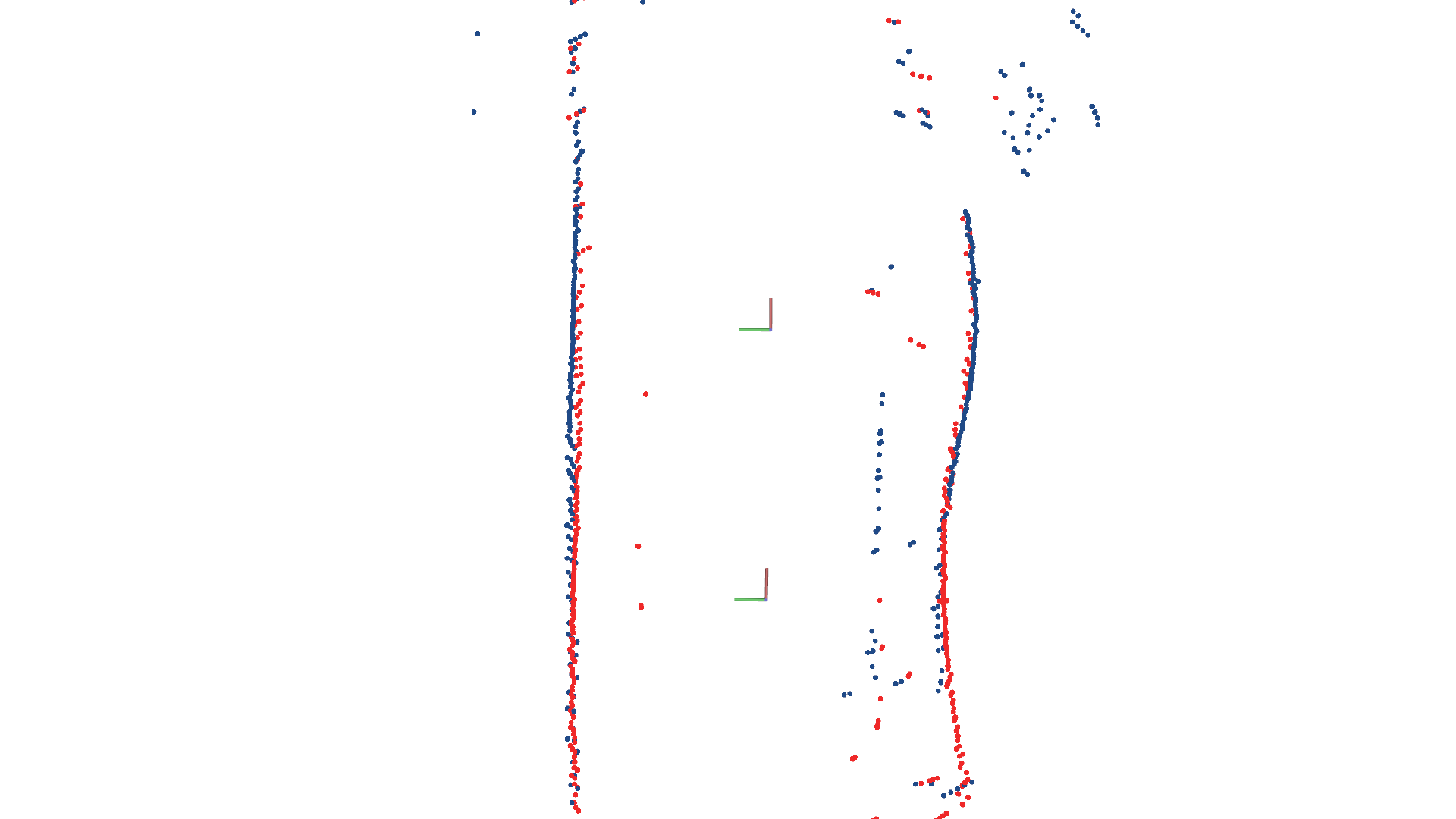}\label{fig:qualitative_radar_overlap_low}}
\caption{\label{fig:qualitative_radar_overlap}Bottom (red) and top (blue) scans acquired with small (a) and large (b) separation in distance. When distance is high, scans incorrectly appear to be misaligned due to a high level of beam divergence in current radar sensors. The impact is higher when landmarks are observed from different angles, which occur more often when the distance between scans is large.
}
\vspace{-0.3cm}
\end{figure}

\begin{figure}
\centering
\subfloat[][Accuracy vs scan spacing distance.]{    \includegraphics[clip,trim={1.4cm 0.5cm 2cm 1cm},width=0.7\linewidth]{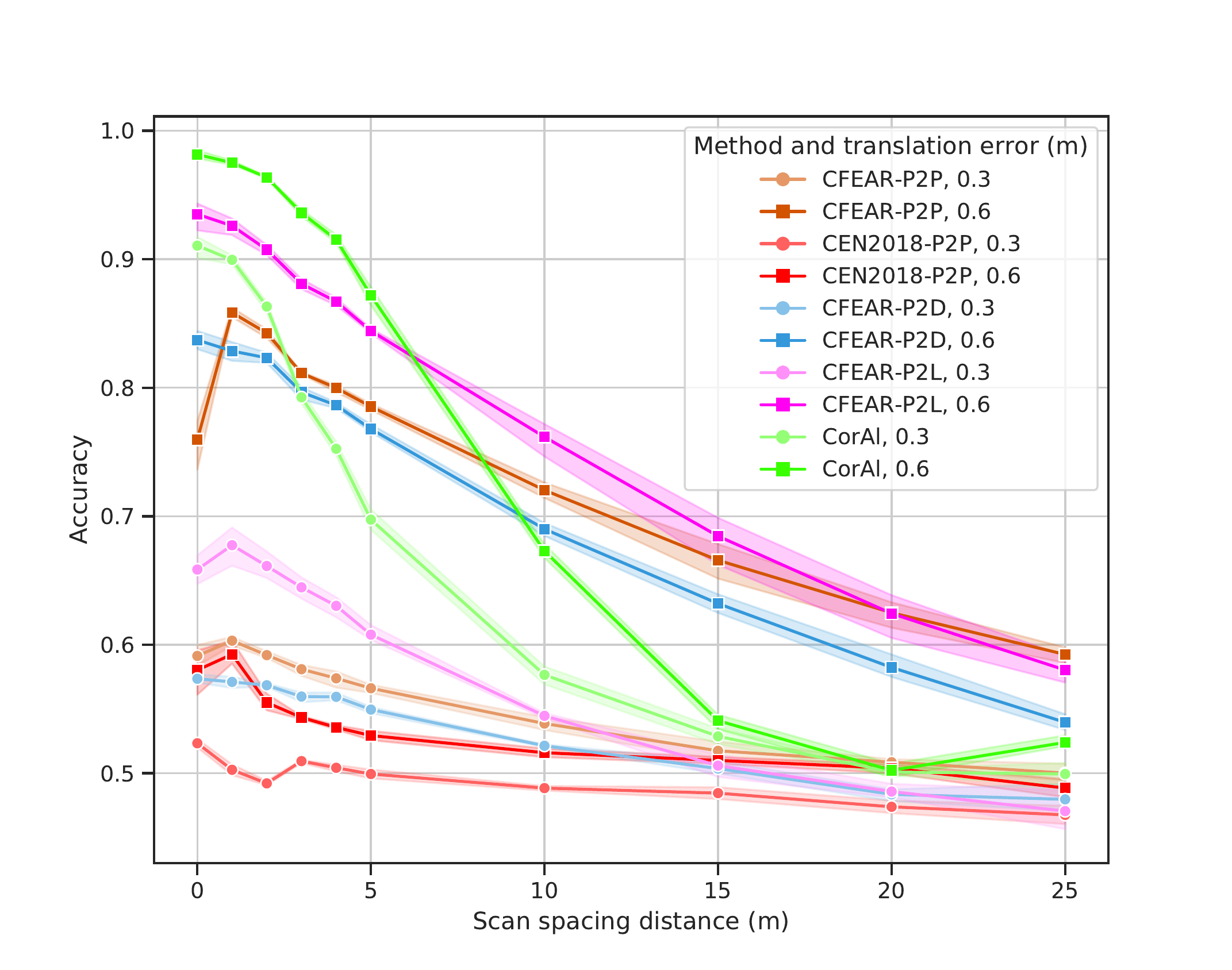}\label{fig:accuracy_scan_space}}\hfill
%\vspace{-0.2cm}
\subfloat[][AUC vs scan spacing distance.]{ \includegraphics[clip,trim={1.4cm 0.5cm 2cm 1cm},width=0.7\linewidth]{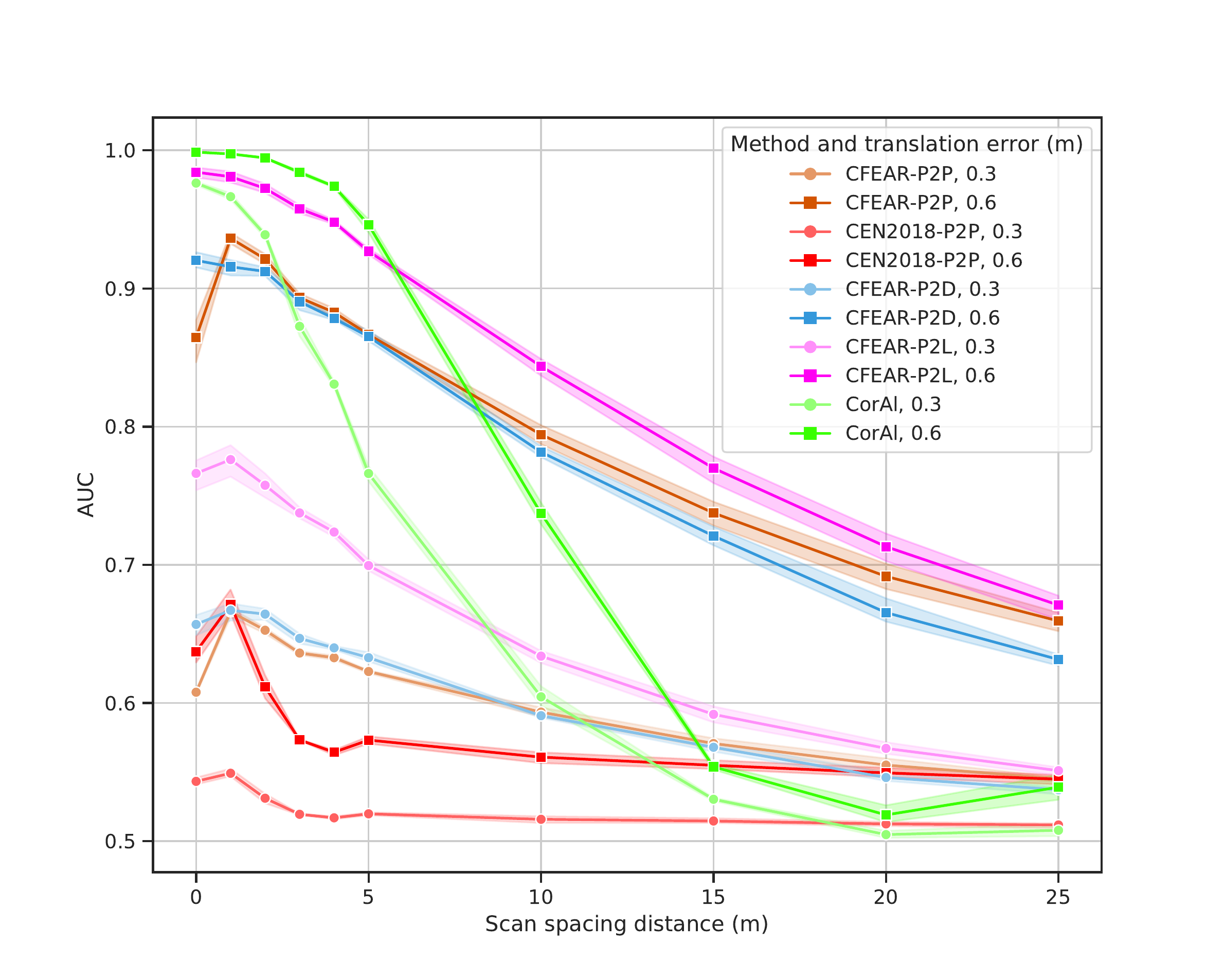}\label{fig:auc_scan_space}}
\caption{\label{fig:scan_space} Accuracy and AUC vs scan spacing (overlap). Increasing distance between scans makes classification more challenging. CorAl is most accurate for small errors while CFEAR-P2L is more accurate when distance between scans is higher.}
\vspace{-0.3cm}
\end{figure}

We have also carried out another set of experiments aimed at analyzing the impact of scan spacing distance. 
%However, in this case we study performance versus scan spacing, which goes from 0 to 25 meters in these experiments. 
In this case, we consider two different levels of translation errors (0.3 and 0.6 meters). Detecting small errors from scans separated by large distances is challenging due to dynamics changes, lower overlap, and because sensor characteristics make observed landmarks appear differently from different perspectives, a phenomenon previously discussed in the literature~\cite{adolfsson_submap_2019}.
As expected, the results in Fig.~\ref{fig:scan_space} show that classification accuracy is reduced for all methods when scan spacing is increased. Despite this, CorAl was able to achieve $~87\%$ accuracy for $0.6$~m errors at $5$~m spacing. After $7$~m, CFEAR-P2L accuracy surpasses CorAl, which accuracy reduces more quickly We believe the worse performance for larger scan spacing depends on radar-specific challenges as depicted in Fig.~\ref{fig:qualitative_radar_overlap}. When spacing is large, walls appear different due to beam divergence. CorAl is more sensitive to small errors and hence not expected to perform well in this scenario.

\subsection{Generalization across environments}
\label{sec:generalization_radar}
\begin{figure}
    \centering
    \subfloat[][Classification results on Oxford dataset when training on the same dataset (intra dataset), and Mulran (generalization).
]{\includegraphics[trim={8cm 0cm 0cm 2cm},clip,width=\linewidth]{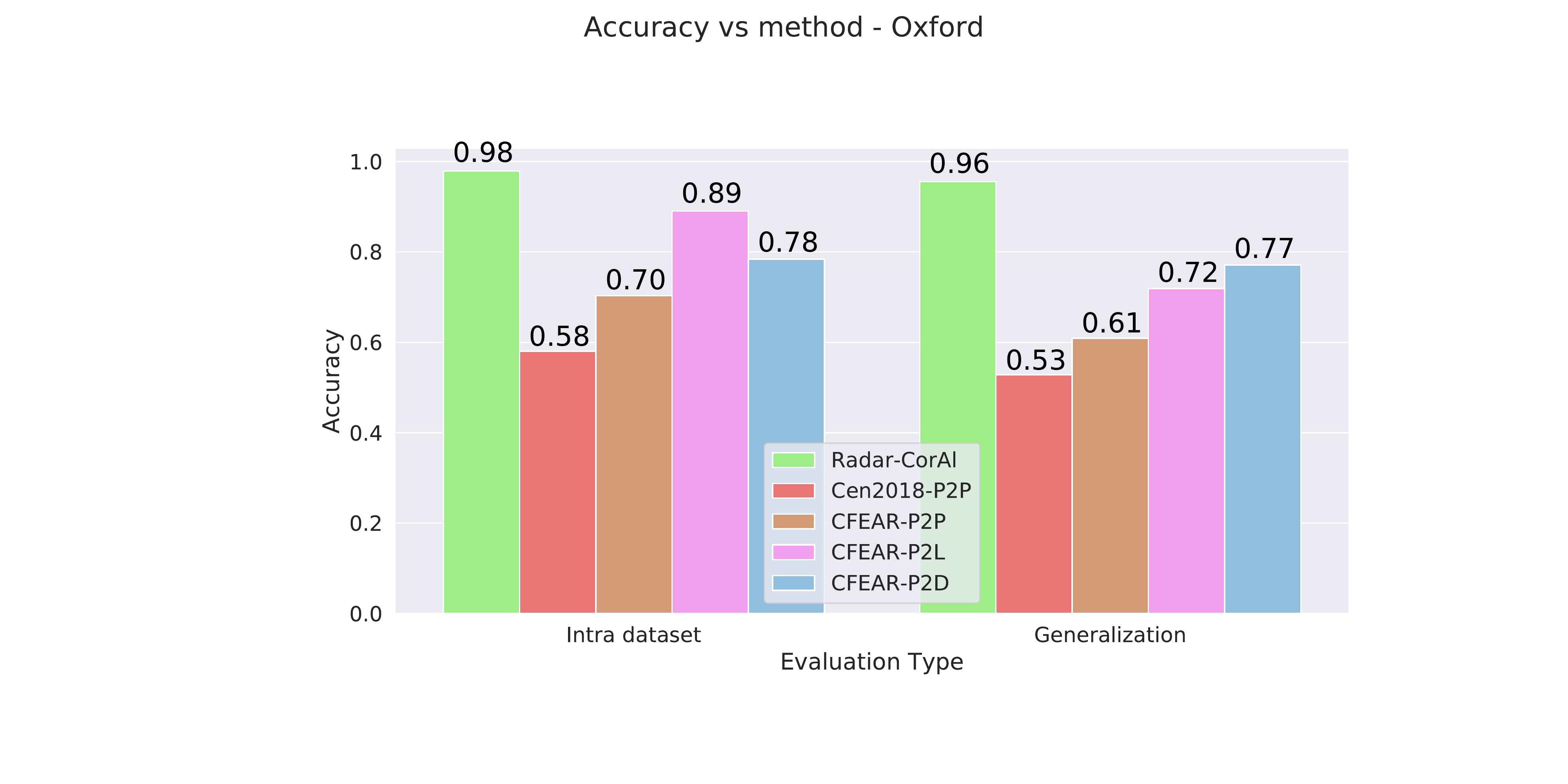}\label{fig:accuracy_oxford_generalzation}}\hfill\\
\subfloat[][Classification results on Mulran dataset when training on left): the same dataset (intra dataset), and right): Oxford (generalization).
]{\includegraphics[trim={8cm 0cm 0cm 2cm},clip,width=\linewidth]{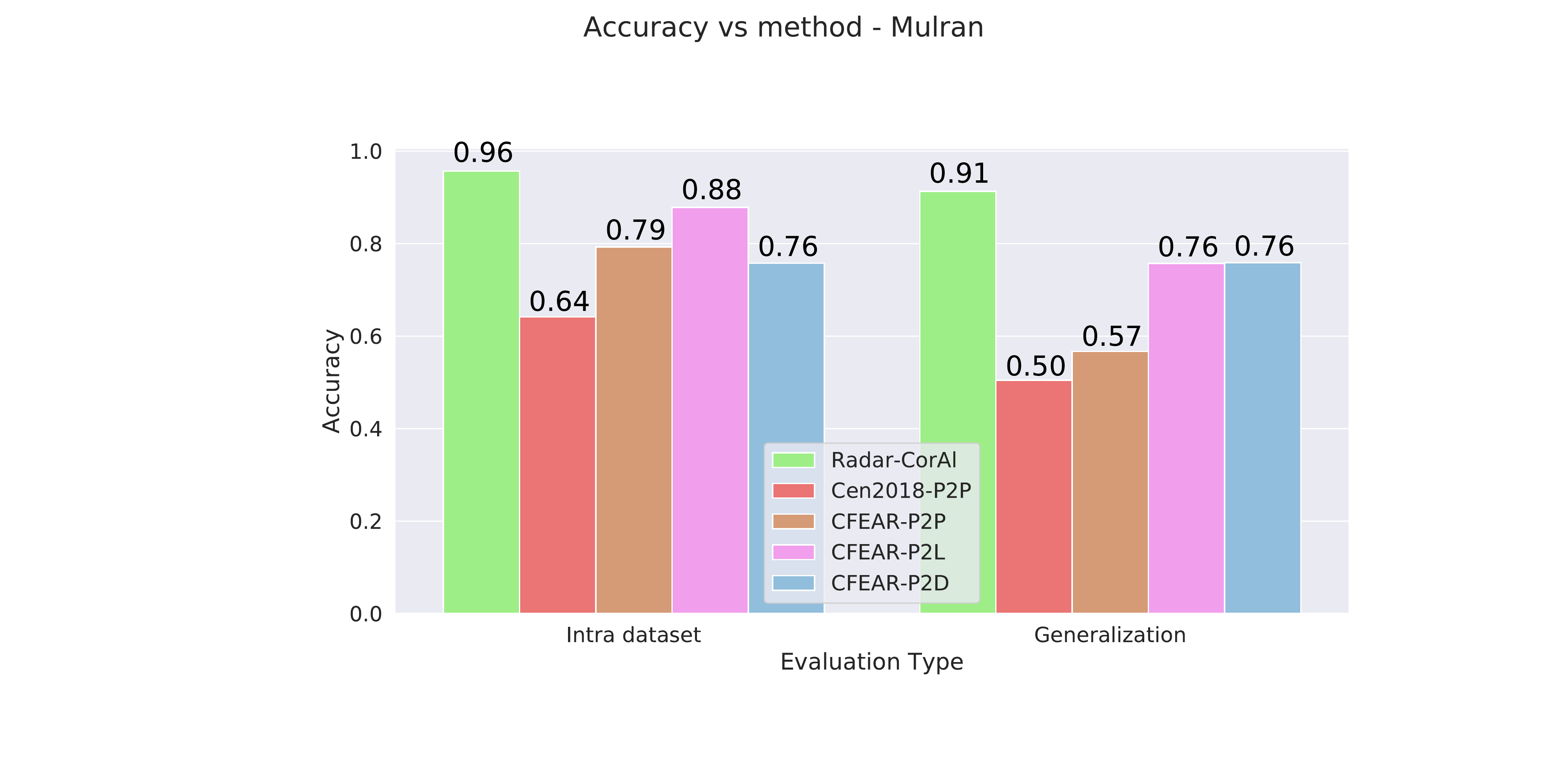}\label{fig:accuracy_mulran_generalzation}}
\caption{\label{fig:accuracy_generalization}Classification accuracy of small errors ($0.5$~m) for a fixed discrimination threshold ($t_h=0.5$) when training and evaluating on the same dataset (intra dataset) and when training and evaluating on different datasets (generalization).}
\end{figure}

%To evaluate the generalization capabilities on radar data, two data sets were utilized; the Oxford and the MulRan dataset where the Oxford dataset contains mainly urban areas (structured) and MulRan contains a variety of different environments (semi-structured), see Fig.~\ref{fig:Mulran_overview}. 
Similarly to the evaluation of generalization capabilities for lidar data (\Cref{sec:lidar-generalization}), we are interested in comparing how the proposed approach can classify the scans when the training and test data sequences are not from the same type of environment. We used the urban Oxford dataset (Fig.~\ref{fig:Oxford_overview}) and the partly semi-structured Mulran dataset (Fig.~\ref{fig:Mulran_overview}).
%To complement previous experiments, we reduce radius of baselines to $radius=1$~m. 
The accuracy for Oxford and Mulran is depicted in Fig.~\ref{fig:accuracy_generalization}. We make the same observation as in the lidar generalization experiments in Sec.~\ref{sec:eval}; classification of a method trained from a semi-structured (more diverse) data set will generalize better. 

\begin{figure}
\centering
\subfloat[][ROC curves when evaluating on the Oxford dataset while using Oxford (Intra dataset) or Mulran (Generalization) for training.
]{\includegraphics[clip,trim={0.5cm 0cm 2cm 0cm},width=0.99\linewidth]{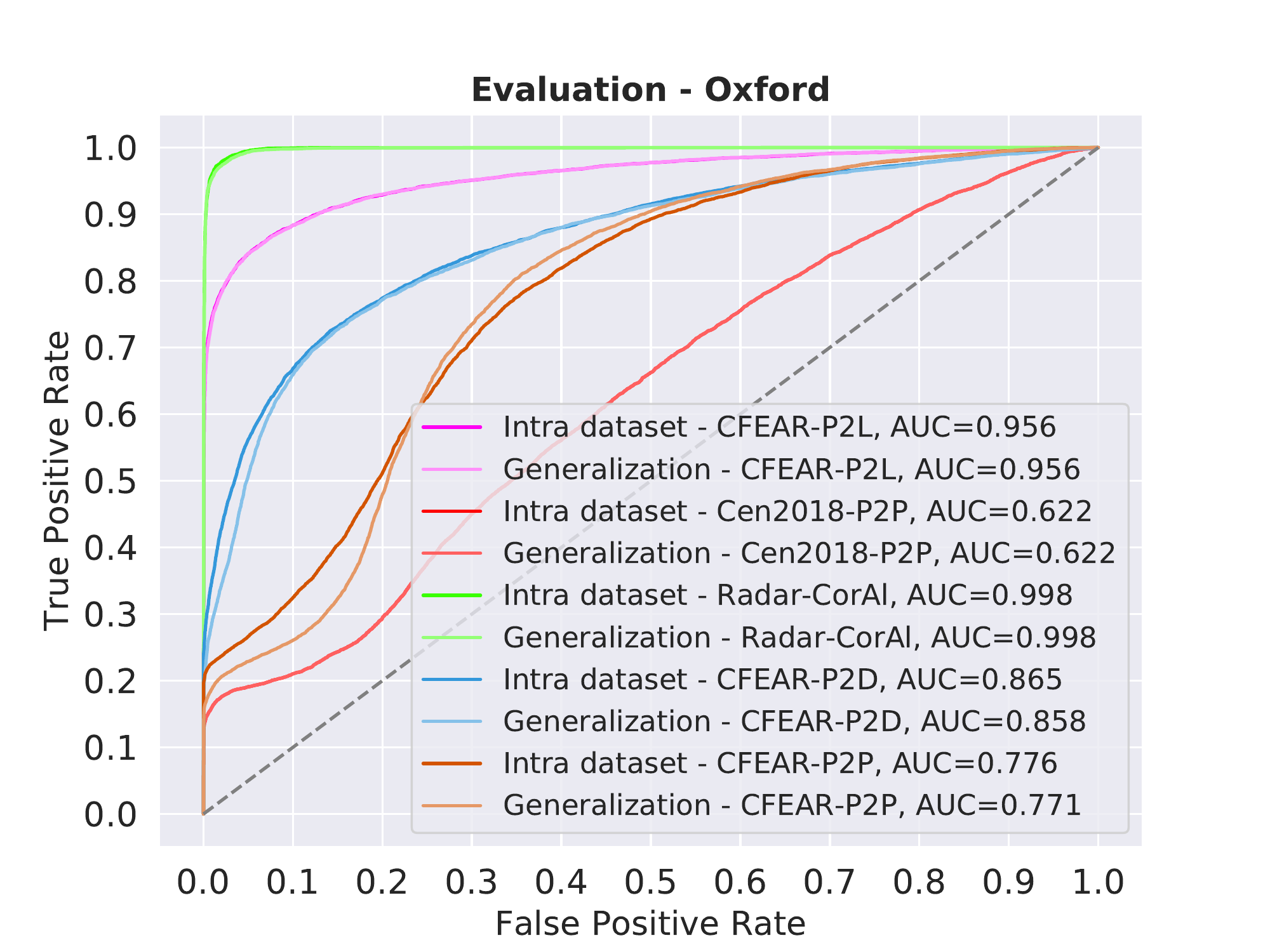}\label{fig:rocl_oxford_generalzation}}\hfill\\
\subfloat[][ROC curves when evaluating on the MulRan dataset while using MulRan (Intra dataset) or Oxford (Generalization) for training.
]{\includegraphics[clip,trim={0.5cm 0cm 1.8cm 0cm},width=0.99\linewidth]{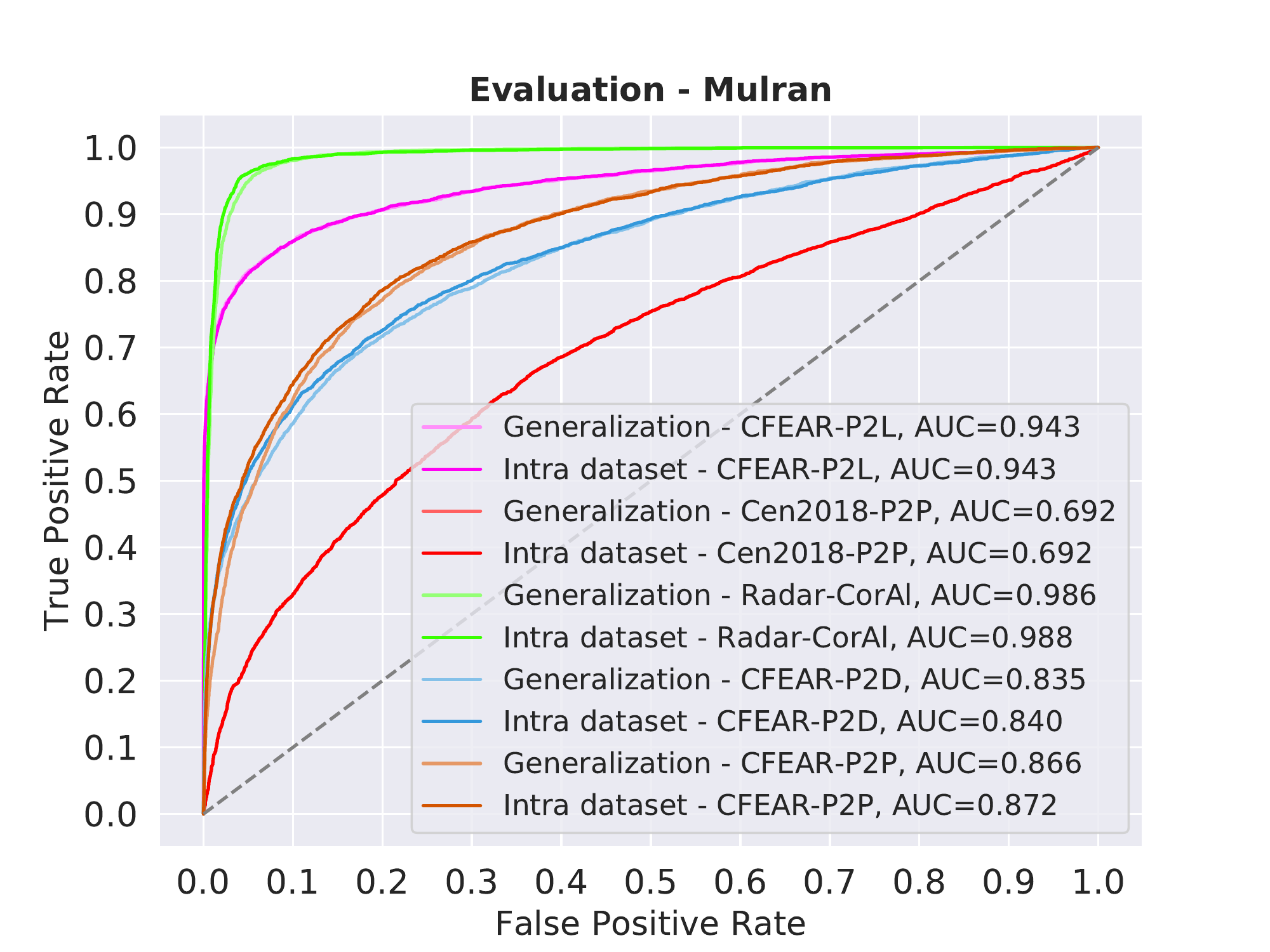}\label{fig:roc_mulran_generalzation}}
\caption{\label{fig:roc_generalization} Comparison where methods are trained and tested on the same (Intra dataset) and separate (Generalization) datasets with varied discrimination treshold. We found that CorAl is the most accurate in both environments regardless of where the method is trained. For our proposed method, the best level of generalization is achieved when training on the semi-structured dataset MulRan and testing on the structured Urban Oxford dataset. }
\vspace{-0.3cm}
\end{figure}

To train with the structured data (Oxford) and testing on the semi-structured data (Mulran) the accuracy obtained is 91\% compared with 96\% if the testing and training data is switched. If the datasets are considered fully separately the accuracy for Oxford and Mulran is 97.9\% and 95.7\% respectively. ROC curves are provided in Fig.~\ref{fig:roc_generalization}, which also illustrates the generalization capabilities. 

\section{Conclusions}
In this paper, we presented CorAl, a principled and intuitive quality measure and self-supervised system that learns to detect small alignment errors between pairs of previously aligned point clouds. CorAl uses dual entropy measurements found in the separate point and in the joint point cloud to obtain a quality measure that substantially outperforms previous methods on the task of detecting small alignment errors within a benchmarking lidar dataset, and within a large-scale urban dataset for spinning radar.

In this work, we proposed a two-step filtering strategy that operates on challenging radar data and produces a high-quality point cloud. By combining our filtering method with CorAl we were able to detect small alignment errors in urban settings using only a spinning radar. We found the method to be accurate in a wide range of environments and can generalize to new unseen environments without retraining. Using a roof-mounted radar within realistic trafficked urban scenarios, we achieve up to $98\%$ accuracy in detection of $0.5$~m errors when trained in the same environment, and up to $96\%$ accuracy when trained on another environment type.
Our experiments on both lidar and radar data demonstrate that CorAl achieves a high level of generalization between structured and semi-structured environments. We also found that learning from more challenging less structured environments results is advantageous for generalization. In our lidar experiments, we even found that CorAl was able to generalize from unstructured (woods) to structured indoor environments. However, none of the evaluated methods generalized well when trained in structured environments only and evaluated in an unstructured environment, and this remains a challenging problem.

% which remains challenging.

We believe that the presented system has great potential to serve as an alignment quality tool for point clouds and can improve localization robustness by equipping odometry, relocalization, and loop closure systems with the capability of introspectively detecting small errors in diverse environments.

%\section*{References}
\bibliography{Journal/references}

\begin{thebibliography}{10}
\providecommand{\url}[1]{#1}
\csname url@rmstyle\endcsname
\providecommand{\newblock}{\relax}
\providecommand{\bibinfo}[2]{#2}
\providecommand\BIBentrySTDinterwordspacing{\spaceskip=0pt\relax}
\providecommand\BIBentryALTinterwordstretchfactor{4}
\providecommand\BIBentryALTinterwordspacing{\spaceskip=\fontdimen2\font plus
\BIBentryALTinterwordstretchfactor\fontdimen3\font minus
  \fontdimen4\font\relax}
\providecommand\BIBforeignlanguage[2]{{%
\expandafter\ifx\csname l@#1\endcsname\relax
\typeout{** WARNING: IEEEtran.bst: No hyphenation pattern has been}%
\typeout{** loaded for the language `#1'. Using the pattern for}%
\typeout{** the default language instead.}%
\else
\language=\csname l@#1\endcsname
\fi
#2}}

\bibitem{9636253}
D.~Adolfsson, M.~Magnusson, A.~Alhashimi, A.~J. Lilienthal, and H.~Andreasson,
  ``{CFEAR} radarodometry - conservative filtering for efficient and accurate
  radar odometry,'' in \emph{2021 IEEE/RSJ International Conference on
  Intelligent Robots and Systems (IROS)}, 2021, pp. 5462--5469.

\bibitem{adolfsson_submap_2019}
D.~Adolfsson, S.~Lowry, M.~Magnusson, A.~Lilienthal, and H.~Andreasson, ``A
  {Submap} per {Perspective} - {Selecting} {Subsets} for {SuPer} {Mapping} that
  {Afford} {Superior} {Localization} {Quality},'' in \emph{2019 {European}
  {Conference} on {Mobile} {Robots} ({ECMR})}, Sept. 2019, pp. 1--7.

\bibitem{della_corte_unified_2019}
B.~Della~Corte, H.~Andreasson, T.~Stoyanov, and G.~Grisetti, ``Unified
  {Motion}-{Based} {Calibration} of {Mobile} {Multi}-{Sensor} {Platforms}
  {With} {Time} {Delay} {Estimation},'' \emph{IEEE Robotics and Automation
  Letters}, vol.~4, no.~2, pp. 902--909, Apr. 2019.

\bibitem{9013051}
A.~C.~M. {Tavares}, F.~J. {Lawin}, and P.~{Forssén}, ``Assessing losses for
  point set registration,'' \emph{IEEE Robotics and Automation Letters},
  vol.~5, no.~2, pp. 3360--3367, 2020.

\bibitem{zhang_loam_2014}
J.~Zhang and S.~Singh, ``{LOAM}: Lidar odometry and mapping in real-time,'' in
  \emph{Robotics: Science and Systems}, 2014.

\bibitem{gcnv2}
J.~Tang, L.~Ericson, J.~Folkesson, and P.~Jensfelt, ``Gcnv2: Efficient
  correspondence prediction for real-time slam,'' \emph{IEEE Robotics and
  Automation Letters}, vol.~4, no.~4, pp. 3505--3512, 2019.

\bibitem{8462890}
S.~{Nobili}, G.~{Tinchev}, and M.~{Fallon}, ``Predicting alignment risk to
  prevent localization failure,'' in \emph{2018 IEEE International Conference
  on Robotics and Automation (ICRA)}, 2018, pp. 1003--1010.

\bibitem{softconstraints}
H.~{Andreasson}, D.~{Adolfsson}, T.~{Stoyanov}, M.~{Magnusson}, and A.~J.
  {Lilienthal}, ``Incorporating ego-motion uncertainty estimates in range data
  registration,'' in \emph{2017 (IROS)}, Sep. 2017, pp. 1389--1395.

\bibitem{p2l_chen_Medioni}
Y.~Chen and G.~Medioni, ``Object modeling by registration of multiple range
  images,'' in \emph{Proceedings. 1991 IEEE International Conference on
  Robotics and Automation}, 1991, pp. 2724--2729 vol.3.

\bibitem{rusinkiewicz-2001-fasticp}
S.~M. Rusinkiewicz, ``Efficient variants of the {ICP} algorithm,'' in \emph{The
  Third International Conference on 3D Digital Imaging and Modeling}, 2001, pp.
  145--152.

\bibitem{magnusson-2009-phd}
M.~Magnusson, ``The three-dimensional normal-distributions transform --- an
  efficient representation for registration, surface analysis, and loop
  detection,'' Ph.D. dissertation, Örebro University, Dec. 2009, Örebro
  Studies in Technology 36.

\bibitem{Almqvist}
H.~Almqvist, M.~Magnusson, T.~P. Kucner, and A.~J. Lilienthal, ``Learning to
  detect misaligned point clouds,'' \emph{Journal of Field Robotics}, vol.~35,
  no.~5, pp. 662--677, 2018.

\bibitem{Stoyanov2012ijrr}
T.~Stoyanov, M.~Magnusson, H.~Andreasson, and A.~J. Lilienthal, ``{Fast and
  accurate scan registration through minimization of the distance between
  compact 3D {NDT} representations},'' \emph{The International Journal of
  Robotics Research}, vol.~31, no.~12, pp. 1377--1393, 2012.

\bibitem{fuzzybnb}
Q.~Liao, D.~Sun, and H.~Andreasson, ``Point set registration for 3d range scans
  using fuzzy cluster-based metric and efficient global optimization,''
  \emph{IEEE Transactions on Pattern Analysis and Machine Intelligence},
  vol.~43, no.~9, pp. 3229--3246, 2021.

\bibitem{Droeschel14localmulti-resolution}
D.~Droeschel and S.~Behnke, ``Local multi-resolution representation for 6d
  motion estimation and mapping with a continuously rotating 3d laser
  scanner,'' in \emph{In Proc. of IEEE Int. Conf. on Robotics and Automation
  (ICRA}, 2014.

\bibitem{barnes_masking_2020}
D.~Barnes, R.~Weston, and I.~Posner, ``Masking by moving: Learning
  distraction-free radar odometry from pose information,'' in \emph{CoRL}, ser.
  CoRL, L.~P. Kaelbling, D.~Kragic, and K.~Sugiura, Eds., vol. 100.\hskip 1em
  plus 0.5em minus 0.4em\relax PMLR, 30 Oct--01 Nov 2020, pp. 303--316.

\bibitem{adolfsson-2021-coral}
D.~Adolfsson, M.~Magnusson, Q.~Liao, A.~J. Lilienthal, and H.~Andreasson,
  ``{CorAl} -- are the point clouds correctly aligned?'' in \emph{European
  Conference on Mobile Robots (ECMR)}.

\bibitem{icp}
P.~J. {Besl} and N.~D. {McKay}, ``A method for registration of 3-d shapes,''
  \emph{IEEE TPAMI}, vol.~14, no.~2, pp. 239--256, Feb 1992.

\bibitem{silva-2005-ga}
L.~Silva, O.~R. Bellon, and K.~L. Boyer, ``Precision range image registration
  using a robust surface interpenetration measure and enhanced genetic
  algorithms,'' \emph{TPAMI}, vol.~27, no.~5, pp. 762--776, May 2005.

\bibitem{Segal_2009}
\BIBentryALTinterwordspacing
A.~Segal, D.~Haehnel, and S.~Thrun, ``Generalized-{ICP},'' in \emph{Robotics:
  Science and Systems V}.\hskip 1em plus 0.5em minus 0.4em\relax Robotics:
  Science and Systems Foundation, jun 2009. [Online]. Available:
  \url{https://doi.org/10.15607%2Frss.2009.v.021}
\BIBentrySTDinterwordspacing

\bibitem{8996777}
H.~Yin, L.~Tang, X.~Ding, Y.~Wang, and R.~Xiong, ``A failure detection method
  for 3d lidar based localization,'' in \emph{2019 Chinese Automation Congress
  (CAC)}, 2019, pp. 4559--4563.

\bibitem{Makadia2006FullyAR}
A.~Makadia, A.~Patterson, and K.~Daniilidis, ``Fully automatic registration of
  3d point clouds,'' \emph{2006 IEEE Computer Society Conference on Computer
  Vision and Pattern Recognition (CVPR'06)}, vol.~1, pp. 1297--1304, 2006.

\bibitem{8500625}
N.~Akai, L.~Y. Moralesl, and H.~Murase, ``Reliability estimation of vehicle
  localization result,'' in \emph{2018 IEEE Intelligent Vehicles Symposium
  (IV)}, 2018, pp. 740--747.

\bibitem{8917111}
R.~Aldera, D.~D. Martini, M.~Gadd, and P.~Newman, ``What could go wrong?
  introspective radar odometry in challenging environments,'' in \emph{2019
  IEEE Intelligent Transportation Systems Conference (ITSC)}, 2019, pp.
  2835--2842.

\bibitem{1642280}
P.~Sundvall and P.~Jensfelt, ``Fault detection for mobile robots using
  redundant positioning systems,'' in \emph{Proceedings 2006 IEEE International
  Conference on Robotics and Automation, 2006. ICRA 2006.}, 2006, pp.
  3781--3786.

\bibitem{David_Landry}
D.~{Landry}, F.~{Pomerleau}, and P.~{Giguère}, ``{CELLO-3D}: Estimating the
  covariance of {ICP} in the real world,'' in \emph{IEEE (ICRA)}, May 2019, pp.
  8190--8196.

\bibitem{bengtsson_robot_2003}
O.~Bengtsson and A.-J. Baerveldt, ``Robot localization based on
  scan-matching—estimating the covariance matrix for the {IDC} algorithm,''
  \emph{Robotics and Autonomous Systems}, vol.~44, no.~1, pp. 29--40, 2003.

\bibitem{nieto_scan-slam_2006}
J.~Nieto, T.~Bailey, and E.~Nebot, ``Scan-{SLAM}: Combining {EKF}-{SLAM} and
  scan correlation,'' in \emph{Field and Service Robotics}, ser. Springer
  Tracts in Advanced Robotics, P.~Corke and S.~Sukkariah, Eds.\hskip 1em plus
  0.5em minus 0.4em\relax Springer, 2006, pp. 167--178.

\bibitem{prakhya_closed-form_2015}
S.~M. Prakhya, L.~Bingbing, Y.~Rui, and W.~Lin, ``A closed-form estimate of 3d
  {ICP} covariance,'' in \emph{2015 14th {IAPR} ({MVA})}, 2015, pp. 526--529.

\bibitem{censi-2007-accurate}
A.~Censi, ``An accurate closed-form estimate of {ICP}'s covariance,'' in
  \emph{Proceedings of the IEEE International Conference on Robotics and
  Automation (ICRA)}, apr 2007, pp. 3167--3172.

\bibitem{5152375}
E.~B. Olson, ``Real-time correlative scan matching,'' in \emph{2009 IEEE
  International Conference on Robotics and Automation}, 2009, pp. 4387--4393.

\bibitem{Bogoslavskyi2017AnalyzingTQ}
I.~Bogoslavskyi and C.~Stachniss, ``Analyzing the quality of matched 3d point
  clouds of objects,'' \emph{2017 IEEE/RSJ International Conference on
  Intelligent Robots and Systems (IROS)}, pp. 6685--6690, 2017.

\bibitem{chandran-ramesh_assessing_2007}
M.~Chandran-Ramesh and P.~Newman, ``\BIBforeignlanguage{en}{Assessing map
  quality and error causation using conditional random fields},''
  \emph{\BIBforeignlanguage{en}{IFAC Proceedings Volumes}}, vol.~40, no.~15,
  pp. 463--468, Jan. 2007.

\bibitem{Fourier_Mellin}
P.~Checchin, F.~G{\'e}rossier, C.~Blanc, R.~Chapuis, and L.~Trassoudaine,
  ``Radar scan matching slam using the fourier-mellin transform,'' in
  \emph{Field and Service Robotics}, A.~Howard, K.~Iagnemma, and A.~Kelly,
  Eds.\hskip 1em plus 0.5em minus 0.4em\relax Berlin, Heidelberg: Springer
  Berlin Heidelberg, 2010, pp. 151--161.

\bibitem{9197231}
Y.~S. {Park}, Y.~S. {Shin}, and A.~{Kim}, ``Pha{R}a{O}: Direct radar odometry
  using phase correlation,'' in \emph{2020 IEEE (ICRA)}, 2020, pp. 2617--2623.

\bibitem{hong2021radar}
Z.~Hong, Y.~Petillot, A.~Wallace, and S.~Wang, ``Radar {SLAM}: A robust slam
  system for all weather conditions,'' 2021.

\bibitem{hong2020radarslam}
Z.~Hong, Y.~Petillot, and S.~Wang, ``Radar{SLAM}: Radar based large-scale slam
  in all weathers,'' in \emph{2020 (IROS)}, 2020, pp. 5164--5170.

\bibitem{burnett_we_2021}
K.~Burnett, A.~P. Schoellig, and T.~D. Barfoot, ``Do we need to compensate for
  motion distortion and doppler effects in spinning radar navigation?''
  \emph{IEEE RAL}, vol.~6, no.~2, pp. 771--778, 2021.

\bibitem{barnes_under_2020}
D.~{Barnes} and I.~{Posner}, ``Under the radar: Learning to predict robust
  keypoints for odometry estimation and metric localisation in radar,'' in
  \emph{(ICRA)}, 2020, pp. 9484--9490.

\bibitem{burnett2021radar}
K.~Burnett, D.~J. Yoon, A.~P. Schoellig, and T.~D. Barfoot, ``Radar odometry
  combining probabilistic estimation and unsupervised feature learning,'' 2021.

\bibitem{8460687}
S.~H. {Cen} and P.~{Newman}, ``Precise ego-motion estimation with
  millimeter-wave radar under diverse and challenging conditions,'' in
  \emph{2018 IEEE International Conference on Robotics and Automation (ICRA)},
  2018, pp. 6045--6052.

\bibitem{8793990}
S.~H. Cen and P.~Newman, ``Radar-only ego-motion estimation in difficult
  settings via graph matching,'' in \emph{2019 International Conference on
  Robotics and Automation (ICRA)}, 2019, pp. 298--304.

\bibitem{kung2021normal}
P.-C. Kung, C.-C. Wang, and W.-C. Lin, ``A normal distribution transform-based
  radar odometry designed for scanning and automotive radars,'' in \emph{2021
  IEEE International Conference on Robotics and Automation (ICRA)}, 2021, pp.
  14\,417--14\,423.

\bibitem{biber_normal_2003}
P.~Biber and W.~Strasser, ``The normal distributions transform: a new approach
  to laser scan matching,'' in \emph{Proceedings 2003 {IEEE}/{RSJ} ({IROS}
  2003)}, vol.~3, Oct. 2003, pp. 2743--2748 vol.3.

\bibitem{makadia-2006-fully}
A.~{Makadia}, A.~{Patterson}, and K.~{Daniilidis}, ``Fully automatic
  registration of 3d point clouds,'' in \emph{2006 IEEE Computer Society
  Conference on Computer Vision and Pattern Recognition (CVPR'06)}, vol.~1,
  June 2006, pp. 1297--1304.

\bibitem{Razlaw2015EvaluationOR}
J.~Razlaw, D.~Droeschel, D.~Holz, and S.~Behnke, ``Evaluation of registration
  methods for sparse 3d laser scans,'' \emph{2015 European Conference on Mobile
  Robots (ECMR)}, pp. 1--7, 2015.

\bibitem{8461000}
D.~{Droeschel} and S.~{Behnke}, ``Efficient continuous-time slam for 3d
  lidar-based online mapping,'' in \emph{ICRA}, May 2018, pp. 1--9.

\bibitem{magnusson-2007-jfr}
M.~Magnusson, A.~J. Lilienthal, and T.~Duckett, ``Scan registration for
  autonomous mining vehicles using {3D-NDT},'' \emph{Journal of Field
  Robotics}, vol.~24, no.~10, pp. 803--827, Oct. 2007.

\bibitem{Pomerleau_2012}
F.~Pomerleau, M.~Liu, F.~Colas, and R.~Siegwart, ``{Challenging data sets for
  point cloud registration algorithms},'' \emph{IJRR}, vol.~31, no.~14, pp.
  1705--1711, Dec. 2012.

\bibitem{sheeny2020radiate}
M.~Sheeny, E.~De~Pellegrin, S.~Mukherjee, A.~Ahrabian, S.~Wang, and A.~Wallace,
  ``Radiate: A radar dataset for automotive perception,'' \emph{arXiv preprint
  arXiv:2010.09076}, 2020.

\bibitem{gskim-2020-mulran}
G.~Kim, Y.~S. Park, Y.~Cho, J.~Jeong, and A.~Kim, ``{M}ul{R}an: Multimodal
  range dataset for urban place recognition,'' in \emph{Proceedings of the IEEE
  International Conference on Robotics and Automation (ICRA)}, Paris, May 2020.

\bibitem{RadarRobotCarDatasetICRA2020}
D.~Barnes, M.~Gadd, P.~Murcutt, P.~Newman, and I.~Posner, ``The {O}xford radar
  robotcar dataset: A radar extension to the {O}xford robotcar dataset,'' in
  \emph{2020 IEEE ICRA}, 2020, pp. 6433--6438.

\end{thebibliography}

\end{document}